\documentclass{article}

\usepackage[nonatbib, final]{neurips}

\usepackage[utf8]{inputenc} \usepackage[T1]{fontenc}    \usepackage{hyperref}       \usepackage{url}            \usepackage{booktabs}       \usepackage{amsfonts}       \usepackage{nicefrac}       \usepackage{microtype}      
\usepackage{wrapfig}
\usepackage{subcaption}
\usepackage{overpic}
\usepackage{graphicx}
\usepackage{xcolor}
\usepackage{amsmath}

\newcommand{\mySpaceTweak}[0]{\vspace{-0.16cm}}

\definecolor{LukasGreen}{HTML}{008800}
\newcommand{\nt}[1]{{\color{purple}#1}}  \newcommand{\bu}[1]{{\color{orange}#1}} 
\renewcommand{\nt}[1]{#1} \renewcommand{\bu}[1]{#1} 
 
\makeatletter
\newcommand{\dummylabel}[2]{\def\@currentlabel{#2}\label{#1}}
\makeatother

\newcommand{\eq}[1]{Eq.~\ref{#1}} 
\title{Guaranteed Conservation of Momentum\\ for Learning Particle-based Fluid Dynamics}

\author{  Lukas Prantl \\
  Technical University of Munich\\
  \texttt{lukas.prantl@tum.de} \\
    \And
  Benjamin Ummenhofer \\
  Intel Labs \\
  \AND
  Vladlen Koltun \\
  Apple \\
  \And
  Nils Thuerey \\
  Technical University of Munich \\
}

\begin{document}

\maketitle

\begin{abstract}
We present a novel method for guaranteeing linear momentum in learned physics simulations. Unlike existing methods, we enforce conservation of momentum with a hard constraint, which we realize via antisymmetrical continuous convolutional layers. We combine these strict constraints with a hierarchical network architecture, a carefully constructed resampling scheme, and a training approach for temporal coherence. In combination, the proposed method allows us to increase the physical accuracy of the learned simulator substantially. In addition, the induced physical bias leads to significantly better generalization performance and makes our method more reliable in unseen test cases. 
We evaluate our method on a range of different, challenging fluid scenarios. 
Among others, we demonstrate that our approach generalizes to new scenarios with up to one million particles. 
Our results show that the proposed algorithm can learn complex dynamics while outperforming existing approaches in generalization and training performance.
An implementation of our approach is available at
\url{https://github.com/tum-pbs/DMCF}.

\end{abstract}

\mySpaceTweak{}
\section{Introduction}

Learning physics simulations with machine learning techniques opens up 
a wide range of intriguing paths 
to predict the dynamics of complex physical phenomena such as fluids \cite{morton2018deep,liLearningParticleDynamics2019,ummenhofer2019lagrangian,sanchez2020learning}. As traditional simulators for fluids employ handcrafted simplifications \cite{pope2000turbulent} and require vast amounts of computational resources \cite{tezduyar1996flow}, recent learning-based methods have shown highly promising results. In face of the impressive performance and accuracy of these techniques, it is surprising to see that they still neglect some of the most basic physical principles, like the symmetries that arise from Newton's third law of motion and the conservation of momentum.
This makes learned simulators less predictable and can lead to severe implications when basing decisions on the simulation outcomes.

Predicting physical properties with neural networks is commonly treated as a regression problem \cite{mohan2019compressed,wang2020towards,wiewel2020latent}, where the training signal is defined as a soft constraint.
This simple and desirable formulation allows to effectively learn and approximate physical processes but also gives way to unwanted shortcuts that deviate from the basic laws of physics.
At the same time network architectures are usually designed with generalizability in mind, 
e.g., with applications ranging from geometry processing \cite{qi2017pointnet++} to physical simulations \cite{battaglia2016interaction}.
In contrast, we propose a lean and efficient architecture that provides a stable and large receptive field, while adhering to the desired physical constraints. At training time, we dynamically take the unrolled evolution of the physical state into account to ensure stability when training with longer sequences.

For the implementation of the inductive bias, our approach is inspired by \emph{Smoothed Particle Hydrodynamics} (SPH) methods \cite{gingold1977smoothed,KBST19}, where physical properties are preserved by the appropriate design of the smoothing kernels \cite{BONET199997}.
Our method uses learned convolution kernels that have inherent benefits over general graph structures in the context of physics simulations, as they can easily process positional information.
Instead of using a soft constraint, e.g., by integration into the loss function, our method enforces physical properties by using inductive biases while keeping the simple training signal of a regression problem.
To this end, we integrate symmetries, which are an integral part of the underlying physical models, into our neural network.
This results in guaranteed conservation of momentum for our method.

\mySpaceTweak{}
\subsection{Related Work}

Numerical computation of fluid-based dynamics is a long and extensively treated topic in research \cite{harlow1955machine,wilcox1998turbulence}.
We discuss the most relevant work here, and refer to surveys \cite{koschier2022survey} for a full review.
Our approach uses a Lagrangian viewpoint, where fluids are represented as smoothed particles.
This viewpoint by design conserves the mass of the fluid, as the number of particles does not change, and limits valuable computation time to where the fluid is.
This idea stems from astrophysics
\cite{gingold1977smoothed} and is known as \emph{Smoothed Particle Hydrodynamics} (SPH).
It was followed by a large number of papers that improved the SPH method in terms of accuracy and performance \cite{Becker2007,solenthaler2009predictive,macklin2013position,bender2016divergence,hu2019difftaichi,adami2012}.

In contrast to many classic simulators that use sophisticated and handcrafted \nt{model equations} to describe the motion of fluids, our method belongs to the class of data-driven solvers, which entirely learns their dynamics from observations.
A seminal example of learning fluid simulation from data is Ladicky et al. \cite{ladicky2015data}, which uses carefully designed density features with random forests to regress fluid motions. However, the method does not guarantee momentum conservation.
In recent work, algorithms have been proposed to use physical priors based on Lagrangians \cite{saemundsson2020,  lutter2019, cramer2020} or Hamiltonians \cite{sanchez2019hamiltonian} to improve the physical accuracy of learning ODEs and PDEs, and compliance with conservation laws. Nevertheless, the methods are limited to low-dimensional problems and do not address complex PDEs such as those required for fluids.
Other works have used convolutional neural networks, or ConvNets, to achieve acceleration for grid-based fluid simulations  \cite{tompson2016accelerating,um2020solver,kochkov2021machine}.
The former proposes ConvNets to accelerate the expensive pressure correction step \cite{tompson2016accelerating},
while the other two propose ConvNets learning corrections to reduce the grid size and, consequently, the runtime of simulations.
Although these methods show large speed-ups, they inherently cannot be applied to particle-based simulations.

For processing point clouds, the community has followed multiple strategies to address the challenges with this representation.
A key challenge is the permutation invariance of particles. 
An approach to address this uses a combination of point-wise MLPs, order-independent set functions, and farthest point sampling to create a point cloud hierarchy \cite{qi2017pointnet++}.  
Another line of work processes point clouds with graph neural networks.
These methods represent fluid particles with graph nodes and define edges between them \cite{battaglia2016interaction,liLearningParticleDynamics2019,sanchez2020learning}.
Here, interaction networks with relation-centric and object-centric functions were developed to predict future object states \cite{battaglia2016interaction}.
Since the framework aims to be general and employs unconstrained MLPs, it cannot be guaranteed that interactions are symmetric.
The same holds for Li et al. \cite{liLearningParticleDynamics2019}, which uses the same framework and builds a dynamic graph connecting nearby particles to model fluids.
SPH simulations can also be treated as 
\nt{message-passing on graphs \cite{sanchez2020learning, pfaff2020learning, brandstetter2022message}.} 
These approaches typically decompose a simulator into multiple MLPs, which act as encoder, processor, and decoder.
The processor performs multiple rounds of general message-passing, updates node and edge features, and does not enforce physical properties like momentum conservation.
Further, graph-based methods are strongly tied to the chosen particle discretization, and changes to it, like the sampling density, may require non-trivial changes to the models.

As an alternative to explicit graph representations, convolutional architectures represent an interesting option for particle-based simulation.
Schenck et al.~\cite{schenckSPNetsDifferentiableFluid2018} implement special layers for particle-particle interactions and particle-boundary interactions.
The work does not aim to learn a general fluid simulator but implements a differentiable position-based fluids (PBF) \cite{macklin2013position} solver, which is used to estimate liquid parameters or train networks to solve control problems by backpropagation through the solver. Moreover, Fourier neural operators \cite{li2020fourier} or their adaptive variants \cite{guibas2021afno} have been proposed to solve PDEs.
Most related to our approach is Ummenhofer et al.~\cite{ummenhofer2019lagrangian}, which implements a trainable continuous convolution operator resulting in compact networks.
The learned kernels describe general continuous functions with a smooth radial falloff but do not account for symmetric interactions of particles.
We extend the continuous convolutions to enable simulations that guarantee momentum conservation and introduce a particle hierarchy for better accuracy and robustness.

Symmetries play an essential role in physics, as isolated systems are invariant to certain transformations, i.e., the behavior of a physical system does not depend on its orientation. For example, Lie point symmetries were used to optimize data augmentation and thus improve the sample efficiency for learning PDEs \cite{brandstetter2022pointsym}.
Similar observations apply for the structural analysis of molecules \cite{eismann2021hierarchical} and vision applications \cite{symconv,Son_2020_ACCV}.
The chemical properties of proteins do not change under rotation, nor should class labels of objects in images.
Rotationally equivariant architectures have been proposed to leverage this concept for improved generalization and sample efficiency, e.g., constraining the convolution kernels for the $E(2)$ group to achieve equivariance \cite{weiler2019general}. A similar approach by Wang et al.~\cite{wang2020incorporating} was applied to physics-based problems.
Kernel constraints can also be transferred to continuous convolutions on 2D point clouds for improved sample efficiency of traffic trajectory predictions \cite{walters2020trajectory}.
For 3D problems, spherical harmonics can be used to define rotationally equivariant filters \cite{thomas2018tensor}, which shows their application to shape classification, molecule generation, and simple problems in classical mechanics.
The method guarantees rotational equivariance, but filter evaluations can become expensive for higher degrees.
In addition, architectures for symmetries in arbitrary dimensions were proposed \cite{villar2021scalars,satorras2021n}, which however require graph-based representations.

\mySpaceTweak{}
\section{Method}
\begin{wrapfigure}{t}{0.5\linewidth} 
	\centering
	\vspace{-3mm}
	\includegraphics[width=\linewidth, trim= 750 380 750 300, clip]{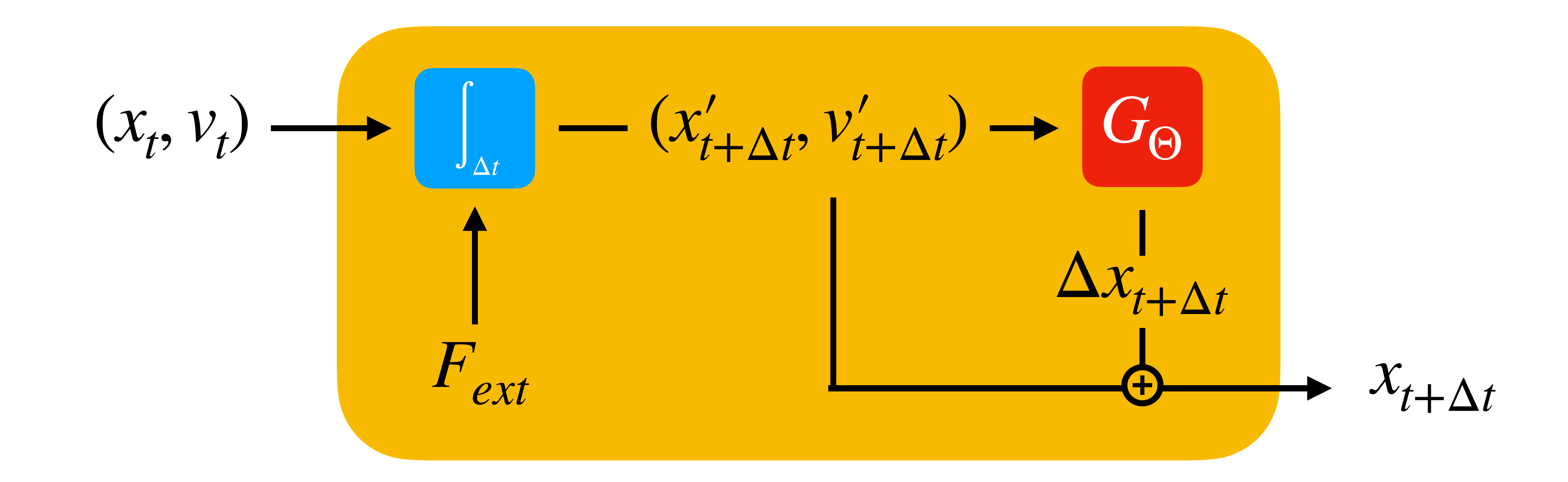}
	\caption{
	    	    Each time step of our method performs a predictive time integration step (blue) for position $x_t$ and velocity $v_t$ of a set of particles at time $t$. A neural network $G$ (red) provides a position update to obtain $x_{t+\Delta t}$. 
	}
	\vspace{-3mm}
	\label{fig:func}
\end{wrapfigure}
Given a physical quantity $P_t \subseteq \mathbb{R}^d$ at a time $t$
in the form of particles,
let $x_t \in P_t$ denote the position, $v_t \in \mathbb{R}^d$ the corresponding velocity and $d$ the spatial dimension.
The central learning objective is to
approximate the underlying physical dynamics to predict the state of $P_{t+\Delta t}$ after a period of time $\Delta t$.
We integrate the velocities $v_{t}$ and positions $x_t$ of the physical system in time while taking into account external forces $F_{ext}$, e.g., gravity, as illustrated in Fig.~\ref{fig:func}. For an initial prediction, an explicit Euler step of the form
$v'_{t+\Delta t} = v_t + \Delta t \frac{F_{ext}}{m}$ and
$x'_{t+\Delta t} = x_t + \Delta t ~ v'_{t+\Delta t}$, is used, where $m$ represents the mass of the particles.
The resulting physics state is passed to a neural network $G$ with trainable parameters $\Theta$, which infers the residual of the position $\Delta x_{t+\Delta t}$ such that particles match a set of desired ground-truth positions $y_{t+\Delta t}$.
Velocities are obtained with a position-based update. This yields:
\begin{equation}
\begin{split}
    x_{t+\Delta t} = x'_{t+\Delta t} + \Delta x_{t+\Delta t} \ ; \quad
    v_{t+\Delta t} = \frac{x_{t+\Delta t} - x_t}{\Delta t}.
\end{split}
\end{equation}
Where the minimization target of the neural network $G$ is as follows:
\begin{equation}
    \min\limits_{\Theta} L(x'_{t+\Delta t} + G(x'_{t+\Delta t},v'_{t+\Delta t},\Theta), y_{t+\Delta t}),
\end{equation}
with $L$ denoting a distance-based loss function, and $y_{t+\Delta t}$ the ground-truth positions.
By separating the external forces and the update inferred by the neural network, the conservation of momentum only depends on the latter. 
In the next section, we derive which properties the network must fulfill to achieve this.

\mySpaceTweak{}
\subsection{Conservation of Momentum}
The linear momentum of a physical system is given by $M = \int_{x \in P} m_x v_x,$,
where $P$ represents an arbitrary control volume in the system, $m_x$ the mass, and $v_x$ the velocity of particle $x$.
It follows that the rate of change of the momentum in the absence of external forces $F_{ext}$, according to Newton's second law $m_x a_x = F_{ext} - F_x$, is given by: 
\begin{equation}
    M' = \int_{x \in P} m_x a_x = -\int_{x \in P} F_x,
\end{equation}
where $F_x$ represents the internal forces of the physical system, and $a_x$ denotes acceleration. Hence the central condition for preserving linear momentum is
\begin{equation}
\label{eq:mom}
    \int_{x \in P} F_x = 0.
\end{equation}
It is theoretically possible to include this conservation law as a minimization goal. However, this would merely impose a soft constraint that needs to be weighed against the other terms of the learning objective.
We instead propose to use the conservation of momentum as an inductive bias at training time in the form of a hard constraint.

In order to guarantee that momentum is conserved,
we work with convolutional layers with specially designed kernels.
We define the convolution in the continuous domain as
\begin{equation}
\label{eq:cconv}
        (f*g)(x) = \int_{y \in P_Q} f(y) g(y-x), \qquad x \in P_D, 
\end{equation}
where $g$ represents a learnable kernel function and $f$ is the feature vector of a quantity that should be processed in the convolution operation.
Here $P_D$ denotes a set of data points, while $P_Q$ denotes a set of query points on which the convolution is performed. 
The use of convolutional layers already guarantees permutation invariance and translation equivariance.

However, instead of just relying on this learned kernel function, we additionally modify the convolution to further assist in ensuring conservation of momentum.
For the deduction, we reformulate the residual position $\Delta x_{t+\Delta t}$ that the network generates for a given position $x_t$ as the result of an internal force $F_x$:
\begin{equation}
    F_x = m \frac{\Delta x_{t+\Delta t}}{\Delta t^2}.
\end{equation}
In general, the internal force at the position $x$ can be expressed as the integral of all pair-wise interaction forces exerted on $x$, i.e. $F_x = \int_{y \in P} F_{xy}$.
Following \eq{eq:mom}, to enforce conservation of momentum we have to ensure
\begin{equation}\label{eq:intf}
    \int_{x \in P} \int_{y \in P} F_{xy} = 0.
\end{equation}
If it is guaranteed that for every internal force $F_{xy}$ a corresponding
opposing force exists, i.e.
\begin{equation}
\label{eq:momentum}
F_{xy} = -F_{yx}, \qquad \forall x, y \in P,
\end{equation}
then the integral of \eq{eq:intf} evaluates to zero,
and the condition for the conservation of momentum is fulfilled. 
This corresponds to an antisymmetry of the internal forces in the fluid.

Applied to convolutional layers, we achieve this property by adjusting the convolution \eq{eq:cconv} as follows:
\begin{equation}
\label{eq:sym_cconv}    
(f*g_s)(x) = \int_{y \in P_Q} (f(x) + f(y)) g_s(y-x), \qquad x \in P_D, 
\end{equation}
where $g_s$ is an antisymmetric kernel with the restriction that 
\begin{equation}
\label{eq:eq_cond}  
    P_D = P_Q ,
\end{equation}
and $f(x)$ is the feature vector at the position $x$, where the convolution is evaluated. 
The condition \ref{eq:eq_cond} is necessary to guarantee that 
there is a bidirectional connection between two points from the two sets. This is only possible if both points exist in both sets.
As before, let us consider the relative features $f_{xy} = (f(x) + f(y)) g_s(y-x)$, isolated for two points in space $x, y \in P$.
The kernel $g_s$ is antisymmetric, i.e. $g_s(x) = -g_s(-x)$. From this follows:
\begin{equation}
f_{xy} = (f_x + f_y) (-g_s(x-y)) = -f_{yx}. 
\end{equation}
Due to the restriction from \eq{eq:eq_cond} 
$f_{yx}$ exists for every $f_{xy}$ 
and hence \eq{eq:momentum} is satisfied.

\mySpaceTweak{}
\subsection{Antisymmetric Continuous Convolution}
\label{sec:ascc}
To implement the convolutional layer fulfilling the constraints above, we make use of continuous convolutions (CConv)  \cite{ummenhofer2019lagrangian}. While the constraints could likewise be realized with other forms of convolutional layers, CConvs are adapted for efficiently processing unstructured data, such as particle-based data from fluids, and hence provide a suitable basis to demonstrate the impact of the antisymmetric constraints.
CConv layers determine the nearest neighbors from a given set of data points $P_D$ for a given set of query points $P_Q$ and aggregate their features $f$. Thereby, the features of the neighbors are weighted with a kernel function depending on their relative position. The kernel functions themselves are discretized via a regular grid with a spherical mapping and contain the learnable parameters.

The original CConvs are defined as:
$\text{CConv}_g(f, P_{D}, P_{Q}) = (f*g)(x) =$ $\sum_{k} f_k g(x_k-x)$ for $k \in \mathcal{N}_r(P_D, x)$ and $x \in P_Q$.
This corresponds to a discretized version of \eq{eq:cconv}, where $\mathcal{N}_r$ represents the nearest neighbor search for a fixed radius $r$.
To conserve momentum, we adapt the equation according to \eq{eq:sym_cconv} as follows:
\begin{equation}
\label{eq:sym_cconv_d}
     \text{ASCC}_{g_s}(f, P_{D}, P_{Q}) = (f*g_s)(x) = \sum_{k \in \mathcal{N}_r(P_D, x)} (f + f_k) g_s(x_k-x), \qquad x \in P_Q,
\end{equation}
\begin{wrapfigure}{t}{0.4\linewidth} 
	\vspace{-4mm}
	\centering
	\includegraphics[width= \linewidth]{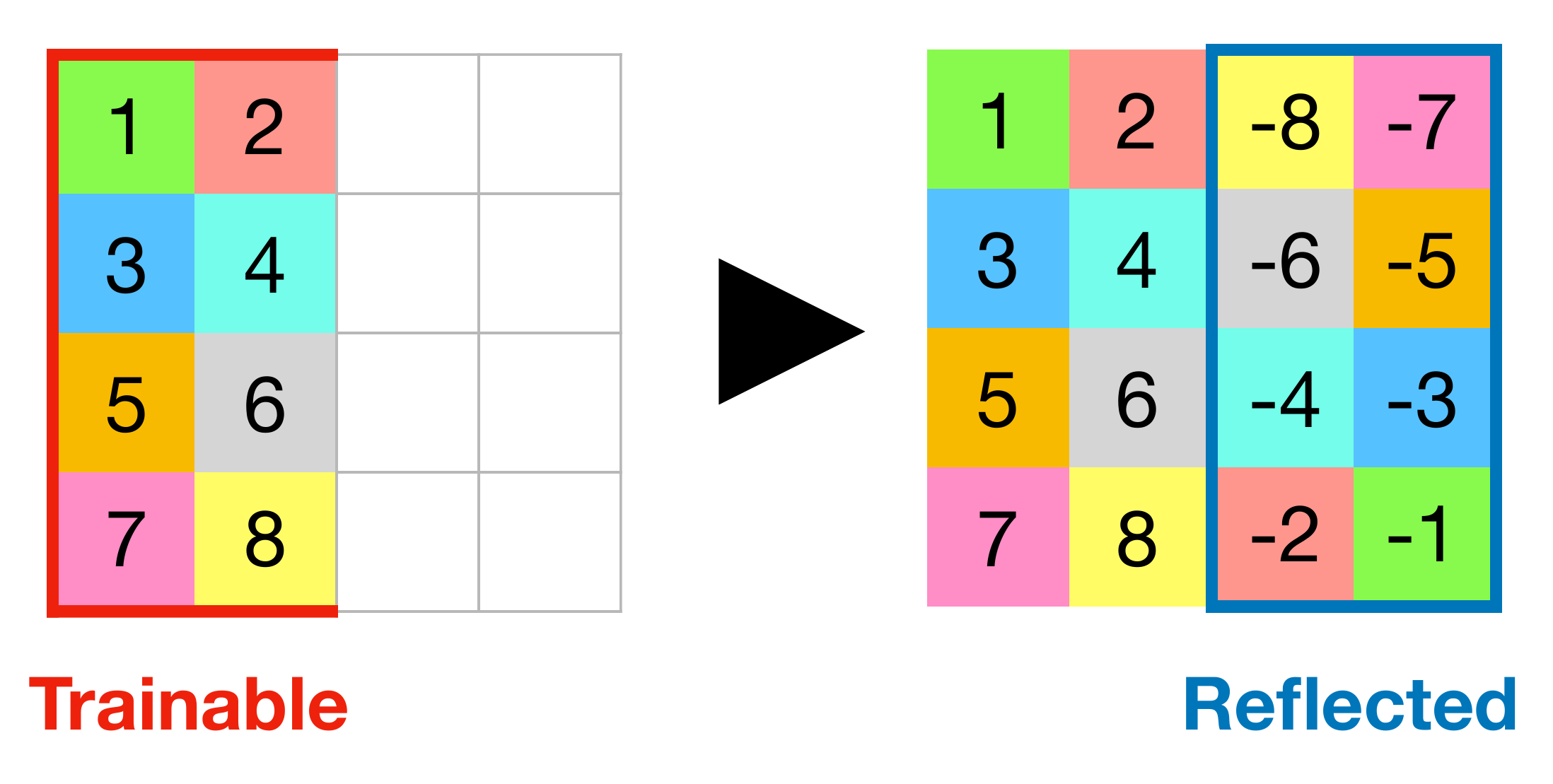}
	\caption{
	    The trainable variables are negated and mirrored by the center point. 
	    This results in an antisymmetric kernel.
	}
	\label{fig:sym_kernel}
	\vspace{-4mm}
\end{wrapfigure}
with the restriction that $P_D = P_Q$.
Here the discrete, antisymmetric kernel $g_s$ consists of a grid with a user-defined size. To obtain the antisymmetric property we halve the learnable kernel parameters along a chosen axis and determine the second half by reflection through the center of the kernel. Additionally, the mirrored values are negated, as shown in Fig.~\ref{fig:sym_kernel}. 
Mathematically, the mirroring fulfills \eq{eq:sym_cconv},
and hence the axis of reflection for the kernel weights can be chosen freely.
The convolution is evaluated with the reflexive term $f+f_k$ from \eq{eq:sym_cconv} to preserve the antisymmetry. 
In the following, we refer to the proposed \emph{antisymmetric CConv} layer as \textit{ASCC}.

\mySpaceTweak{}
\subsection{Neural Network Formulation}
\label{sec:arch}
\begin{figure}[t]
	\centering
	\includegraphics[width=\linewidth, trim= 150 150 150 250 clip]{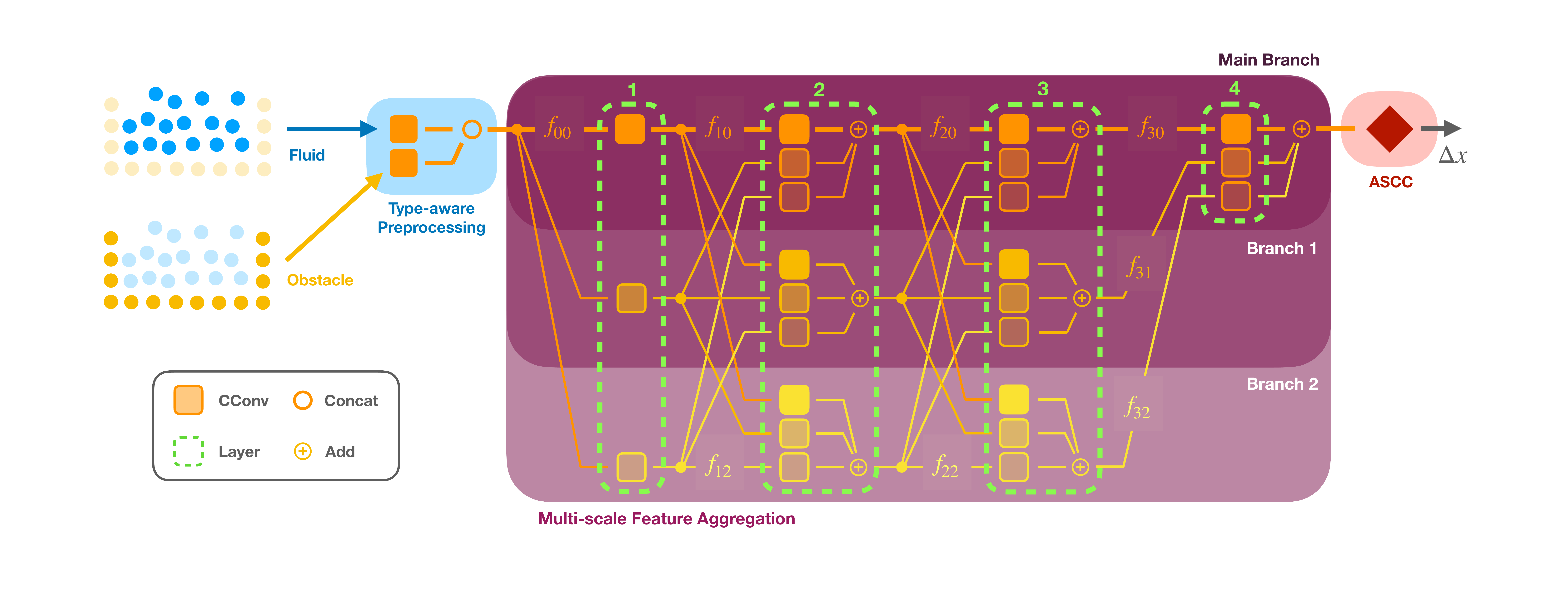}
	\caption{
	    Neural network architecture: The colored squares symbolize the different CConv blocks, whereas the rotated square in red represents the antisymmetric layer. 
	    The color shows the used query point set. Orange corresponds to the original point set, with each set below halving the resolution. 
	    	    Search radii are enlarged accordingly.
	    	    	    	    	}\label{fig:arch}
\end{figure}

To construct a full neural network model that preserves momentum, we build on a CConv architecture, replacing the output layer with an ASCC layer, as shown in Fig.~\ref{fig:arch}. We deliberately replace only the last layer, as this maximizes the flexibility of the previous layers and still guarantees momentum preservation.
While antisymmetric kernels have highly attractive physical properties, they intentionally restrict the scope of actions of a neural network. The resulting networks generalize much better, but the initial learning objective is more complex than for unconstrained networks. 
Regular networks can take unphysical shortcuts by overfitting to problem-specific values. A corresponding evaluation of the generalization capabilities of ASCCs is provided in Sec.~\ref{sec:column}.

\mySpaceTweak{}
\paragraph*{Layer Hierarchy}
\begin{wrapfigure}{t}{0.4\linewidth} 
    \vspace{-0.65cm}
	\centering
	\includegraphics[width=0.49\linewidth, trim= 500 0 500 0, clip]{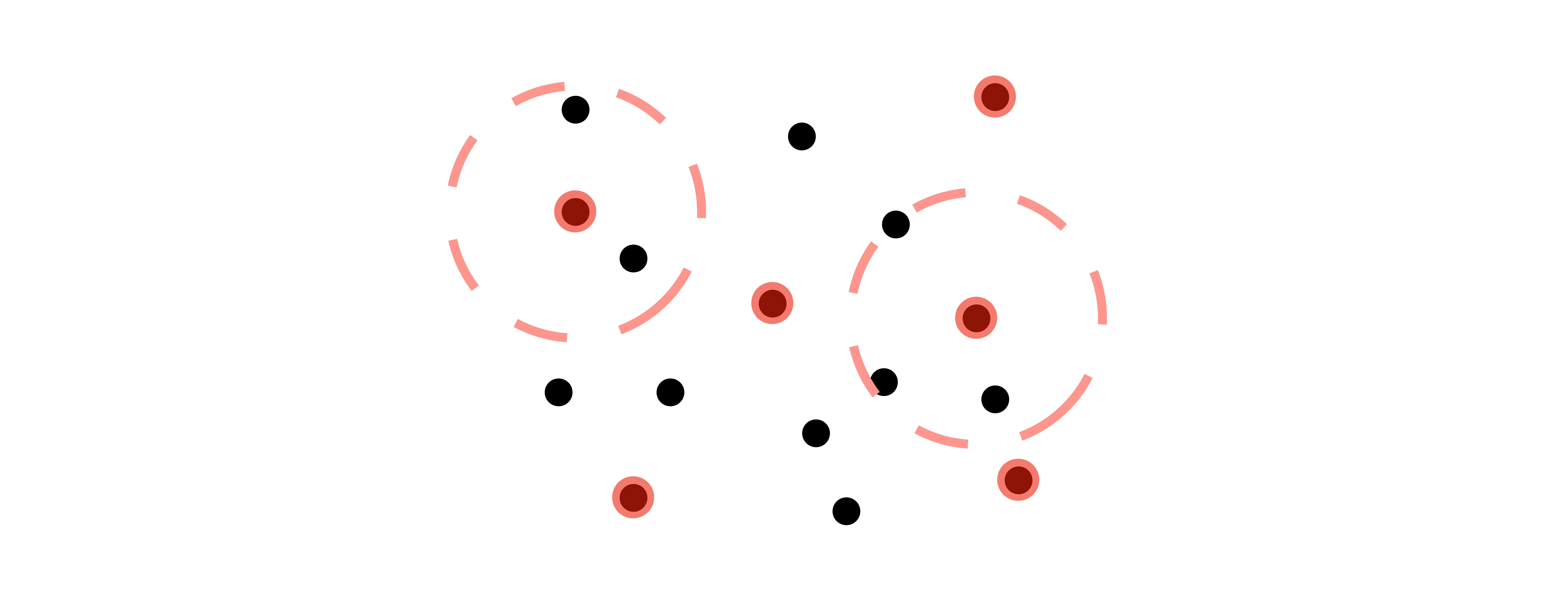}
	\includegraphics[width=0.49\linewidth, trim= 500 0 500 0, clip]{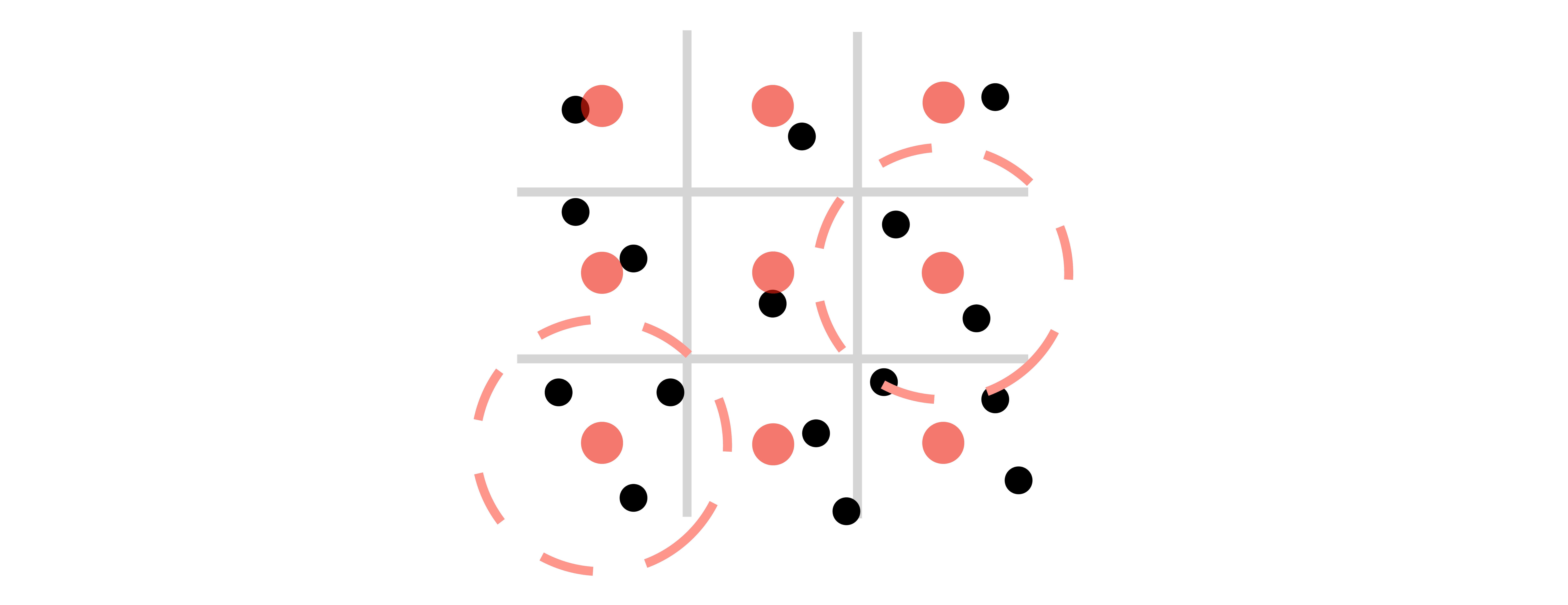}
	\caption{
	    Farthest point sampling (left) vs. voxel sampling (right). Data points in black, sampling points in red.
	    The red circles indicate sampling region's 	    for feature aggregation.
	}\label{fig:sampling}
    \vspace{-4mm}
\end{wrapfigure}
To support learning generalizable physical dynamics, we employ a hierarchical network structure that allows for a large receptive field.
Specifically, we use an architecture inspired by the parallel processing of the multi-scale feature aggregator of HRNet \cite{hrnet}. 
It processes sets of extracted features at different spatial scales with separate branches of the network. Akin to pooling layers, the density of the sampling points is reduced successively in each branch.
The structure of the network and its branches are shown in Fig.~\ref{fig:arch}.

The main branch of the network works on the full set of sample points $P_{Q0}$ which is in this case equal to the data points $P_D$. For the secondary branches, a new set of sampling points $P_{Qi}$ is generated from the input points $P_D$ with a scaling factor $s_i$ for the corresponding branch $i$ such that 
$s_i \approx \rho(P_{Qi})/\rho(P_{Q0})$, with $\rho$ denoting the density function of the given point set.
Additionally, points in each set should be spatially distributed as evenly as possible so that the scaling factor remains similar across all partial volumes. To satisfy condition \ref{eq:eq_cond}, the point features are resampled back to original sample points $P_{Q0}$ before processing them in the ASCC layer.

Generating the sampling points $P_{Qi}$ of a layer is a non-trivial task, for which previous work typically employs 
farthest-point sampling (FPS)  \cite{qi2017pointnet++}.
While FPS extracts a uniformly distributed subset from a point cloud,
its time complexity of $\mathcal{O}(N log N)$ is not favorable.
We have found that a \emph{voxelization} approach is preferable, which in a similar form is already used in established computer vision methods \cite{voxelnet, song2015}. We choose the centers of the cells of a regular grid, i.e. voxels, as sampling points, as  illustrated in Fig.~\ref{fig:sampling}. The spacing of the grid is determined from the initial particle sampling modified by the scaling factor $s_i$. The voxelization can be performed in $\mathcal{O}(N)$, and results in an evenly distributed set of sampling points.

\mySpaceTweak{}
\paragraph*{Feature Propagation}
To compute the features $f_{ij}$ for level $i$ and branch $j$, we use a set of CConvs that process the features of all branches from the previous pass for each branch. This results in a multi-level feature aggregation across all branches. For merging the features, the different features are accumulated via summation:
\begin{equation}
    f_{ij} = \sum\limits_{k} \text{CConv}(f_{(i-1),k}, P_{Qk}, P_{Qj}),
\end{equation}
where $k$ denotes all branch indices of the previous layer, 
as also illustrated in Fig.~\ref{fig:arch}.
Here, it is important 
to choose the sampling radius of the convolution operator based on the scale of the corresponding input branch $s_k$. This ensures that the convolutions on average always process a neighborhood of similar size. I.e., for a small input resolution a larger radius is used, and vice versa. 
Finally, the accumulated features of the main branch are processed by an ASCC layer. 

\mySpaceTweak{}
\subsection{Training and Long-term Stability}
\label{sec:training}
Following previous work \cite{liLearningParticleDynamics2019,ummenhofer2019lagrangian,sanchez2020learning}, we use a mean absolute error of the position values between prediction and ground truth weighted by the neighbor count as our loss function:
\begin{equation}
    L(t) = \frac{1}{|P|} \sum_i e^{-\frac{c_i}{c_{\text{avg}}}} |x_{t,i} - y_{t,i}|
\end{equation}
Here, $c_i$ denotes the number of neighbors for particle $i$, and $c_{\text{avg}}$ the average neighbor count.
We purposefully formulate the loss to work without additional loss terms, such as physical soft constraints, and instead rely on the conservation of momentum from the ASCC.

For temporal stability, we use a rollout of $T$ frames at training time. That is, for each training iteration, we run the network for $T$ time steps, where the input for the next step is based on the result from the previous prediction. The loss is evaluated and averaged across all $T$ frames with 
$L_r = \frac{1}{T} \sum_{t=0}^T L(t)$.
This integrates the temporal behavior in the training and stabilizes it 
for a medium time range.

While larger $T$ are preferable for improving long-term predictions, they lead to significantly increased memory requirements for backpropagation.
To counteract this effect, we prepend the rollout by a sequence of $W$ simulated preprocessing steps using the current training state. Altogether $N=W+T$ frames are generated, but irrespective of the number of preprocessing steps, the loss and hence gradient, is only computed for $T$ steps after preprocessing. 
This lets the network learn to correct its own errors that occur over longer time spans and improves long-term stability even in complex scenarios without increasing the memory. During training, we adapt $W$ based on the training progress and the example difficulty. We provide the details of the implementation in App. \ref{sec:train_details}.

\mySpaceTweak{}
\section{Results}\label{sec:results}
We perform a series of tests to show the physical accuracy, robustness, and generalizability of our method compared to other baseline methods. 
We primarily use data from a high-fidelity SPH solver with adaptive time stepping \cite{adami2012}. The resulting, two-dimensional dataset \texttt{"WBC-SPH"} consists of randomly generated obstacle geometries and fluid regions. Gravity direction and strength are additionally varied across simulations. In addition to this primary dataset, we also use the MPM-based fluid dataset \texttt{"WaterRamps"} from Sanchez-Gonzalez et al. \cite{sanchez2020learning}, and the three-dimensional liquid dataset \texttt{"Liquid3d"} from Ummenhofer et al. \cite{ummenhofer2019lagrangian} for additional evaluations. Both consist of randomized fluid regions with constant gravity. A more detailed description of the datasets is provided in App. \ref{sec:sim_data}.

To quantify the accuracy of the methods, we use several different metrics with respect to the ground truth data from the corresponding dataset.
We use the root-mean-squared error (RMSE) for single simulation steps to evaluate short-term accuracy. However, over long sequences, it is common for fluid particles to mix chaotically regardless of whether the general fluid behavior resembles the reference, leading to artificial per-particle errors.
Therefore, 
the earth mover's distance (EMD) is used as an assessment of the long-time accuracy over the fluid volume as a whole \cite{FanSG16}: $\min_{\phi :x \to y}\sum_{x}\|\phi(x)-y\|_2^2$, where $x$ and $y$ are positions from two point sets and $\phi$ denotes a bijective mapping of minimal distances.
EMD matches the particles as probability distributions, such that the global distance error is minimized. The deterministic assignment of the particles makes EMD agnostic to the ordering of particles, which counteracts the assessment errors caused by mixing. 
Intuitively, EMD evaluates the similarity of density-weighted volumes instead of the individual particles. Thus, it quantifies the error of the simulated configuration relative to the reference, neglecting small-scale changes in particle order.
In addition, we evaluate the difference between the velocity distributions to evaluate the accuracy of the velocity content of a simulation. We compute a histogram of the magnitude of the velocities and compute its Jensen-Shannon divergence to the histogram from ground truth data:
$\text{JSD}(V_x\|V_y) = \frac{1}{2} (D_{KL}(V_x\|V) + D_{KL}(V_y\|V))$,
where $V = \frac{1}{2}(V_x + V_y)$ and $D_{KL}$ is the Kullback-Leibler divergence.
We also evaluate deviations in terms of maximum density, $|1 - (\max_{i} \rho(x_i) / \max_{i} \rho(y_i))|$,
with $\rho$ being a function to compute the density of particle $i$.
This evaluation correlates with the compressibility of the fluid and gives an indication of the stability of a method. Lastly, we also compute the linear momentum change as $\sum\limits m_i a_i - F_{ext}$,
where $a_i$ is the acceleration and $m_i$ the mass of a particle, and $F_{ext}$ denotes external forces.

The source code of our approach, together with the associated evaluation tools and links to datasets, is available at
\url{https://github.com/tum-pbs/DMCF}.

\mySpaceTweak{}
\paragraph{Standing Liquid} 
\label{sec:column}
\begin{wrapfigure}{t}{0.5\linewidth} 
    \vspace{-4mm}
	\centering
    \begin{subfigure}[c]{0.49\linewidth}
	\hspace{4mm}
	\includegraphics[height=\linewidth, trim= 1000 0 1000 0, clip]{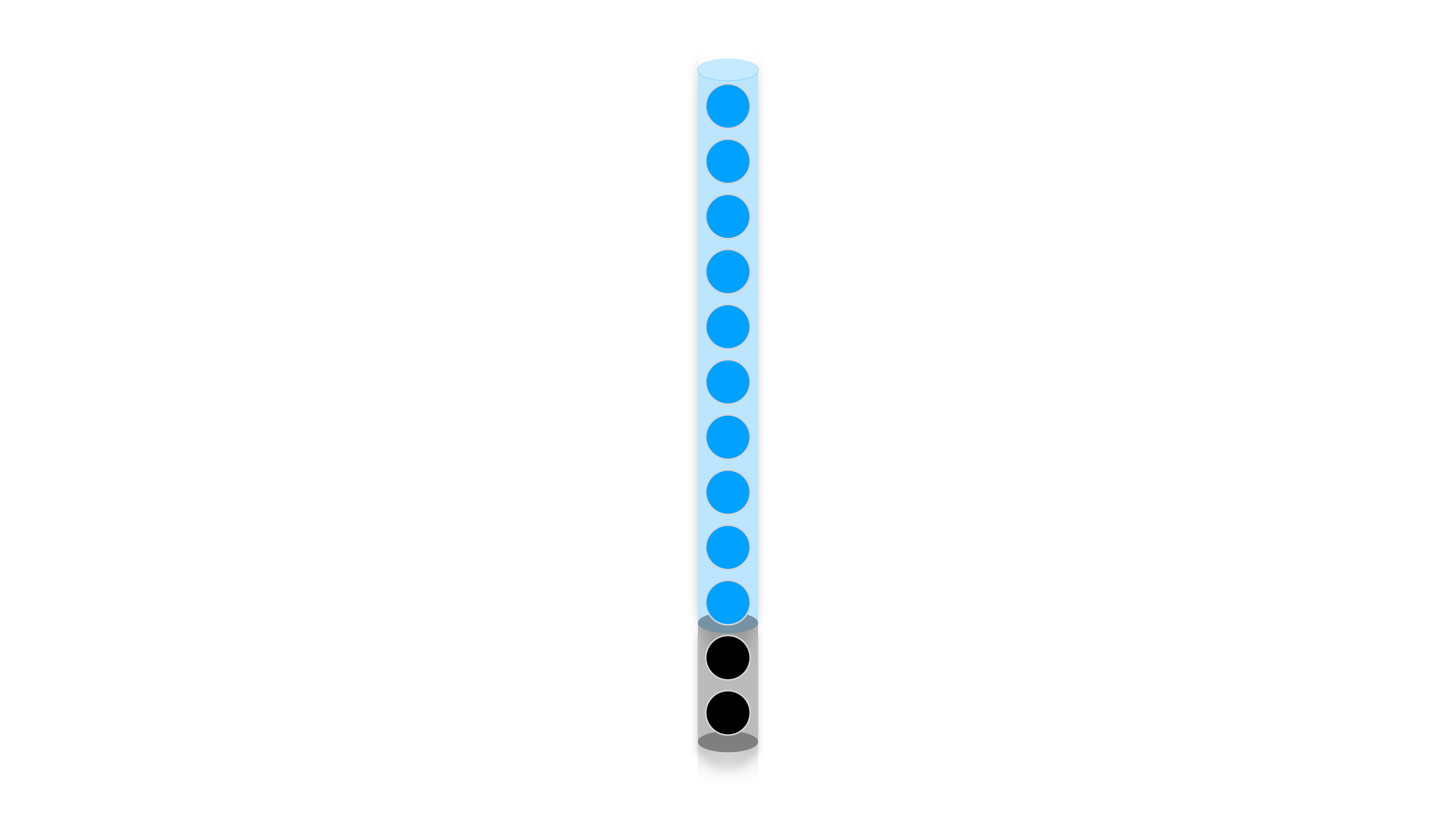}
	\hspace{4mm}
	\includegraphics[height=\linewidth]{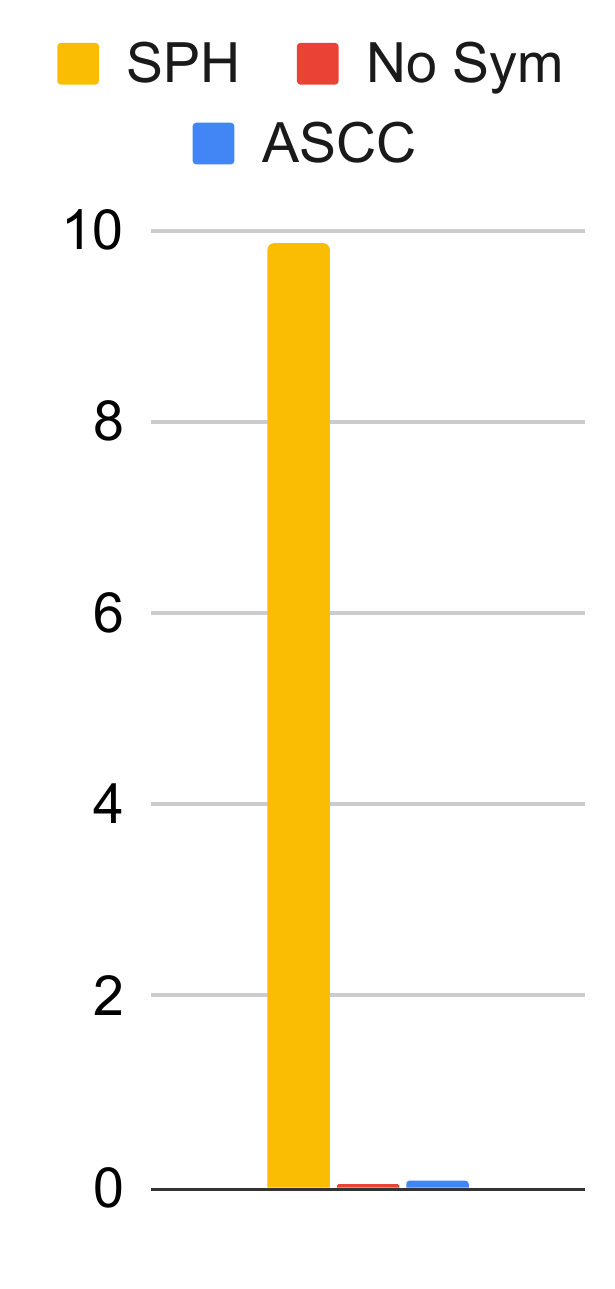}
	\vspace{-2mm}
	\subcaption*{\tiny{Column}}
    \end{subfigure}
	\hspace{-2mm}
	\vline
    \begin{subfigure}[c]{0.49\linewidth}
	\hspace{8mm}
	\includegraphics[height=\linewidth, trim= 1000 0 1000 0, clip]{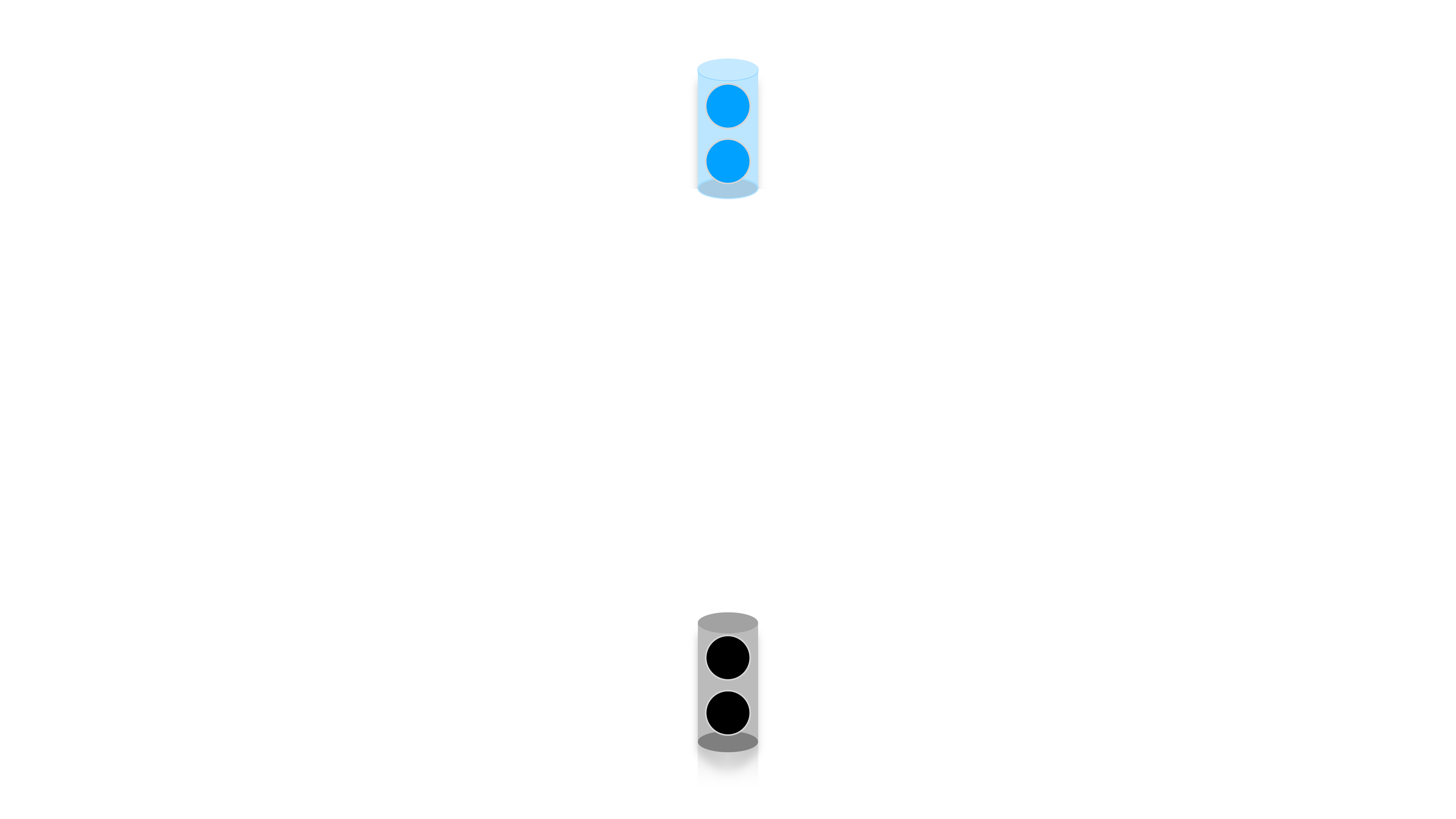}
	\hspace{2mm}
	\includegraphics[height=\linewidth]{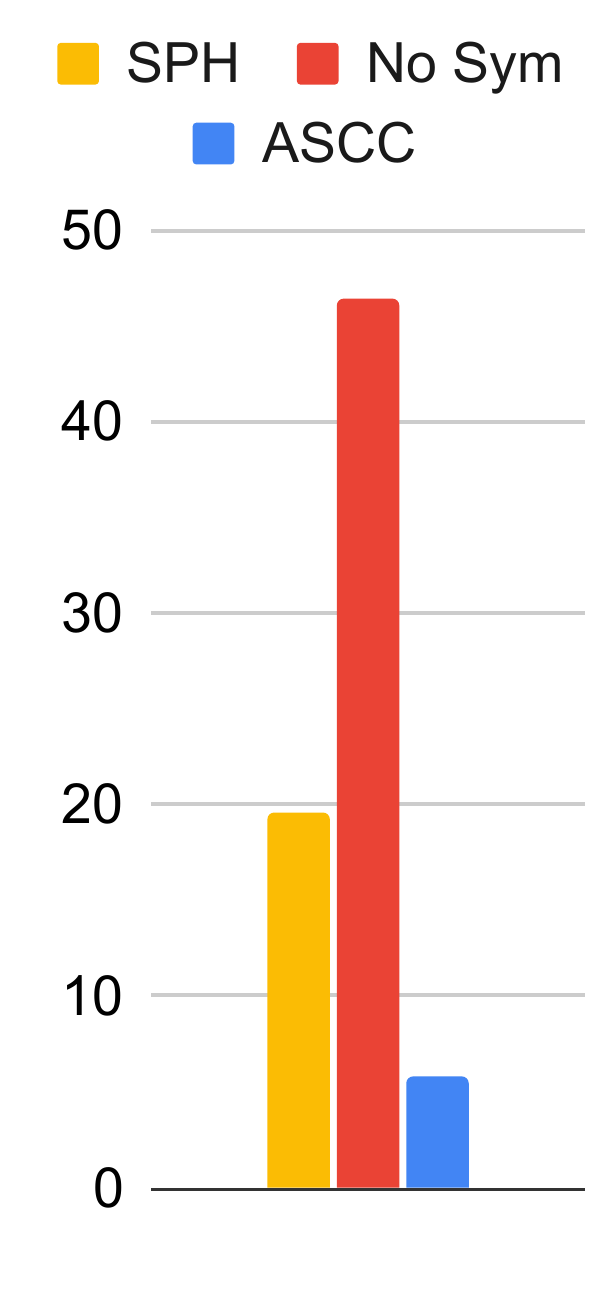}
	\hspace{-8mm}
	\vspace{-2mm}
	\subcaption*{\tiny{Free Fall}}
    \end{subfigure}	\caption{
	    RMSE ($\times 10^{-3}$) evaluation over 100 steps for a set of \textit{Column} scenes and \textit{Free Fall} scenes. A schematic is shown on the left side of the graphs: fluid particles (blue) are affected by gravity; solid boundary particles are shown in black. The ASCC network shows the best performance in terms of generalizing to the free fall cases.
	}\vspace{-4mm}
	\label{fig:column}
\end{wrapfigure}
We first consider one-dimensional simulations of a hydrostatic column of liquid. The training data consists of a different number of fluid particles which, under the effect of gravity, rest on a fixed boundary at the bottom. 
The reference stems from a solver with implicit time stepping \cite{he2012local} that can handle
global effects instantaneously.
To test the generalization, we also evaluate the methods for a free fall scene. Here we generate a certain number of fluid particles in free fall, which substantially deviates from the behavior of the training data.

\begin{figure}[t]
	\centering
	\includegraphics[width=\linewidth]{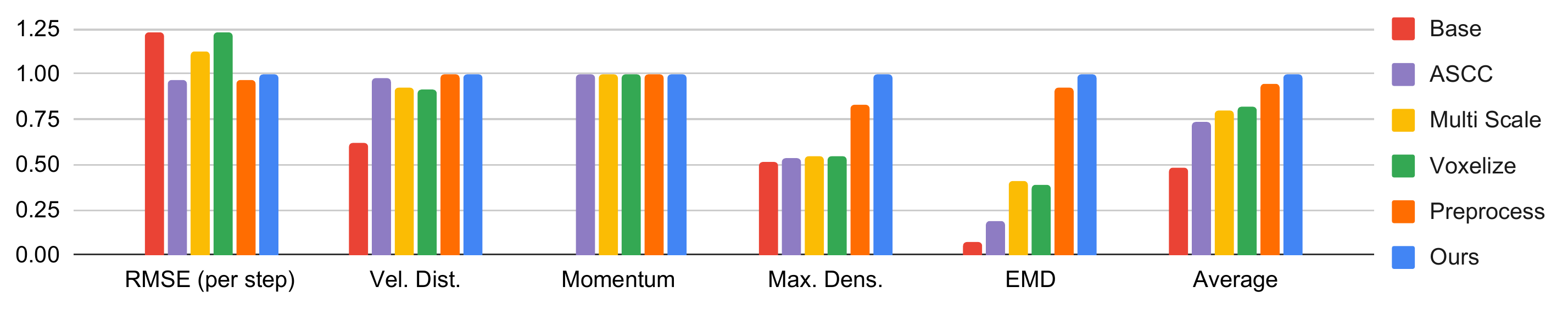}    \mySpaceTweak{}    \mySpaceTweak{}	\caption{
	    Quantitative evaluation for the 
	    ablation study: We consider a relative accuracy to the final method (Ours as 1.0) with higher values meaning better performance.
	    	    The metrics are computed as described in Sec.~\ref{sec:results}, with \textit{Vel. Dist.} denoting the JSD of the velocity distributions.
	    While \textit{RMSE (per step)} shows the RMSE after one time step, all other metrics are evaluated over a whole sequence. 
	    	    \textit{Average} corresponds to the mean of all previous metrics.
	}\vspace{-1mm}
	\label{fig:ablation}
\end{figure}
\mySpaceTweak{}
We trained two networks with the architecture from Sec.~\ref{sec:arch}, once with the proposed ASCC layer (ASCC) and once with a regular CConv layer instead (No Sym). 
Despite the seeming simplicity of the test case, even small inaccuracies can lead to unstable behavior over time.
Without the antisymmetric constraint, a network can simply overfit to the gravity and negate it. This reduces the problem to a local per particle problem. With ASCC, on the other hand, the network relies on particle interactions to solve the problem. This makes it necessary to determine a global context, and infer a pressure-like counter force to stabilize the fluid.
We compare the performance of both learned versions with the solution obtained by an explicit SPH solver with a small time step \cite{premvzoe2003particle}.
For new column heights, Fig.~\ref{fig:column} shows that both learned versions quickly compensate for compression effects in the liquid and yield very small errors, while the SPH solution oscillates in place.
Whereas SPH handles the free fall case on the right reasonably well with an error of $19.68$, the network without antisymmetry largely fails and produces erroneous motions with an error of $46.36$. The antisymmetric version, having learned a physically consistent response, yields an error that improves over SPH by a factor of $3.3$.

\paragraph{Ablation Study} 
\begin{figure}[t]
	\centering
	\begin{subfigure}[c]{0.72\textwidth}
	\begin{overpic}[width=0.195\textwidth, trim=440 0 1100 0, clip]{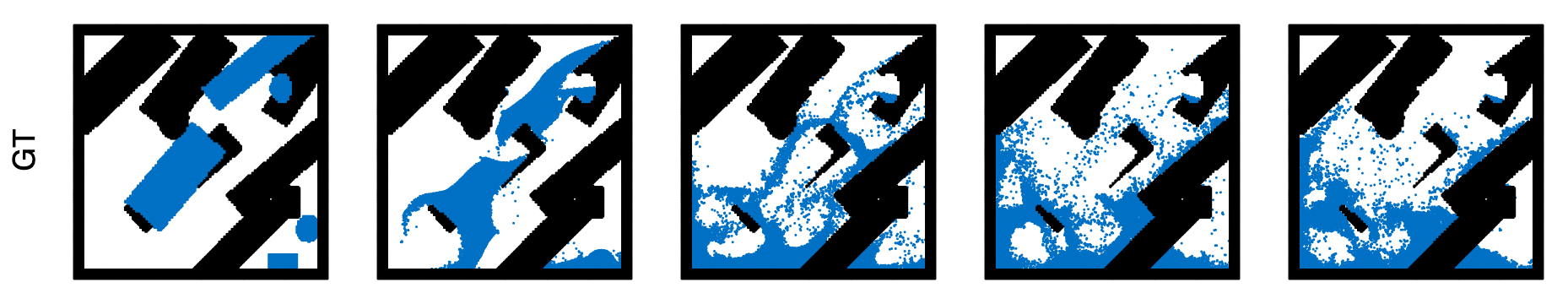}
	    \put(12,75){\tiny\color{white}{$GT$}}\end{overpic}	\begin{overpic}[width=0.195\textwidth, trim=440 0 1100 0, clip]{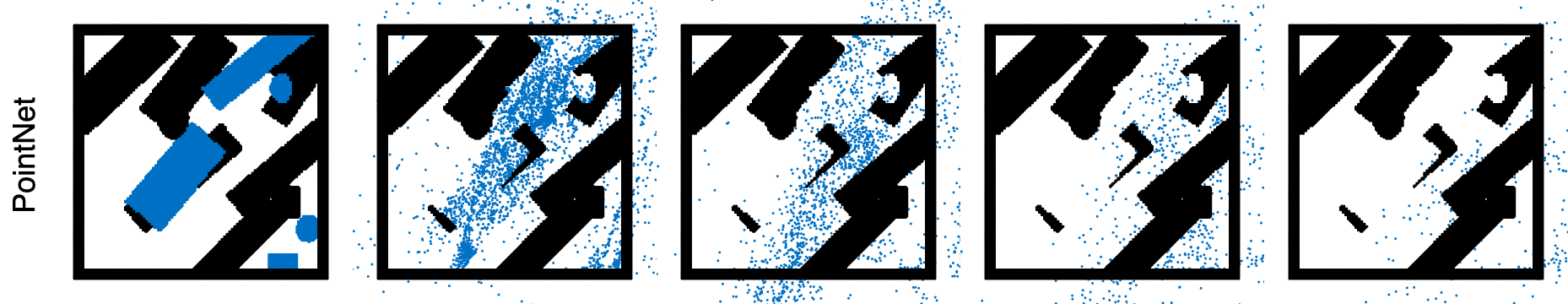}
	    \put(16,75){\color{white}{$a$}}\end{overpic}	\begin{overpic}[width=0.195\textwidth, trim=440 0 1100 0, clip]{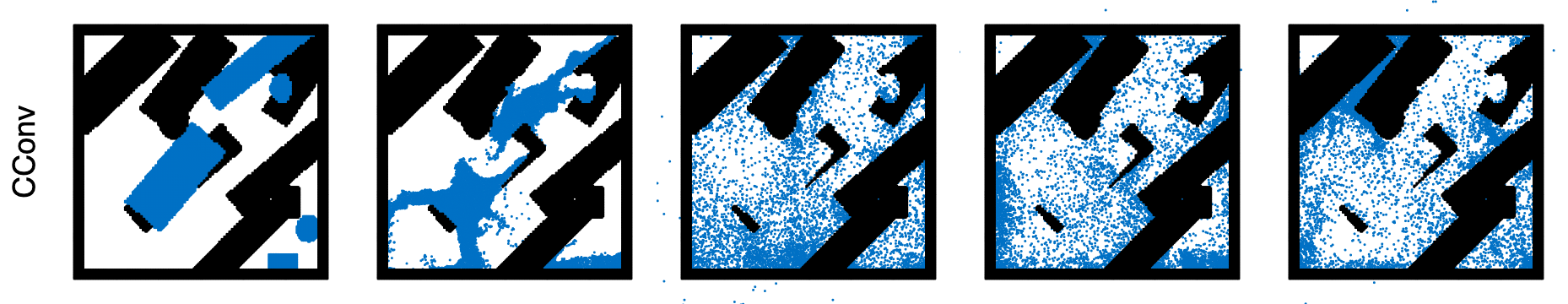}
	    \put(16,75){\color{white}{$b$}}\end{overpic}	\begin{overpic}[width=0.195\textwidth, trim=440 0 1100 0, clip]{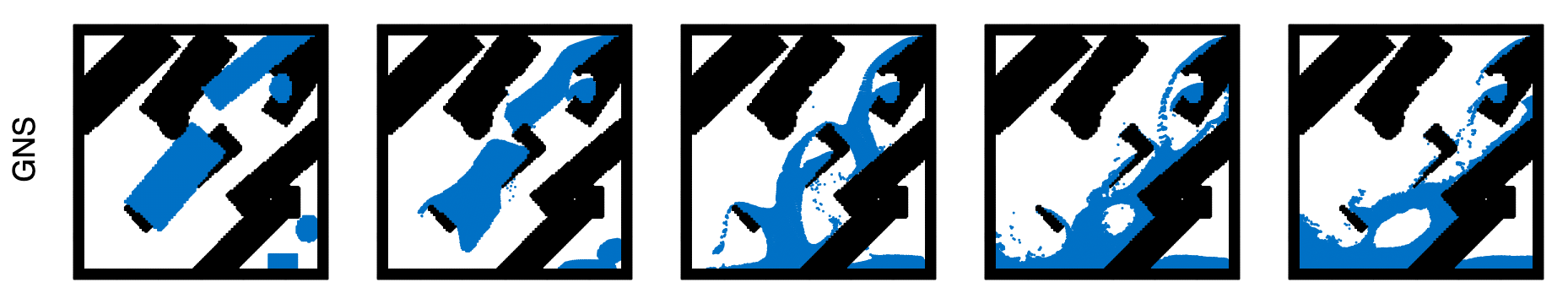}
	    \put(16,75){\color{white}{$c$}}\end{overpic}	\begin{overpic}[width=0.195\textwidth, trim=440 0 1100 0, clip]{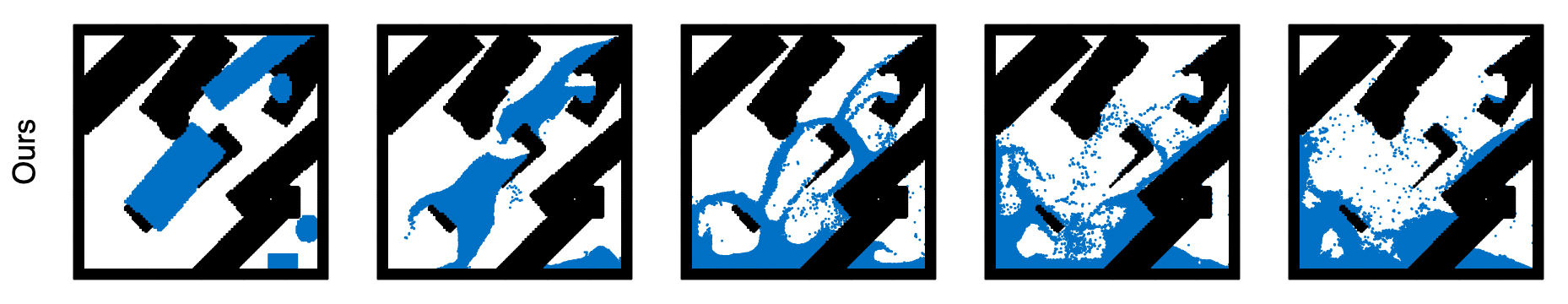}
	    \put(16,75){\color{white}{$d$}}\end{overpic}	\end{subfigure}
	\vline
	\hspace{1mm}
	\begin{subfigure}[c]{0.24\textwidth}
	\includegraphics[width=0.9\textwidth, trim= 10 260 0 0, clip]{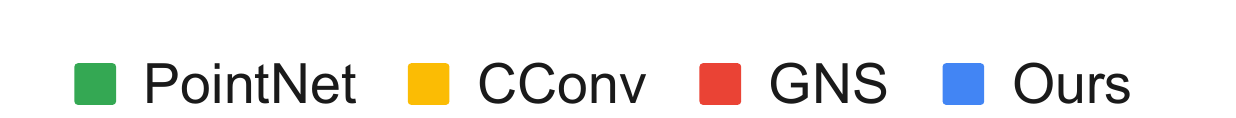}
	
	\includegraphics[width=0.45\textwidth, trim= 0 0 0 40, clip]{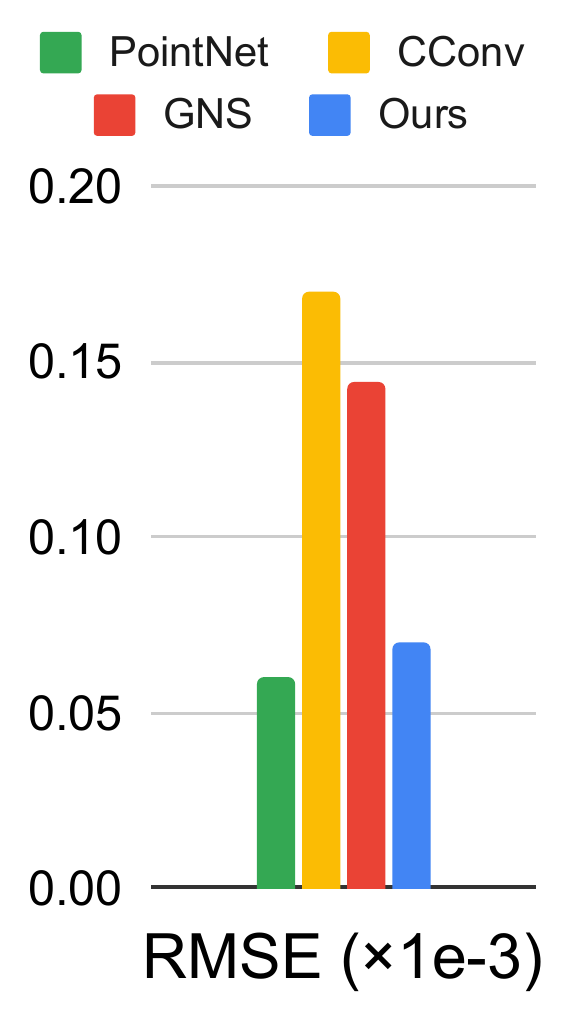}
	\includegraphics[width=0.45\textwidth, trim= 0 0 0 300, clip]{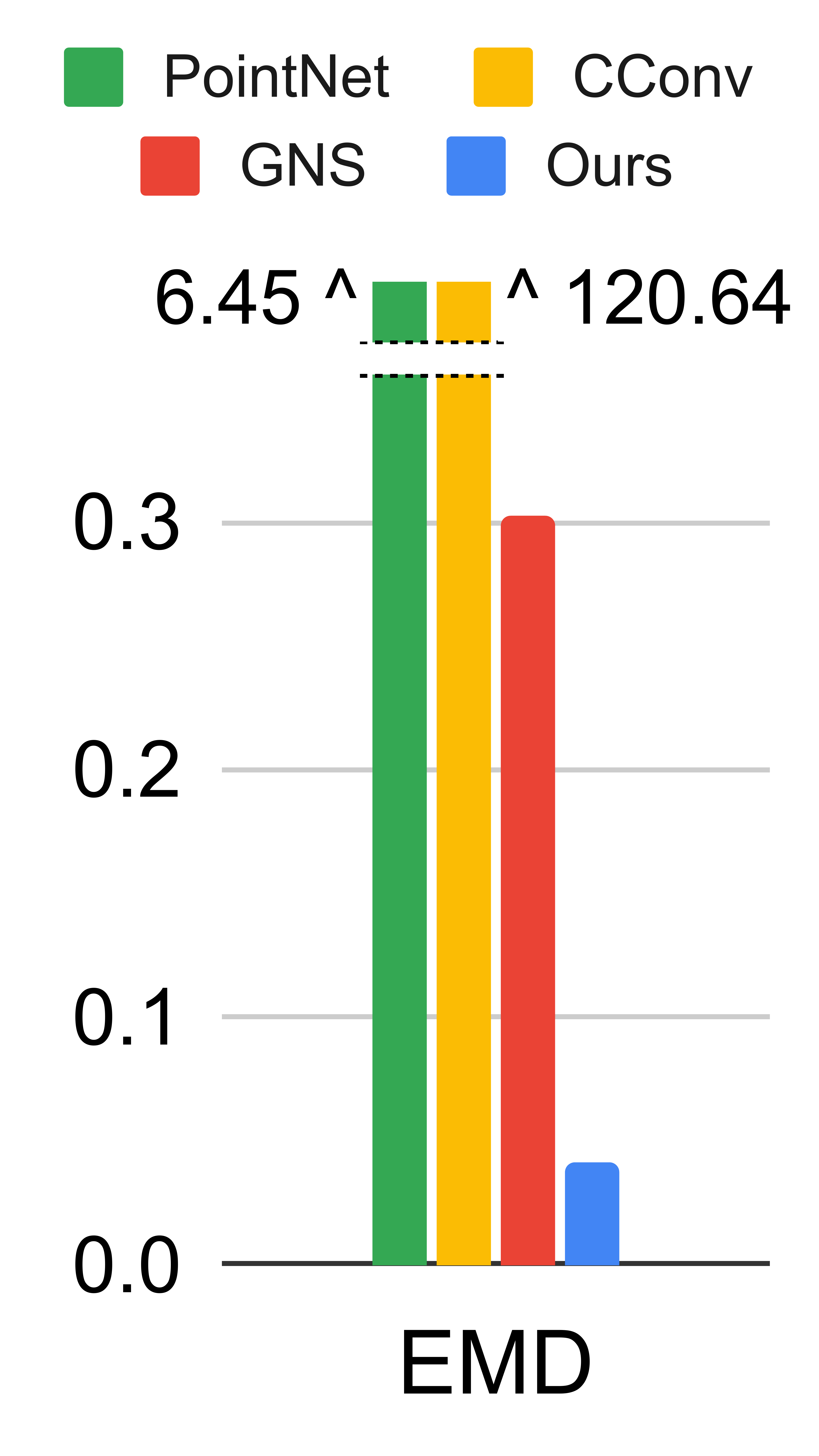}
	\end{subfigure}    
	\caption{Excerpts from a test sequence with the high-fidelity \texttt{WBC-SPH} dataset and a quantitative evaluation on the right. The frames on the left represent, f.l.t.r.,  ($GT$) the ground-truth, ($a$) PointNet, ($b$) CConv, ($c$) GNS, and ($d$) Ours.}
	\mySpaceTweak{}    
	\mySpaceTweak{}	
	\label{fig:fluid_eval}
\end{figure}
We evaluate the relevance of individual features introduced for our method with the ablation study using the \texttt{WBC-SPH} dataset. 
We start with a base model (\textit{Base}) without an ASCC layer and gradually add relevant features. While this model fares very well for single steps,
as shown in Fig.~\ref{fig:ablation} it performs poorly with respect to all other evaluations. Here, the graphs show the performance relative to the full method, i.e., $L_{\text{Ours}}/L$, with $L$ denoting the loss of a variant under consideration.
For the second version, we add the antisymmetrical layer (\textit{ASCC}), which ensures the conservation of momentum and significantly improves the motion in terms of velocity distribution.
Here the momentum score of $0.0$ for \textit{Base} corresponds to a effective error of $0.095$, all following variants having a score of $1.0$ with zero error.
Next, we evaluate the relevance of the multi-scale processing. We tested two different variants, one with FPS (\textit{Multi Scale}) and one with the voxelization (\textit{Voxelize}). The performance is very similar for both, with the latter variant having clear advantages in terms of resource usage.
While improvements following the multi-scale handling are not obvious first it starts to more clearly pay off when combined with preprocessing for temporal coherence training in
(\textit{Preprocess}). The improved long-term stability and accuracy are clearly visible
in the maximum density and EMD metrics. 
Lastly, our full method (\textit{Ours}) includes additional input rotation normalization and a more advanced boundary processing step, as explained in more detail in App. \ref{sec:impl_details}. This leads to further slight improvements and yields the final algorithm, which we will compare to previous work in the following sections.

\paragraph{Comparisons with Previous Work} 
In the following, we compare our method with several established methods as baselines: 
an adapted variant of PointNet \cite{pointnet}, the CConv network \cite{ummenhofer2019lagrangian} and GNS \cite{sanchez2020learning}. More information about the employed networks and the training procedures are given in App. \ref{sec:train_details}.
We train and evaluate networks with our \texttt{WBC-SPH} dataset.
For fairness, we additionally evaluate the different approaches with their respective data sets, i.e., with the \texttt{WaterRamps} dataset of GNS and with the \texttt{Liquid3d} dataset of CConv.

When trained on the \texttt{WBC-SPH} dataset, which consists of sequences with $3200$ time steps,
neither CConv nor PointNet are able to generate stable results. While the short-term RMSE accuracy is good, both networks fail to stabilize the challenging dynamics of our test sequences. Examples are shown in Fig.~\ref{fig:fluid_eval}.
The outputs from GNS are stable, but the network seems to maintain stability by smoothing and damping the dynamics. Also, despite receiving multiple frames over time, the GNS model has difficulty predicting the dynamics of changing gravity directions.
For this scenario, the antisymmetric ASCC network clearly outperforms GNS and the other variants in quantitative as well as qualitative terms. In particular, its EMD error is $7.22$ times smaller than the closest competitor.
\begin{figure}[t] 
	\centering
        \hspace{-2mm}
	\begin{overpic}[width=0.07\textwidth, trim=110 51 1493 51, clip]{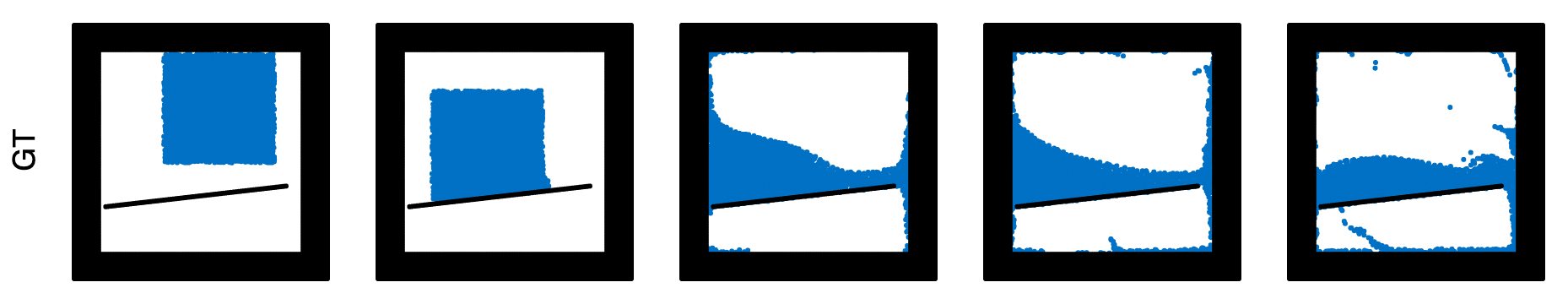}
	    \put(1,108){\tiny \color{black}{GT:}}
	    \put(85,-18){\tiny \color{black}{$0$}}\end{overpic}	\hspace{0.2mm}
	\begin{overpic}[width=0.07\textwidth, trim=830 51 773 51, clip]{images/gns/gt/16.png}
	    \put(70,-18){\tiny \color{black}{$160$}}\end{overpic}	\hspace{0.2mm}
	\begin{overpic}[width=0.07\textwidth, trim=1550 51 53 51, clip]{images/gns/gt/16.png}
	    \put(70,-18){\tiny \color{black}{$320$}}\end{overpic}	\hspace{2mm}
    	\begin{overpic}[width=0.07\textwidth, trim=110 51 1493 51, clip]{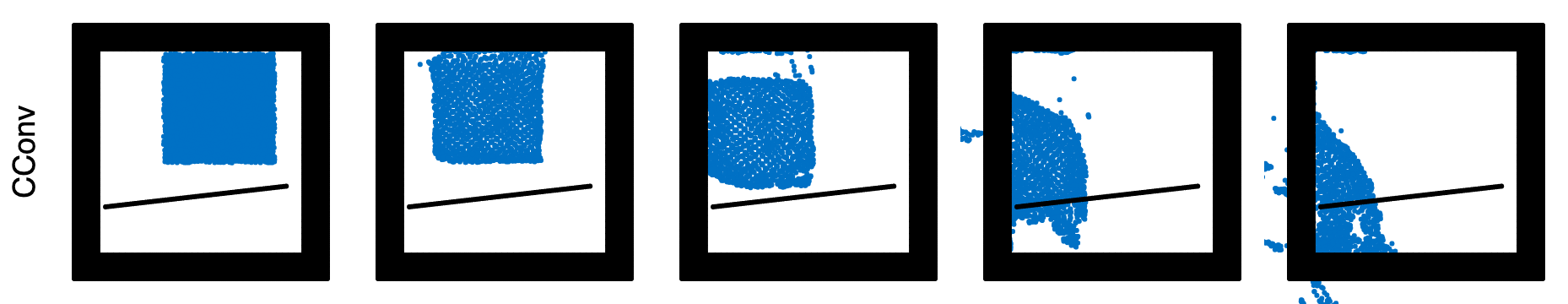}
	    \put(1,108){\tiny \color{black}{CConv:}}
	    \put(85,-18){\tiny \color{black}{$0$}}\end{overpic}	\hspace{0.2mm}
	\begin{overpic}[width=0.07\textwidth, trim=830 51 773 51, clip]{images/gns/cconv/16.png}
	    \put(70,-18){\tiny \color{black}{$160$}}\end{overpic}	\hspace{0.2mm}
	\begin{overpic}[width=0.07\textwidth, trim=1550 51 53 51, clip]{images/gns/cconv/16.png}
	    \put(70,-18){\tiny \color{black}{$320$}}\end{overpic}	\hspace{2mm}
    	\begin{overpic}[width=0.07\textwidth, trim=110 51 1493 51, clip]{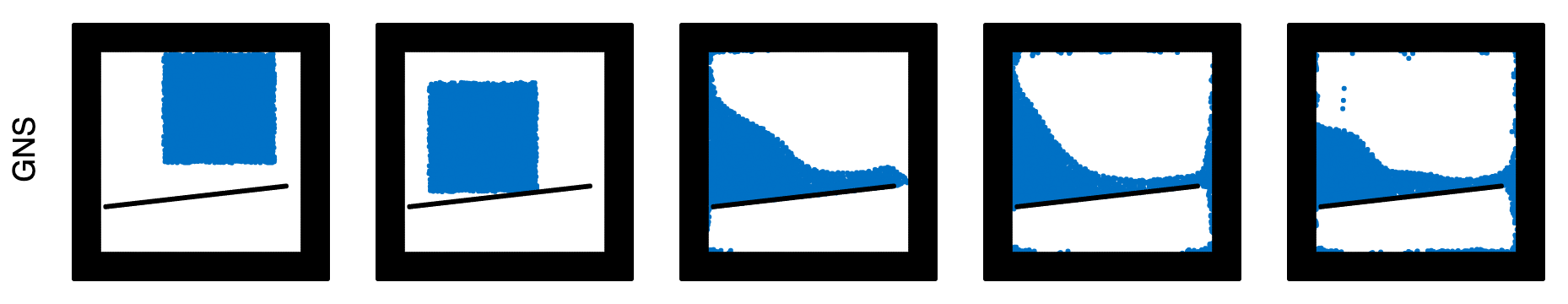}
	    \put(1,108){\tiny \color{black}{GNS:}}
	    \put(85,-18){\tiny \color{black}{$0$}}\end{overpic}	\hspace{0.2mm}
	\begin{overpic}[width=0.07\textwidth, trim=830 51 773 51, clip]{images/gns/gns/16.png}
	    \put(70,-18){\tiny \color{black}{$160$}}\end{overpic}	\hspace{0.2mm}
	\begin{overpic}[width=0.07\textwidth, trim=1550 51 53 51, clip]{images/gns/gns/16.png}
	    \put(70,-18){\tiny \color{black}{$320$}}\end{overpic}	\hspace{2mm}
    	\begin{overpic}[width=0.07\textwidth, trim=110 51 1493 51, clip]{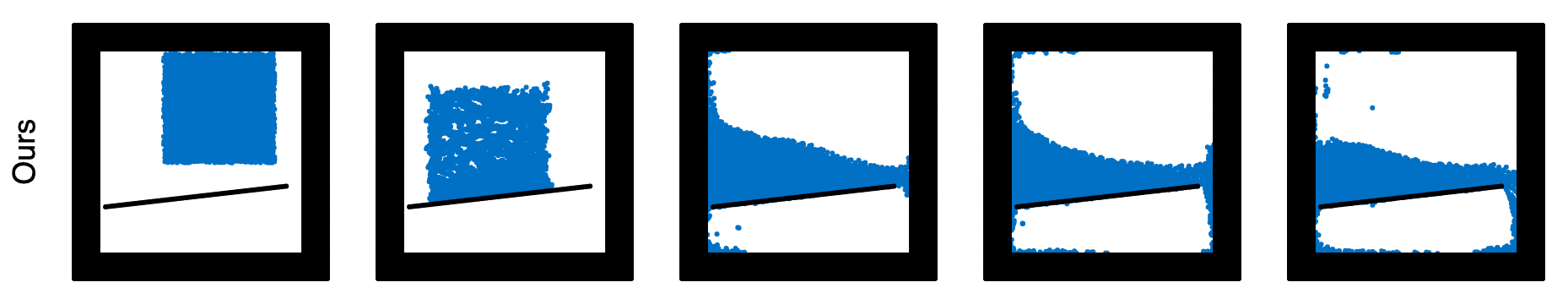}
	    \put(1,108){\tiny \color{black}{Ours:}}
	    \put(85,-18){\tiny \color{black}{$0$}}\end{overpic}	\hspace{0.2mm}
	\begin{overpic}[width=0.07\textwidth, trim=830 51 773 51, clip]{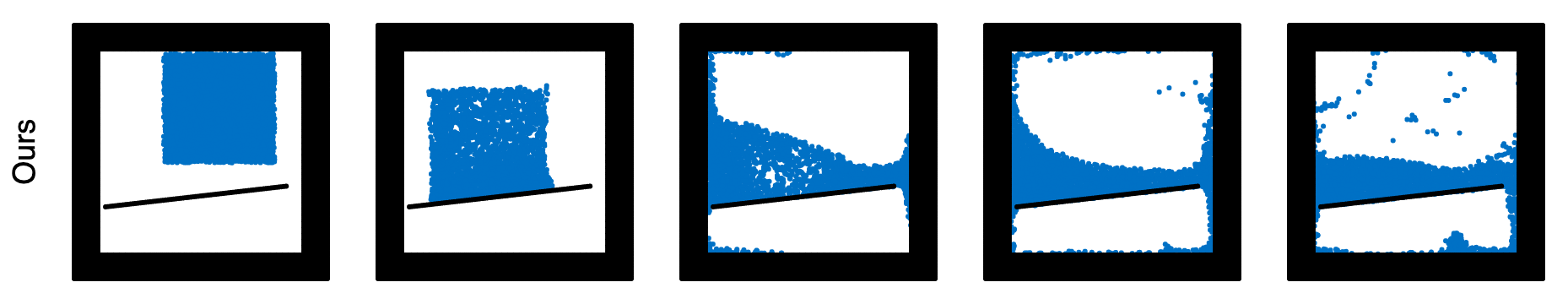}
	    \put(70,-18){\tiny \color{black}{$160$}}\end{overpic}	\hspace{0.2mm}
	\begin{overpic}[width=0.07\textwidth, trim=1550 51 53 51, clip]{images/gns/ours_long/16.png}
	    \put(70,-18){\tiny \color{black}{$320$}}\end{overpic}    \hspace{-2mm}
	\caption{Snapshots over time from models for the \texttt{WaterRamps} dataset.}
		\label{fig:gns_eval}
\end{figure}
\begin{wrapfigure}{t}{0.46\linewidth} 
	\includegraphics[width=0.25\textwidth, trim= 0 300 0 0, clip]{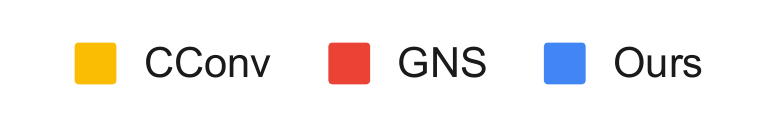} 
	
	\includegraphics[width=0.11\textwidth, trim= 0 0 0 40, clip]{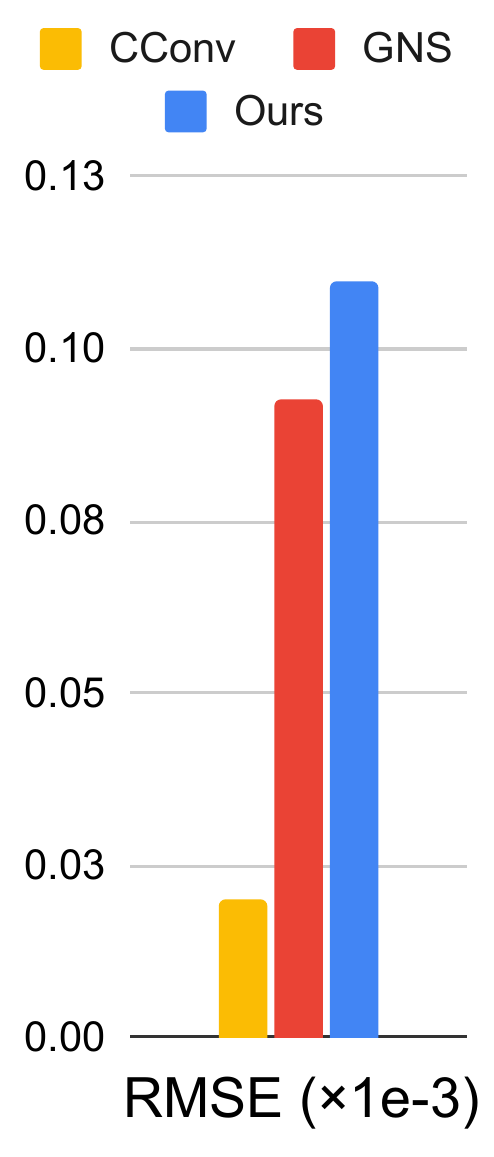} 
				\includegraphics[width=0.11\textwidth, trim= 0 0 0 40, clip]{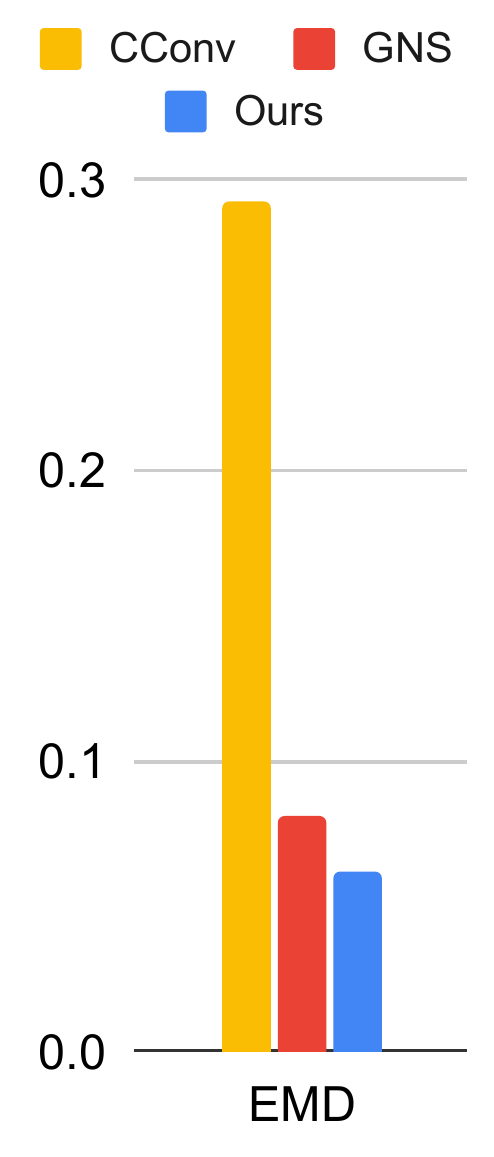} 
	\includegraphics[width=0.11\textwidth, trim= 0 0 0 40, clip]{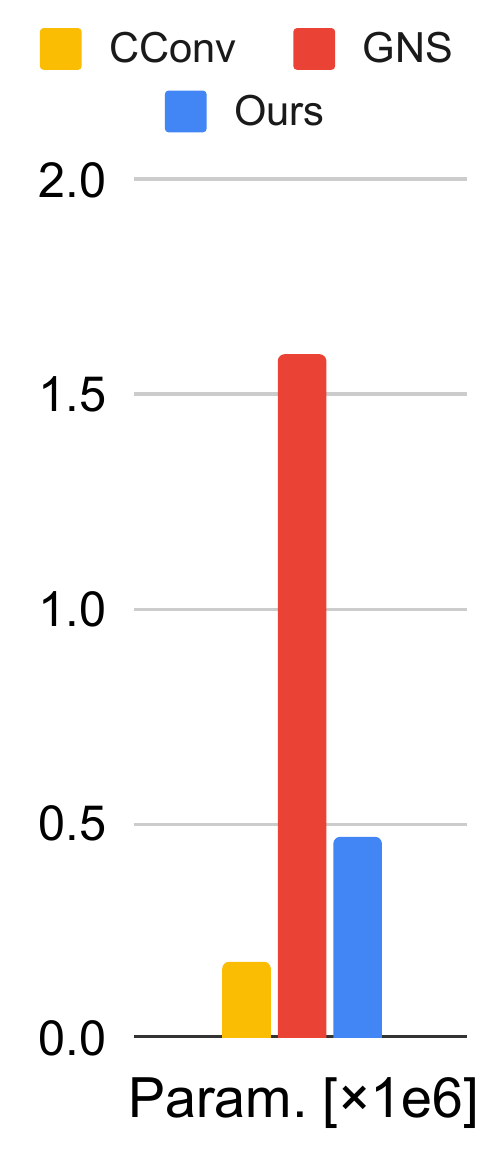}  
	    \includegraphics[width=0.11\textwidth, trim= 0 0 0 40, clip]{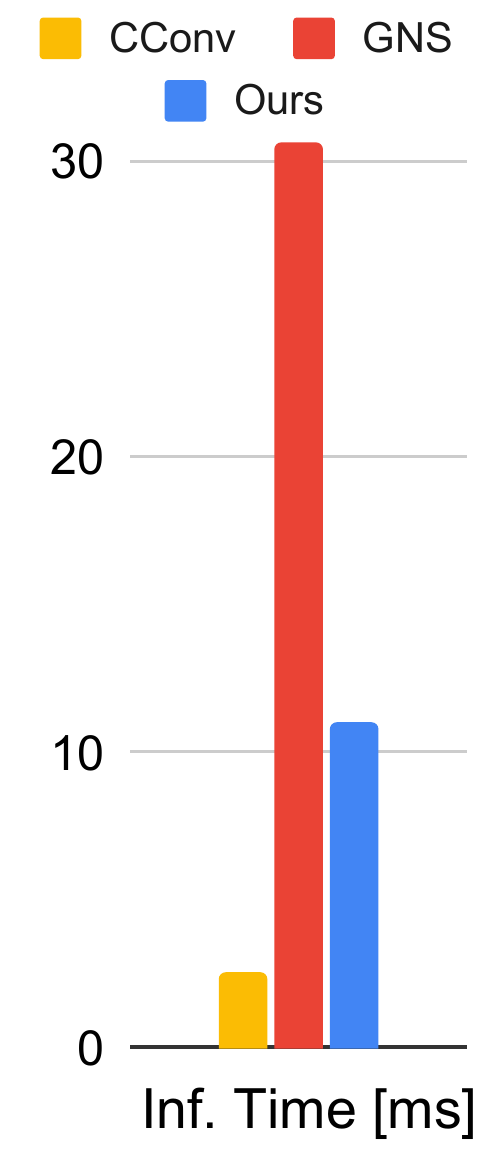}
    \mySpaceTweak{}	\caption{Quantitative evaluations with the \texttt{WaterRamps} dataset. We show the accuracy of the models along with their size and computational performance.}
		\label{fig:gns_quan}
	\vspace{-2mm}
\end{wrapfigure}
For fairness of the comparison, we also retrain our model and the CConv model with the \texttt{WaterRamps} dataset.
While the accuracy of GNS with this data is very high, our method nonetheless outperforms GNS with a decrease of $24\%$ in terms of EMD. In comparison, CConv results in a $3.61$ times larger EMD.
Over short periods the dynamics of GNS behave similarly to ours.
For longer periods of more than 30 frames, the results diverge, as shown in Fig.~\ref{fig:gns_eval}.
Most importantly, our network yields a high accuracy while requiring only a fraction of the computational resources that GNS requires (Fig.~\ref{fig:gns_quan}):
the ASCC network has $29\%$ of the weights of GNS, and inference is $2.79$ times faster, while yielding the aforementioned improvements in terms of accuracy.  
In addition to the conservation of  momentum, this advantage can be explained by the inductive biases of our 
convolutional architecture: our networks can efficiently process positional information, which is difficult for generic graph networks. We discuss the inherent advantages of the convolution architectures over graph networks in more detail in appendix~\ref{app:discuss}. 
While CConv performs best in terms of resources, it does so at the expense of inference accuracy: its EMD is three times larger than the one obtained with our method.

\mySpaceTweak{}
\paragraph{Generalization} 
\begin{figure}[t]
	\centering
    \begin{subfigure}[c]{0.1\textwidth}
	\includegraphics[width=\textwidth]{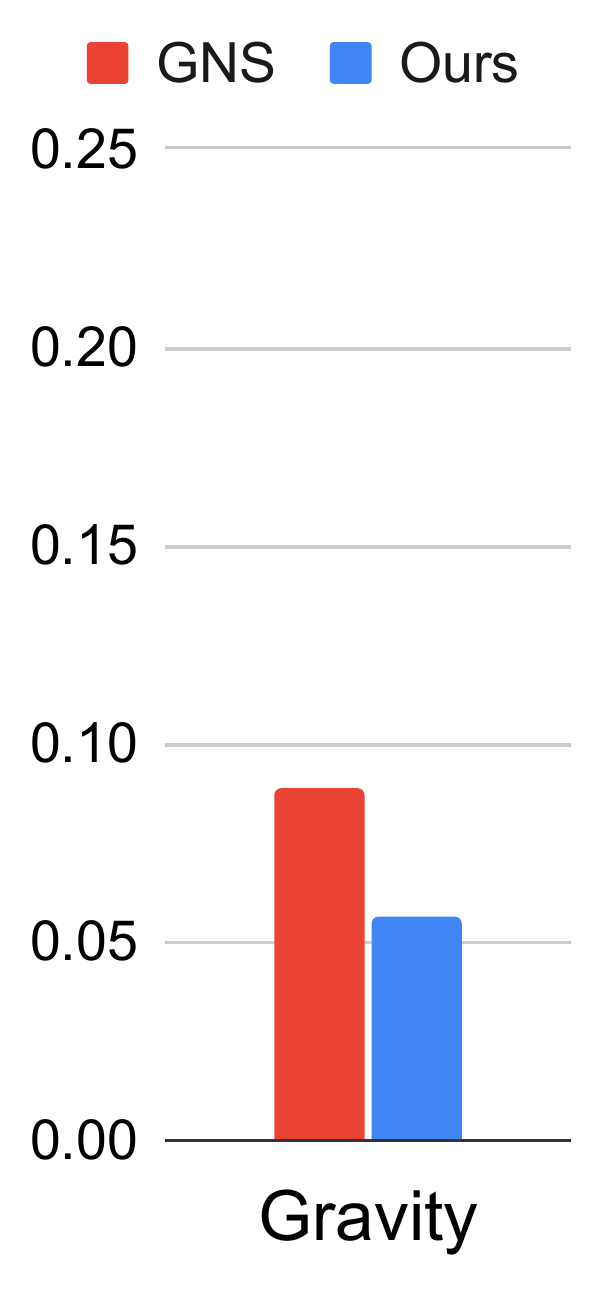} 
	\end{subfigure}
    \begin{subfigure}[c]{0.38\textwidth}
	\includegraphics[width=\textwidth]{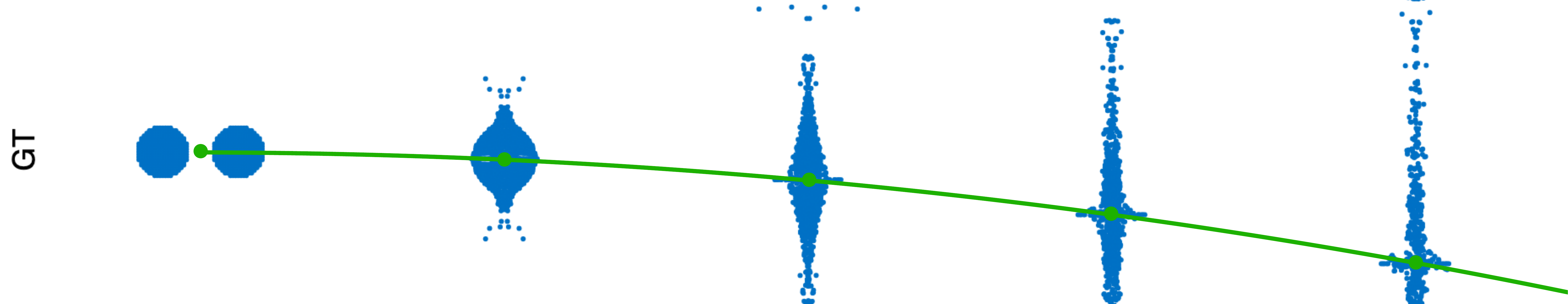} 
	\includegraphics[width=\textwidth]{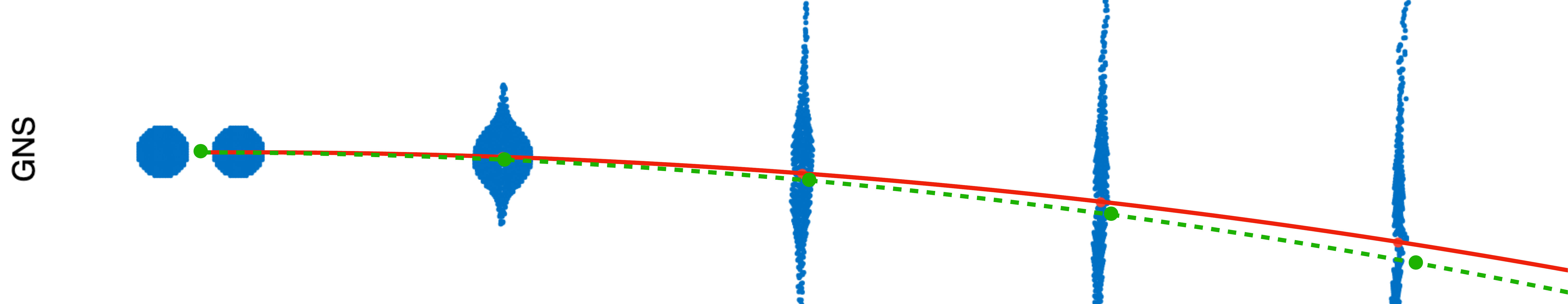} 
	\includegraphics[width=\textwidth]{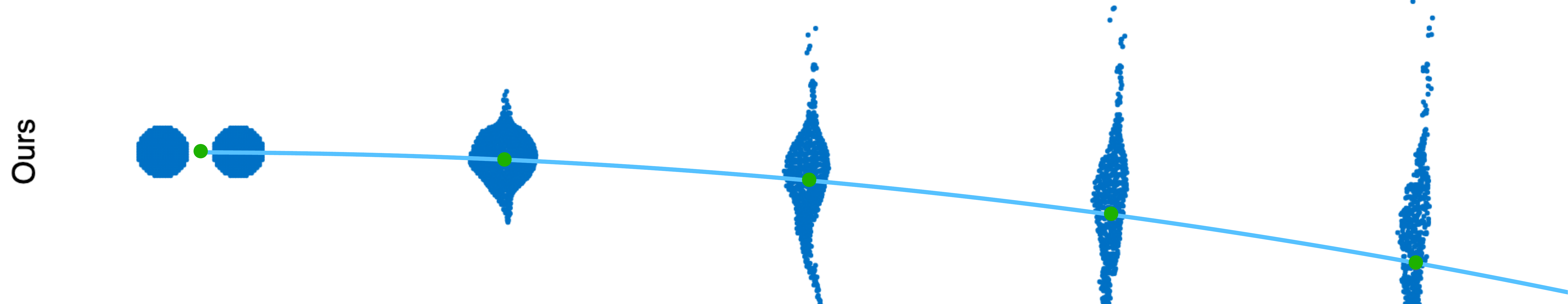} 
	\end{subfigure}
	\vspace{1mm}
	\vline
	\vspace{1mm}
    \begin{subfigure}[c]{0.1\textwidth}
	\includegraphics[width=\textwidth]{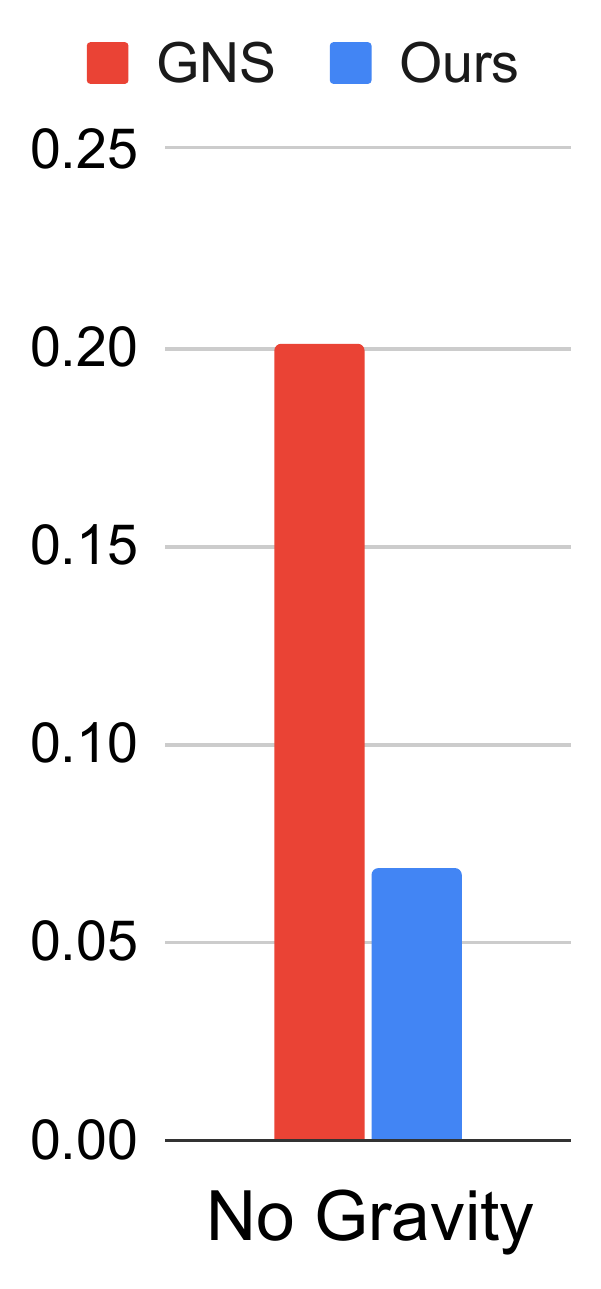} 
	\end{subfigure}
    \begin{subfigure}[c]{0.38\textwidth}
	\includegraphics[width=\textwidth]{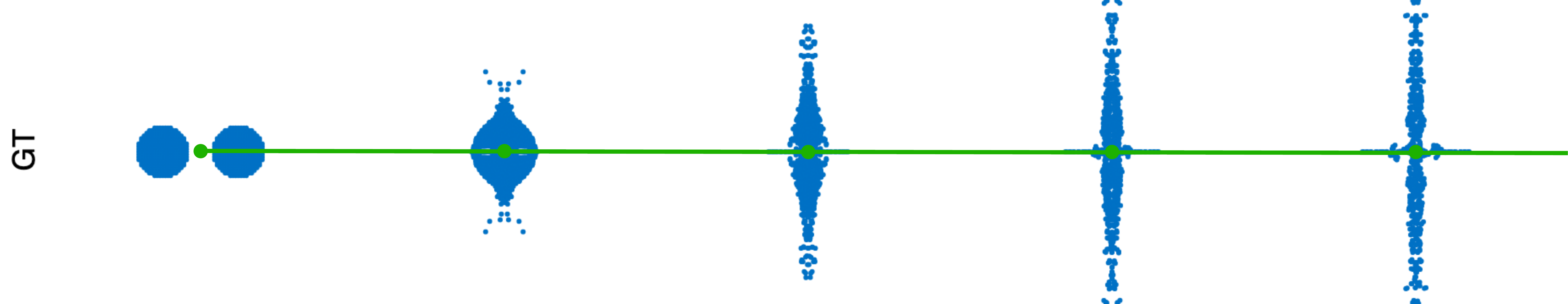} 
	\includegraphics[width=\textwidth]{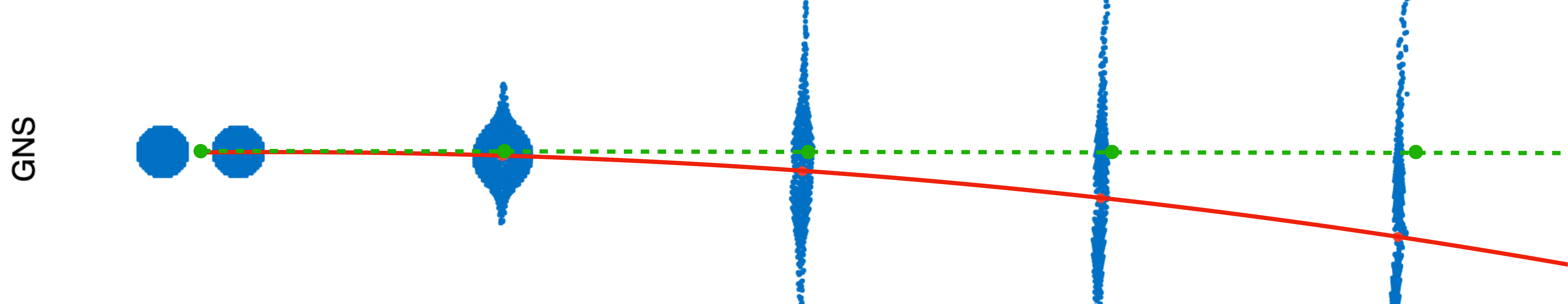} 
	\includegraphics[width=\textwidth]{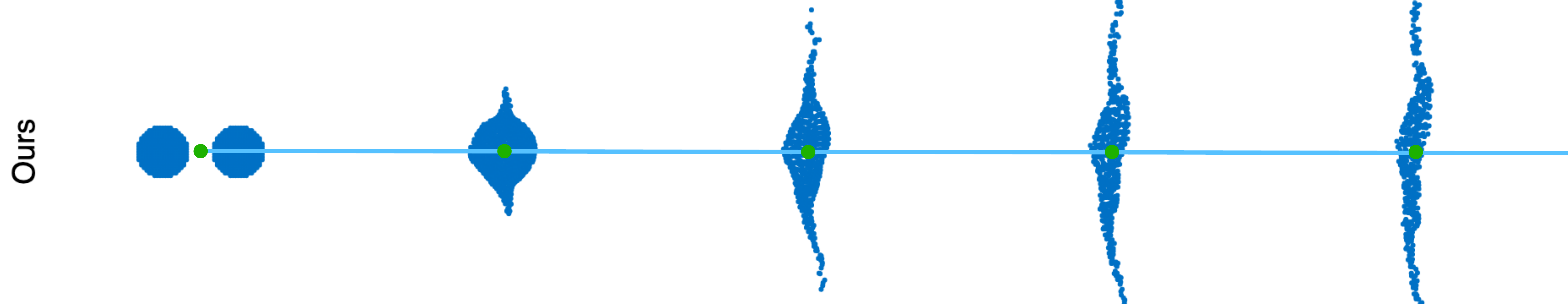} 
	\end{subfigure}
    \mySpaceTweak{}    \mySpaceTweak{}	\caption{
	Assessment of generalization with droplet collisions: left with, right without gravity. Each side shows an EMD evaluation and three centers of mass trajectories with overlaid snapshots over time. The reference trajectories are shown in green. While our approach closely matches these trajectories, hiding them behind the blue lines, GNS noticeably deviates, yielding larger errors in terms of EMD.
		}
	\label{fig:mom}
\end{figure}
To assess the generalizing capabilities of our network and GNS trained with the \texttt{WaterRamps} data, we perform tests with droplets colliding under the influence of varying gravity.
This case allows for an accurate evaluation in terms of the center of mass trajectory of the liquid volume. Considering a case with default gravity, the ASCC network 
already yields an EMD error reduced by $57\%$. 
When reducing the gravity to zero, the GNS trajectory noticeably deviates from the ground truth trajectory and yields a $195\%$ larger error than our method. The GNS has encoded the influence of gravity in its learned representation and hence has difficulties transferring the dynamics to the changed physics environment.

\begin{figure}[t]
	\centering
	\begin{subfigure}[c]{0.7\linewidth}
	\begin{subfigure}[c]{0.105\linewidth}
	\includegraphics[width=\linewidth, trim={2.cm 1.cm 3.cm 6.cm},clip]{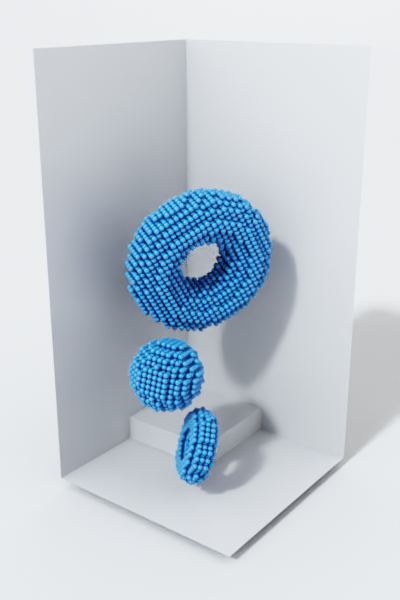}
	\includegraphics[width=\linewidth, trim={2.cm 1.cm 3.cm 6.cm},clip]{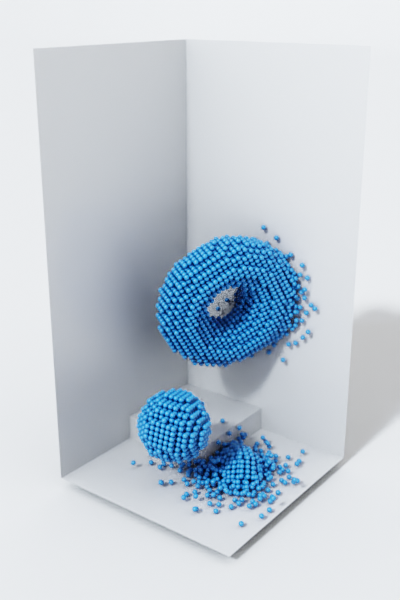}
	\end{subfigure}
	\begin{subfigure}[c]{0.21\linewidth}
	\begin{overpic}[width=\linewidth, trim={2.cm 1.cm 3.cm 6.cm},clip]{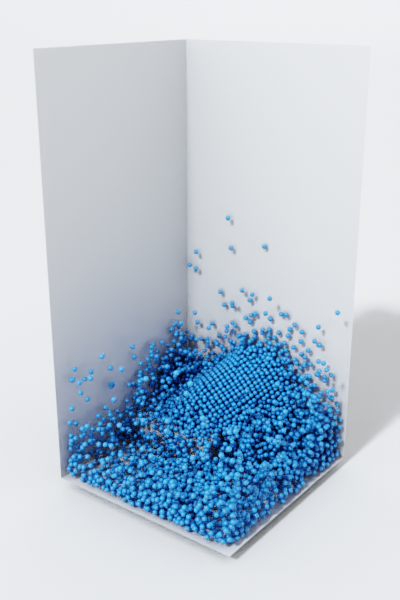}
	    \put(5,90){\small \color{white}{GT}}\end{overpic}
	\end{subfigure}
	\begin{subfigure}[c]{0.105\linewidth}
	\includegraphics[width=\linewidth, trim={2.cm 1.cm 3.cm 6.cm},clip]{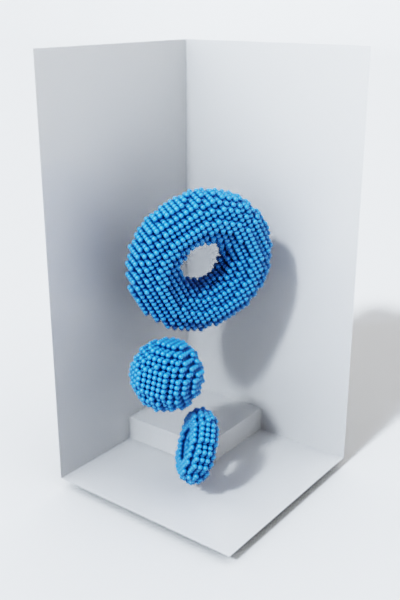}
	\includegraphics[width=\linewidth, trim={2.cm 1.cm 3.cm 6.cm},clip]{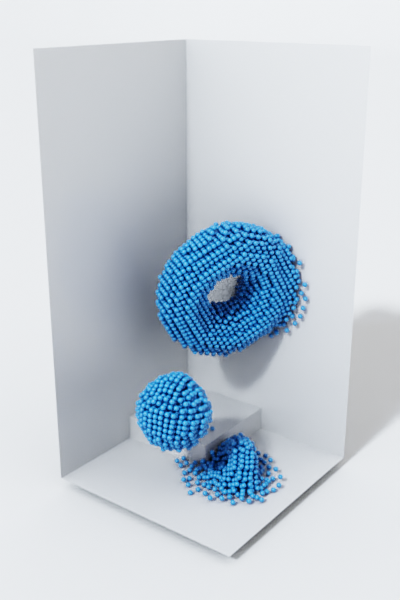}
	\end{subfigure}
	\begin{subfigure}[c]{0.21\linewidth}
	\begin{overpic}[width=\linewidth, trim={2.cm 1.cm 3.cm 6.cm},clip]{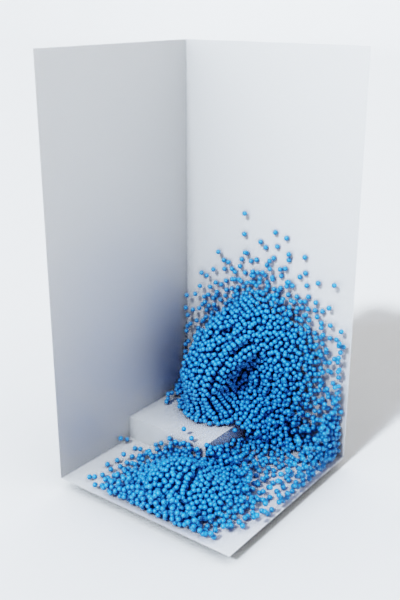}
	    \put(5,90){\small \color{white}{CConv}}\end{overpic}
	\end{subfigure}
	\begin{subfigure}[c]{0.105\linewidth}
	\includegraphics[width=\linewidth, trim={2.cm 1.cm 3.cm 6.cm},clip]{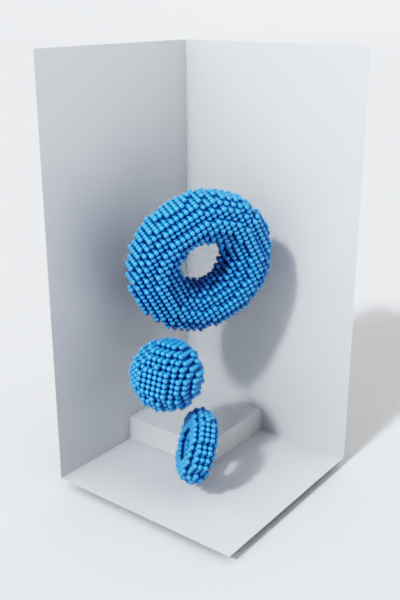}
	\includegraphics[width=\linewidth, trim={2.cm 1.cm 3.cm 6.cm},clip]{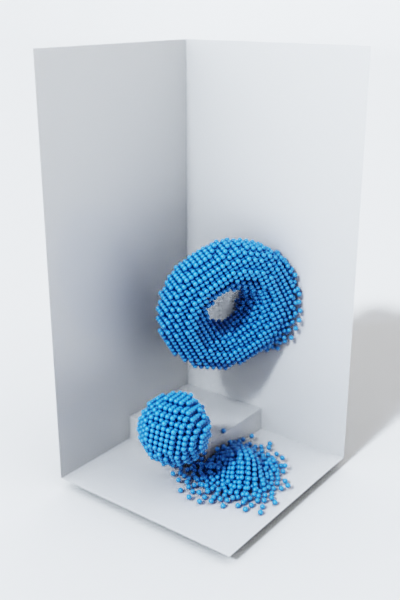}
	\end{subfigure}
	\begin{subfigure}[c]{0.21\linewidth}
	\begin{overpic}[width=\linewidth, trim={2.cm 1.cm 3.cm 6.cm},clip]{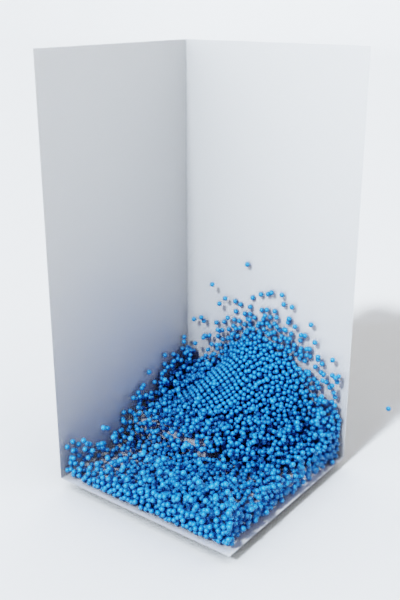}
	    \put(5,90){\small \color{white}{Ours}}\end{overpic}
	\end{subfigure}
	\end{subfigure}
		\begin{subfigure}[c]{0.285\linewidth}
	\includegraphics[width=0.48\textwidth, trim= 0 0 0 25, clip]{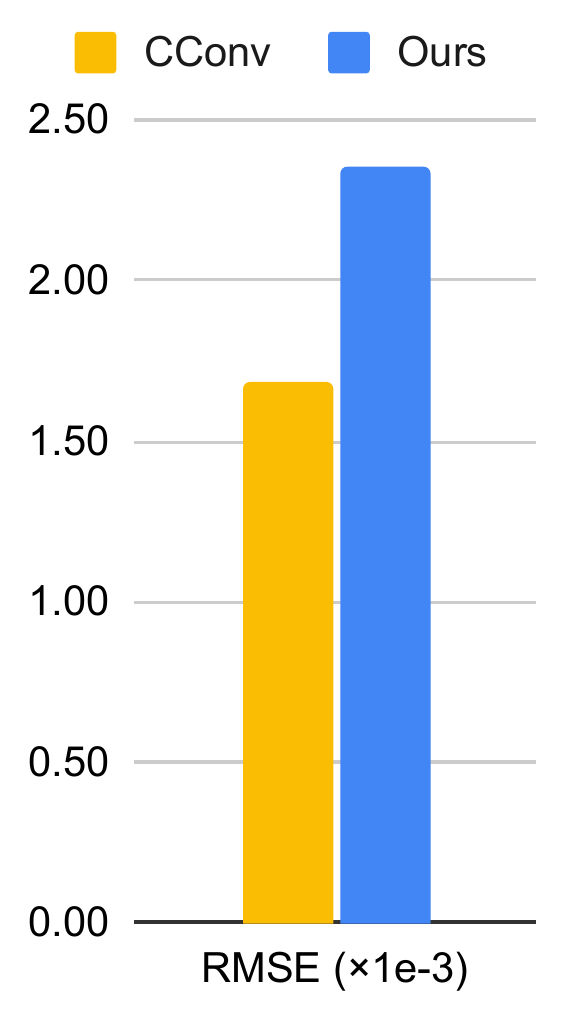} 
	\includegraphics[width=0.48\textwidth]{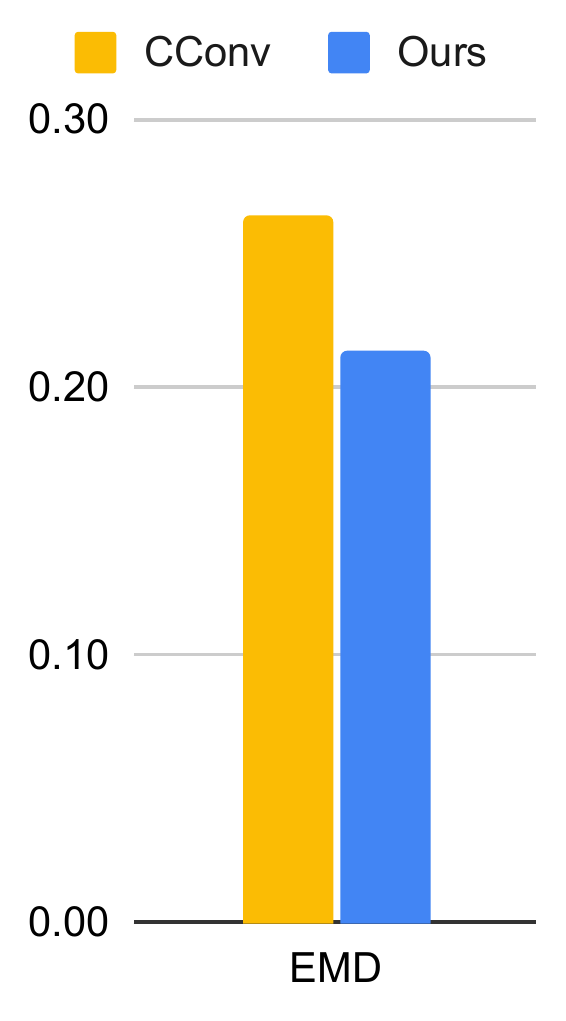} 
	\end{subfigure}
    \mySpaceTweak{}	\caption{Example of a 3D sequence based on the \texttt{Liquid3d} dataset. CConv visibly dampens the acceleration by gravity while our method closely matches the dynamics of the ground truth. 		}
	\label{fig:3d_qual}
\end{figure}

\begin{figure}[t]
\centering
\includegraphics[width=\linewidth]{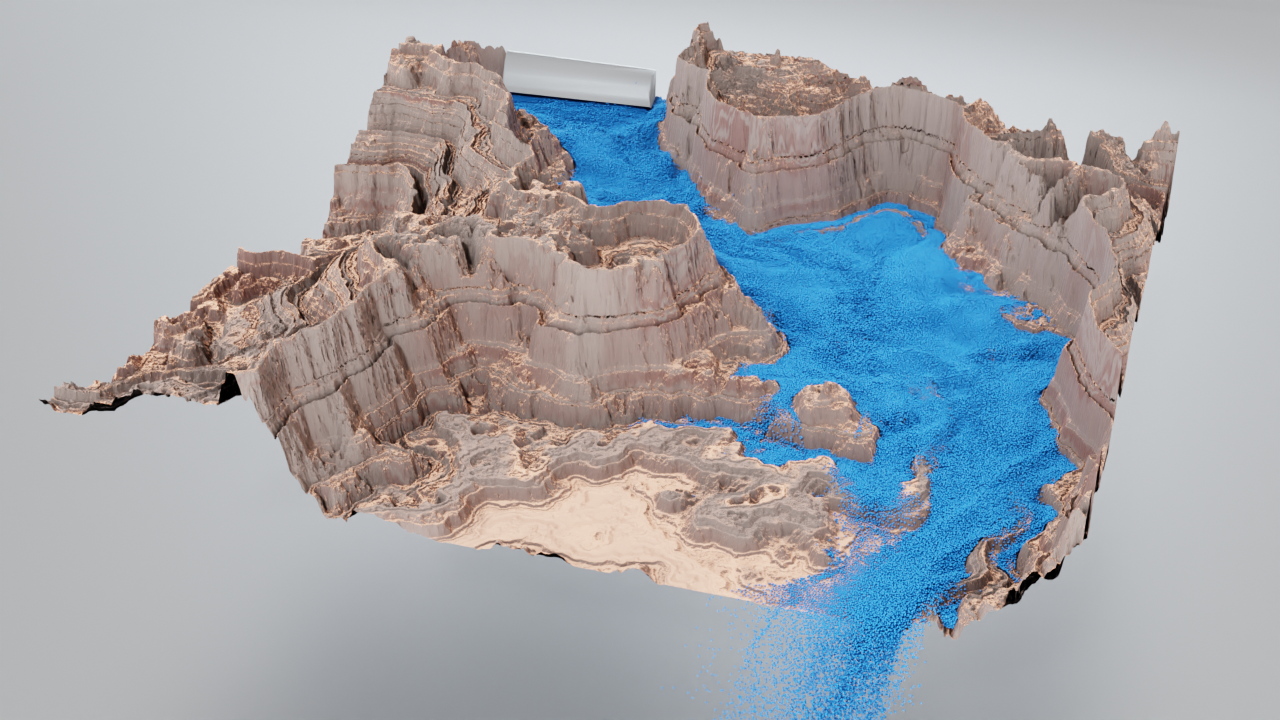}
\caption{Test scene with over one million particles after 800 time steps generated with our method.}
\label{fig:canyon}
\end{figure}

\mySpaceTweak{}
\paragraph{3D Simulations} 
Finally, we test our network with 3D data using the \texttt{Liquid3d} dataset. 
Unlike the other datasets, its simulations use a significantly larger time step ($8$ times larger than \texttt{WBC-SPH}).
Combined with the higher dimensionality, it is substantially more difficult to obtain stable solutions. Again, our method achieves comparable accuracy with much better generalization. Similar to the previous examples, it becomes apparent
that CConv tends to overly dampen the acceleration of gravity, which, unlike our method, leads to discrepancies in terms of the motion compared to the ground truth, as shown in Fig.~\ref{fig:3d_qual}.
To show the scalability and generalizability of our method, we apply it to multiple large-scale scenes with up to $165\times$ more particles than in the training set. Qualitative results can be found in Fig.~\ref{fig:canyon} and App. A.4 (Fig. \ref{fig:3d_big}). Our model robustly predicts the dynamics of these scenes using the same time step as for the smaller scenes.

\mySpaceTweak{}
\paragraph{Performance} 
We have also evaluated the computational performance of our method, 
with detailed measurements given in App. \ref{sec:more_results}.
E.g., for the 2D data \texttt{WBC-SPH} our method requires $67.25ms$ per frame on average. 
For comparison, the corresponding reference solver requires $10925ms$ per frame. Thus, our network achieves a speed-up of $162$ over the reference solver.

\mySpaceTweak{}
\section{Conclusion}
\mySpaceTweak{}
We have presented a method to guarantee the conservation of momentum in neural networks and demonstrated the importance of this concept for learning the complex dynamics of liquids. In particular, our approach outperforms state-of-the-art methods in terms of accuracy and generalization.
Nonetheless, the limits of symmetric constraints are far from exhausted. E.g., while our network is antisymmetric, permutation invariant, and translation invariant, 
an interesting avenue for future work will be to investigate more generic forms of $E(n)$ equivariance \cite{taco2016groupeq, walters2020trajectory, satorras2021n}.

\clearpage

\bibliographystyle{plain}
\bibliography{main}

\clearpage

\appendix

\section{Appendix}

In the following sections, we provide additional details about the network architecture, training, and experiments. The source code and WBC-SPH data set are published at \url{https://github.com/tum-pbs/DMCF}.

\subsection{Implementation Details}
\label{sec:impl_details}
We implement our neural network with Tensorflow (\url{https://www.tensorflow.org}), and use the Open3D library (\url{http://www.open3d.org}) for the continuous convolutions (CConvs). They also serve as the basis for the implementation of our antisymmetric CConv (ASCC) layer. 

\paragraph{Axis for Mirroring}
As mentioned in the main text, the mirror axis for ASCC layers can be chosen freely while fulfilling the requirements from theory. This provides a degree of freedom for implementation.
We decided to use a fixed axis, which in our case corresponds to the spatial y-axis.
While the mirroring could potentially be coupled to the spatial content of features, we found that a single, fixed axis for mirroring simplifies the implementation of the ASCCs, and hence is preferable in practice.

\paragraph{Additional Modifications}
In addition to the properties of our algorithm as discussed in Section \ref{sec:arch} and the ablation study in Section \ref{sec:results}, 
we normalize the input data depending on the given gravitational direction in the model. 
We have found that this slightly improves different directions of vector input quantities.
The output is denormalized again before continuing with the time integration steps. 
Moreover, to satisfy the condition of the constraint \ref{eq:eq_cond} from above, i.e. $P_D = P_Q$, it is important for the antisymmetric model to process the boundary particles in addition to the fluid particles as the input to the ASCC layer, even if only the output for the latter is used. 
This allows the ASCC layer to incorporate reactions to boundary conditions in its output directly, 
and it ensures the pairwise symmetry for \ref{eq:eq_cond}.
In this aspect, our method differs from previous methods such as CConvs, which process the boundary particles only in the input part and ignore them in the rest of the network. 
While it would be sufficient to add the boundary particles before the ASCC layer at the end of the model, in practice, we include the boundary particles in all layers. 
This affects performance due to the somewhat larger number of particles to be processed. However, in combination, the normalization and boundary handling lead to slight improvements in accuracy. We measure the influence of these two additions over the \emph{Preprocess} model in our ablation study, which shows an increase from 0.94 to the final 1.0 (\emph{Ours}) in terms of averaged relative performance.

\paragraph{Ablation Study}
The main text mentions that the ablation study score is evaluated relative to the final version $L_{Ours}/L$. In cases where the values become zero, however, e.g., for the change of momentum, the evaluation of this term is no longer well defined. In practice, we add a small constant epsilon via: $(L_{Ours} + \epsilon)/(L + \epsilon)$. We chose $10^{-100}$ as the value for $\epsilon$ to compute the relative scores provided in the main paper.

\subsection{Training Details}
\begin{wrapfigure}{t}{0.5\linewidth} 
    \vspace{-1cm}
 	\centering
 	\includegraphics[width=\linewidth, trim= 650 200 650 90, clip]{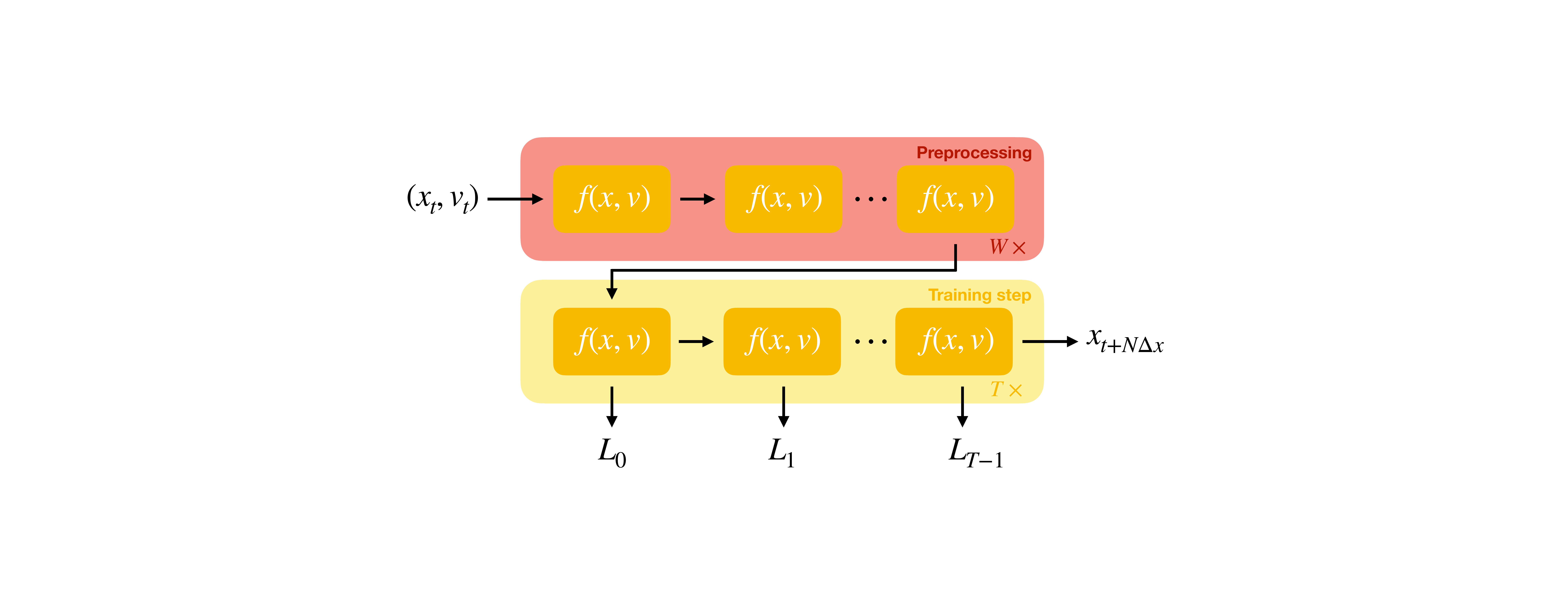}
 	\caption{
 	     Rollout at training time: Each $f(x,v)$ represents one time step. After a random number of  $W_{max}$ precalculation steps (red), the network is executed for $T$ time steps to compute the training loss (yellow). The gradient is evaluated only for the last part (yellow). 	      	}
 	\label{fig:rollout}
    \vspace{-5mm}
 \end{wrapfigure}
\label{sec:train_details}
For training, we use Adam as the optimizer and train with a batch size of $2$ and an initial learning rate of $0.001$. We use a scheduled learning rate decay and halve the learning rate in intervals of $5k$ iterations, starting at iteration $20k$.
The training has a total duration of $50k$ in iterations. For all convolutions, we use the random uniform initializer with range $[-0.05, 0.05]$. 
For additional temporal coverage of the training, we train our model with a rollout of $N=3$ frames. This value is increased to $N=5$ from step $15k$ onwards.
Similar to GNS \cite{sanchez2020learning}, a noise with standard distribution is added to the training input. We use a standard deviation of size $0.1r$, where $r$ corresponds to the particle radius of the data. We ran our training on an NVidia 2080ti GPU (12GB) for the 2D data sets and with an NVidia A6000 GPU (48GB) for the 3D data set.

\paragraph*{Preprocessing}
As discussed in Section \ref{sec:training}, we use custom preprocessing steps to improve long-time stability. 
We evaluate the network for a random number $W \in [0, W_{max})$ of time steps before providing the final state to the training step, as illustrated in Figure \ref{fig:rollout}. 
$W_{max}$ is continuously increased throughout the training, as long rollouts are not meaningful in earlier stages of the training process. We enable preprocessing starting from step $10k$ with a starting value of $W_{max}=5$. At step $20k$ and $30k$ we double the value of $W_{max}$.

Despite this scheduled increase, we found that the preprocessed states, due to their randomized nature, can lead to overly difficult states throughout the training. In the context of fluids, the maximum density of the fluid is a good indicator for problematic states. Hence, we stop the preprocessing iterations for a sample if it exceeds a chosen threshold in terms of
\begin{equation}
E = |1 - (\max\limits_{i} \rho(x_i) / \max\limits_{i} \rho(y_i))| ,
\end{equation}
with $\rho$ being a function to compute the density of particle $i$.
This ensures that the states at preprocessing time do not deviate too much from the ground truth. At the same time, the threshold preserves challenging situations during which it is essential to train the network such that it learns to stabilize the state of the system.

\subsubsection{Neural Network Architecture}
The neural network we employ has three distinct parts, as illustrated in Fig. \ref{fig:arch} of our main paper. The first part \emph{Type-aware Preprocessing} consists of several parallel CConv layers, one for each particle type in the input. We use two layers to process the two types of particles (fluid and obstacle). 
\nt{The feature dimension of the CConvs is 8 per particle type. 
As shown in previous work \cite{ummenhofer2019lagrangian}, 
this approach, performs better than type-specific input labels for particles.} 
The type-specific features generated in this way are concatenated for further processing in the following layers. 
Additionally, type-aware handling benefits using different features for the different particle types.
In addition to the spatial position, the input features are velocity and acceleration for the fluid particles and surface normals for the obstacle particles.

In the main body of our architecture, the \emph{Multi-scale Feature Aggregation} part, the preprocessed features are passed through several layers. The number of layers in our final network is $4$, each consisting of multiple CConvs working on $4$ \nt{different branches with different resolutions}. The first branch of these four retains the original scaling and is referred to as the main branch.
In the first layer $L1$ of the feature aggregation part, we split the features into the $4$ branches with $4$ different CConvs. For each CConv we use different query points, which we generate using the voxelization approach. The density of the selected query points determines the resolution of the output. After this first layer, we obtain different features with different resolutions for each branch. 
The scaling factor of the different branches is $1, \frac{1}{2}, \frac{1}{4}$ and $\frac{1}{8}$, respectively, with corresponding radii of $r, 2r, 4r$ and $8r$, where $r$ is the particle radius of the input data. The voxel size for voxelization is given by $\frac{r}{2}$ and is also divided by each branch's corresponding scaling factor.
The feature dimensions of the CConvs used in the first layer are $16, 8, 4$ and $4$, starting with the convolution of the main branch. 
In the second $L2$ and third $L3$ layer, the division into $4$ branches is maintained. We use $4$ CConvs per branch, each of which processes a multi-scale feature from the previous layer. The result is merged with an addition and corresponds to the new feature for the corresponding branch. This results in a total of 16 CConvs per layer. The feature dimension remains the same for all $4$ CConvs per branch but varies with the respective branch. For both layers, we use $32, 16, 8$, and $4$ feature channels. In the fourth layer, $L4$, the branches are merged back into the main branch. The $4$ CConvs have a feature dimension of $32$ each.

In the final and third sections, we use the anti-symmetric ASCC as the output layer, enforcing the desired conservation of momentum. The feature dimension of the ASCC layer is determined by the desired spatial output dimension. 

For all CConv layers we use a kernel size of 8 in all dimensions with a \emph{poly6} kernel \cite{muller2003particle} as window function. The same applies to the ASCC layer used, with the difference that we use a \emph{peak} kernel \cite{koschier2022survey}.
We found this prevents clustering of the particles compared to a \emph{poly6} kernel.

\begin{wrapfigure}{t}{0.32\linewidth} 
    \centering
	\includegraphics[height=0.9 \linewidth, trim=15 0 10 0, clip]{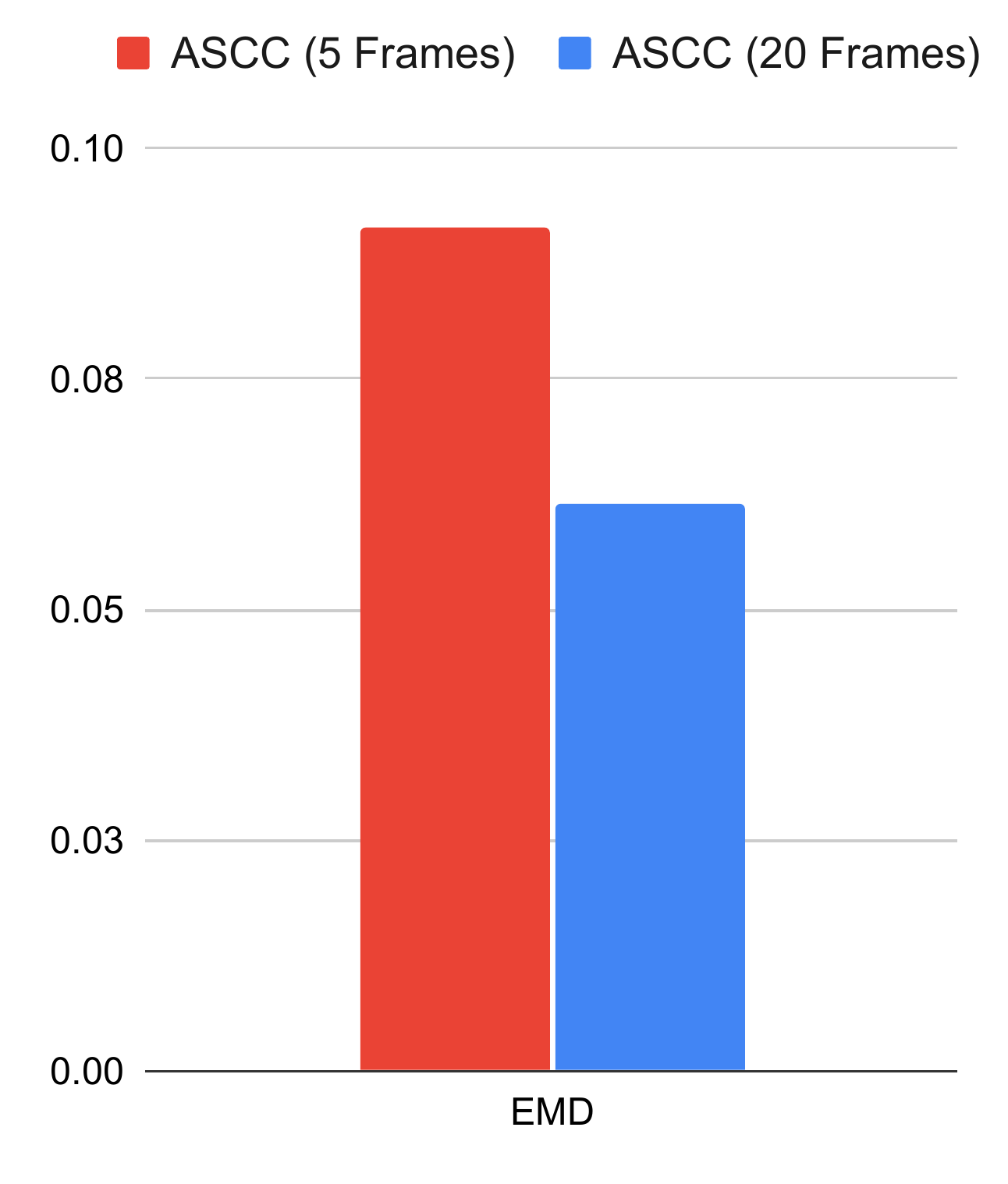} 
	\caption{Accuracy comparison for our \texttt{WaterRamps} model trained with $5$ and $20$ steps.}
	\label{fig:long_emd}
	\vspace{-6mm}
\end{wrapfigure}
The parameterization explained above is the default for the \texttt{WBC-SPH} data set. 
The \texttt{WaterRamps} data set instead contains smaller particle neighborhoods and less complex dynamics.
Hence, for training with this data set, we use the same hyperparameters, while the number of branches is reduced to $3$ by removing the branch with the lowest resolution. 
As with \texttt{WBC-SPH}, we use a maximal rollout of $5$ in training for the model used for the generalization (Section \ref{sec:results}) and the robustness test (Section \ref{sec:more_results}). However, for our final version for the comparison with GNS we found that a longer rollout further improves results. Here we use a rollout of $20$ steps, which results in a EMD error of $0.06156$, compared to $0.09155$ for the 5-step model. This is also shown in Figure \ref{fig:long_emd}.
The same applies to the network for the \texttt{Liquid3d} data set. Here we have additionally removed the third layer and reduced the size of the CConv kernel to $4$ and of the ASCC kernel to $6$.

\paragraph*{PointNet}
PointNet can be seen as a fundamental baseline, similar to a fully-connected network in other problem settings.
The original method was initially designed for classification instead of regression tasks. The input to PointNet is usually the complete point cloud, from which a single scalar/vector is generated. To adapt it for our problem setup, we evaluate the PointNet for all input particles, thus generating output for each particle. Additionally, we use only the considered particle's neighborhood as input.

For the PointNet \cite{pointnet} baseline, we used a fully-connected neural network with $5$ layers that process particles individually. In order to establish a relation amongst the neighboring particles, we additionally accumulated the features of the neighboring particles after each fully-connected layer with a \emph{poly6} kernel for each particle. The number of neurons per layer was $64, 128, 128, 128$, and $3$.

\paragraph*{GNS}
For our tests with the GNS \cite{sanchez2020learning} model, we use the official implementation provided by the authors.
We trained the GNS model at first with the provided \texttt{WaterRamps} data set, using the code from the original paper without modifications, for 5M iterations until the validation loss converged. 

In addition, we trained a GNS model with our \texttt{WBC-SPH} data set. For this, we modified the hyperparameters of the network to fit our data set by halving the search radius to $0.0075$, setting the batch size to 1, 
and reducing the input noise to $3.3e^{-4}$.
Again, we trained the network until the loss converged after 1.25M steps.
For training, we proceed as for the \texttt{WaterRamps} data set: the GNS receives a sequence of 6 temporal frames as input from which velocity and acceleration are constructed. It is worth noting that our ASCC model only receives a single frame as input, with the current acceleration and velocity as additional features. Thus, the varying external forces of the data set can be constructed for the GNS from the data sequence, whereas for our method, the acceleration of the particles depends on it.

The set of six time steps used as input provides the GNS network with additional temporal information. According to the authors, this plays a vital role in generating stable results \cite{sanchez2020learning}. E.g., the network can reconstruct the acceleration acting on the particles from the provided sequence. This approach has the limitation that the six frames must be provided, pre-computed, and processed each time. 
In our case, we restrict the network to work with a single frame as input while the velocity and acceleration of the particles are provided explicitly as input features. E.g., this allows our method to work with a static initialization frame without the need to generate a sequence beforehand. 

Another difference in terms of architecture is that the GNS does not add gravity accelerations to the model outputs, which, according to the authors, does not influence the performance. Hence, GNS learns the acceleration due to gravity for free-falling particles as well, unlike our method, where the gravitational acceleration is applied independently of the network. This has the advantage that GNS does not have to compensate for the gravitational effect in hydrostatic conditions, which we found essential for stable simulations.
On the other hand, our model directly generalizes to different external forces than gravity. If necessary, these forces likewise would have to be provided as inputs.

\paragraph*{CConv}
We likewise used the author's implementation for the CConv \cite{ummenhofer2019lagrangian} model.
When training the CConv model with our \texttt{WBC-SPH} or with the \texttt{WaterRamps} data set, 
we halved the search radius of the CConvs compared to the original. Similar to our method, the CConv model does not receive a sequence of data as input, and hence we pass the acceleration as an additional feature as input to the network.

\subsubsection{Discussion: Comparing CConv and GNS}\label{app:discuss} Our method builds on CConvs as central building blocks for our neural network architecture.
As an alternative to CConvs, Graph Neural Networks (GNNs)
provide an established framework for processing unstructured data. 
Even if the data is not given in the form of an explicit graph structure, a graph can be created dynamically by creating new edges based on the proximity of particles and a distance threshold. 
While graphs and spatial convolutions can be seen as equivalent representations that can be transformed into one another for a given discretization with particles, CConvs contain an explicit inductive bias in the form of positional information of the convolution kernel. The relative positions of query points are directly put into correlation. This is a crucial operation of classical Lagrangian discretizations such as SPH and is supported by CConvs without having to be learned and encoded in parameters. Consequently, CConvs yield leaner networks with correspondingly faster evaluation and training times. 
Additionally, in contrast to graphs, the kernels are regularized by construction through the discretization. This improves the generalization to different sampling densities, as we show below.

\subsection{Simulation Data Sets}
\label{sec:sim_data}
The data sets for the evaluation of our method are based on particle-based fluid simulations. The data was generated with different solvers, the properties of which we discuss in detail below. As spatial units, we use meters.

\paragraph*{Liquid Column}
For the liquid column data set, we use an iterative solver following He et al.~\cite{he2012local} with an error threshold of $0.01$. We use a particle radius of $0.005m$ and a time step of $2.5ms$. The fluid viscosity was set to $1e^{-4}$, and the stiffness for the pressure computation to $10$. 
For training data, we use columns with a particle count from $1$ to $40$ over $100$ time steps, where the boundary consists of two particles. For the evaluation, we use a subset of $10$ scenes from the data set and additionally generate $5$ scenes with $1$ to $5$ particles in free fall with a starting height of $1cm$.
For the explicit solver used as a comparison in the evaluation, we use the method by Prem\v{z}oe et al.~\cite{premvzoe2003particle}, with the same settings as for the iterative solver, apart from a smaller time step of $0.25ms$. This was necessary as the simulations were not stable with a larger time step.

\paragraph*{WBC-SPH Data Set}
This data set is based on the WBC-SPH solver by Adami et al.~\cite{adami2012}.
We use a particle radius of $0.005m$ and a time step of $2.5ms$. The scenes consist of randomly generated  fluid volumes and obstacles with a static, square-shaped outer boundary. 
The fluid particles are simulated over $3200$ frames, with gravity having a random magnitude and direction for each scene. The gravitational strength can be up to $1.5g$. The randomization of gravity generates data with high variance and allows for a high degree of generalizability, e.g., fluid simulation without gravity or with other external forces than earth's gravity. With this setup, we generate 50 scenes for training, $10$ scenes for validation, and $5$ scenes for the test data set. In our test data set, two simulations of a hydrostatic tank with a liquid height of $10cm$ and $25cm$ are added for diversification, as well as two simulations of colliding fluid drops, once with and once without gravity.

\paragraph*{WaterRamps Data Set}
The \texttt{WaterRamps} data set is based on an MPM solver \cite{taichi} and stems from GNS \cite{sanchez2020learning}. The data is two-dimensional, and the particle radius is twice as large at $0.01m$ compared to the \texttt{WBC-SPH} data set, with a time step of $2.5ms$.

\paragraph*{Liquid3d Data Set}
For the 3D evaluations we use the \texttt{Liquid3d} data set from Ummenhofer et al.~\cite{ummenhofer2019lagrangian}. The data set is based on the DFSPH solver \cite{bender2016divergence} with a particle radius of $0.025m$, and a time step of $0.02s$. Hence, the data is sampled more coarsely than the other two data sets, while the time step is $8$ times as large.
A large time step makes the data difficult to learn because the discrepancy between the input to our network and the targeted reference becomes larger after a position update based on the integration of the external forces. Thus, the network must learn a much larger correction. In addition, the data set is three-dimensional, which introduces more degrees of freedom and additional complexity. This makes the data set a suitable and challenging environment to evaluate for our method.

\clearpage

\subsection{Additional Results}
\label{sec:more_results}

\begin{figure}[t]
 \centering
 \begin{subfigure}[c]{0.49\textwidth}
     \includegraphics[height=3cm, trim=0 0 0 20, clip]{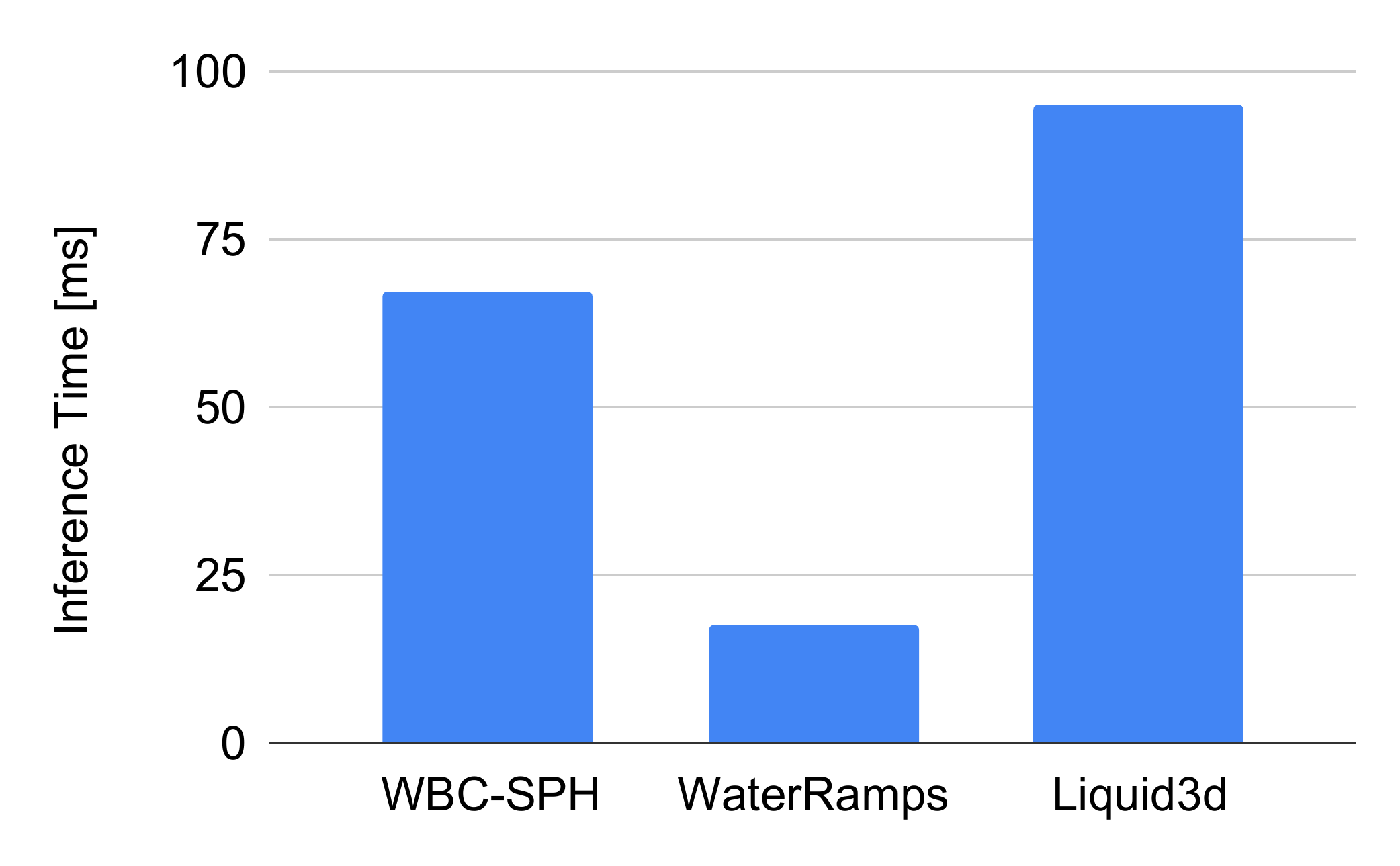}
     \subcaption{
          Average inference time for single frames.
     }
 \label{fig:time}
 \end{subfigure}
 \begin{subfigure}[c]{0.49\textwidth}
    \includegraphics[height=3cm, trim=0 0 0 20, clip]{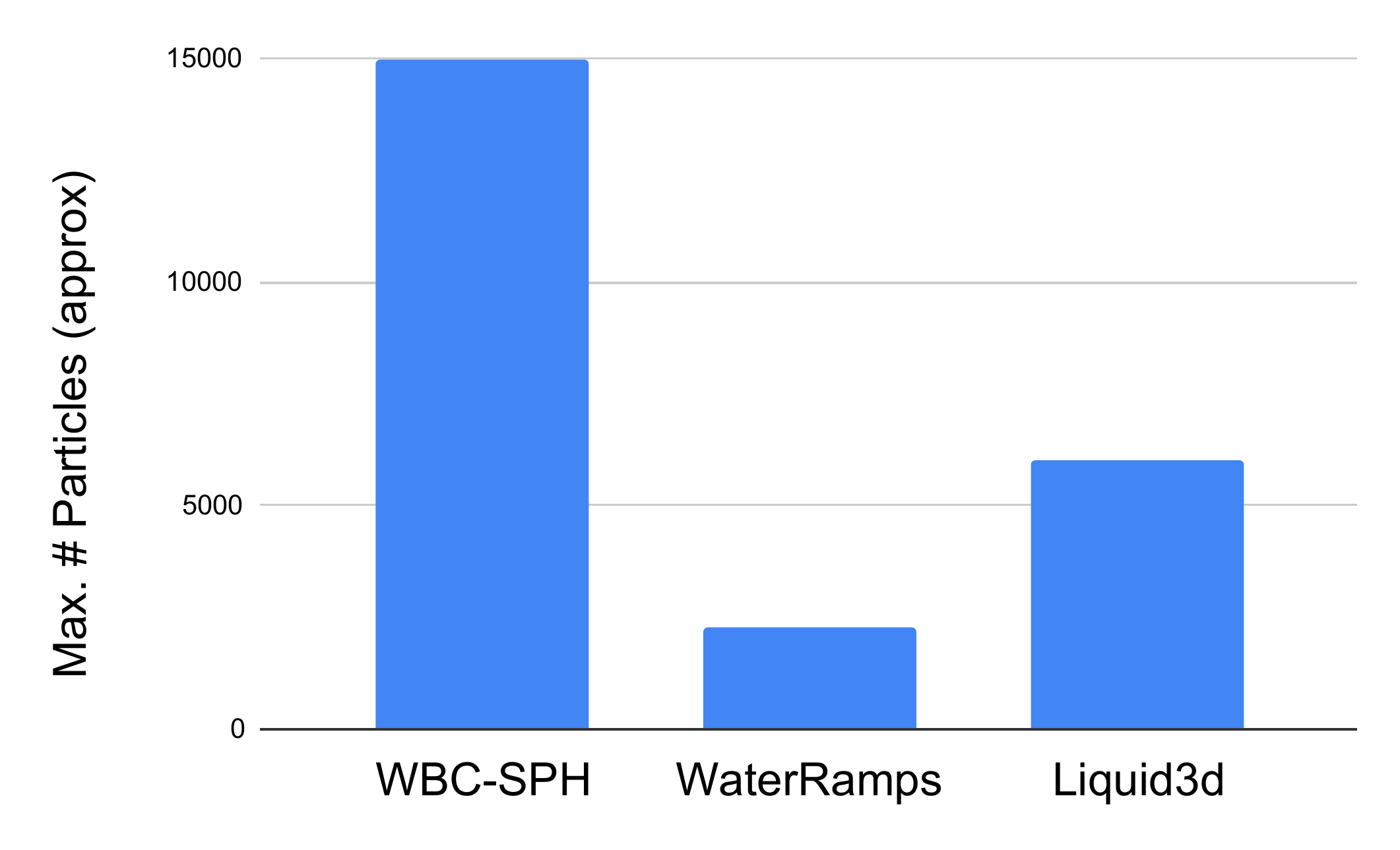}
     \subcaption{
          Approximate maximum number of particles per randomized initialization for each data set.
     }
 \label{fig:cnt}
 \end{subfigure}
 \caption{Runtime (a) for our method and the number of particles (b) for the used datasets.} 
 \label{fig:time_cnt}
\end{figure}

\paragraph{Evaluation of Performance}
\begin{wrapfigure}{t}{0.2\linewidth} 
    \vspace{-5mm}
 \centering
     \includegraphics[width=0.2\textwidth]{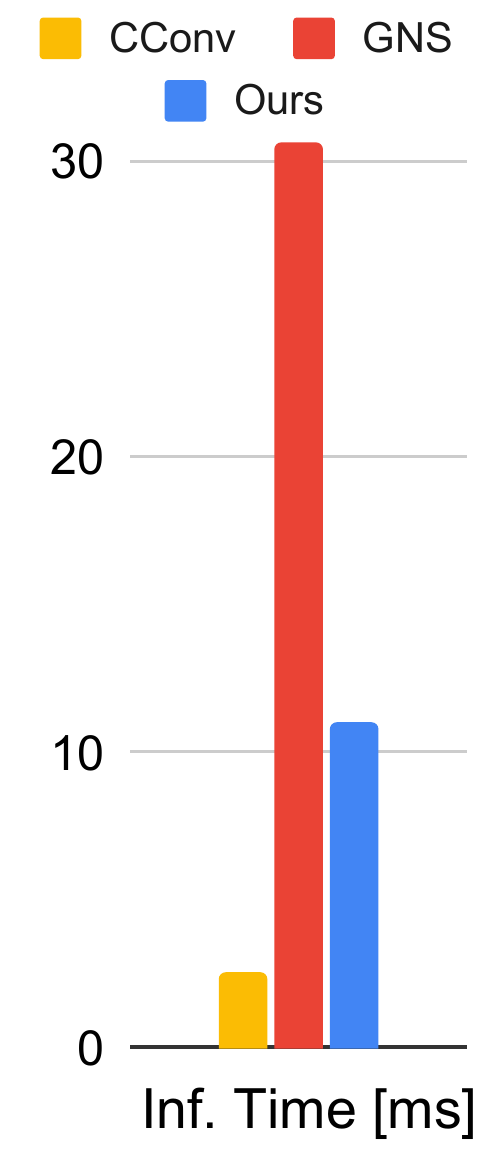}
    \caption{Runtime for different models.} 
    \label{fig:time_inf}
    \vspace{-7mm}
\end{wrapfigure}
Figure~\ref{fig:time} shows the average execution time of our network for the inference of a single frame of simulation for each data set. 
It is noticeable that the inference time for the 2D \texttt{WBC-SPH} data set is larger than for the other 2D case. This behavior is caused by the fact that there are significantly more particles in one frame of the high-resolution data set \texttt{WBC-SPH}, as shown in Figure \ref{fig:cnt}. Hence, the performance directly correlates with the number of particles that needs to be processed.
In addition, we evaluated the inference performance of the different approaches.
As can be seen in Figure \ref{fig:time_inf},
the performance correlates with the model sizes, our model having ~0.47m parameters, GNS with ~1.59m, and CConv with ~0.18m. 
The smallest model, CConv, is the fastest with $2.57$ms. Our model has the second fastest inference time with $10.98$ms and is almost three times faster than GNS with $30.63$ms. Thus, our model provides the best tradeoff between inference time and accuracy among the three methods.

\paragraph{Robustness}
\begin{figure}[b]
\centering
\begin{minipage}[t]{.45\textwidth}
 \centering
     \includegraphics[width=\textwidth]{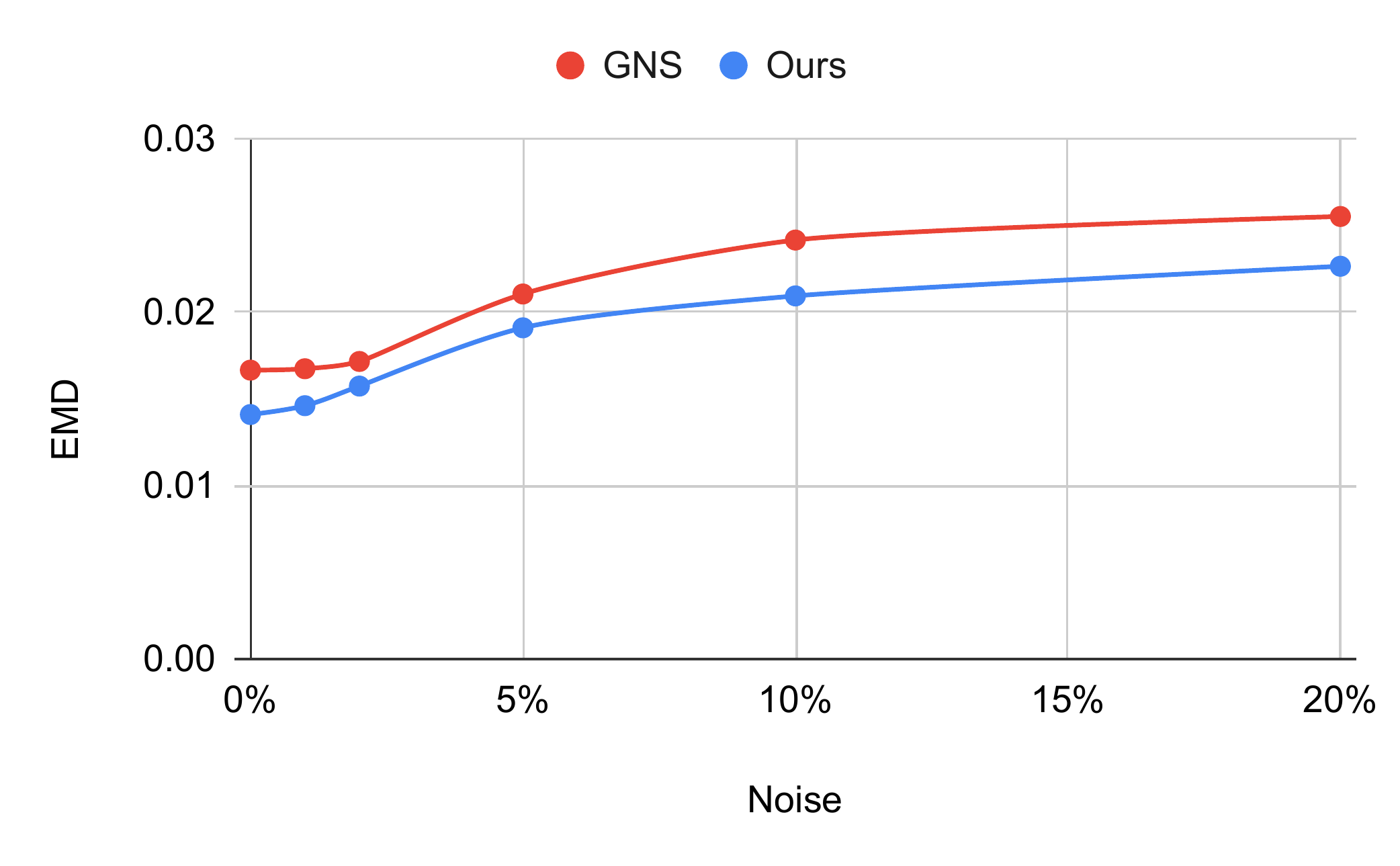}
    \caption{Inference accuracy for varying amounts of input noise. 
    The standard deviation of the noise is expressed as a percentage of the particle radius.} 
    \label{fig:noise}
\end{minipage}\hspace{1cm}
\begin{minipage}[t]{.45\textwidth}
 \centering
    \includegraphics[width=\textwidth]{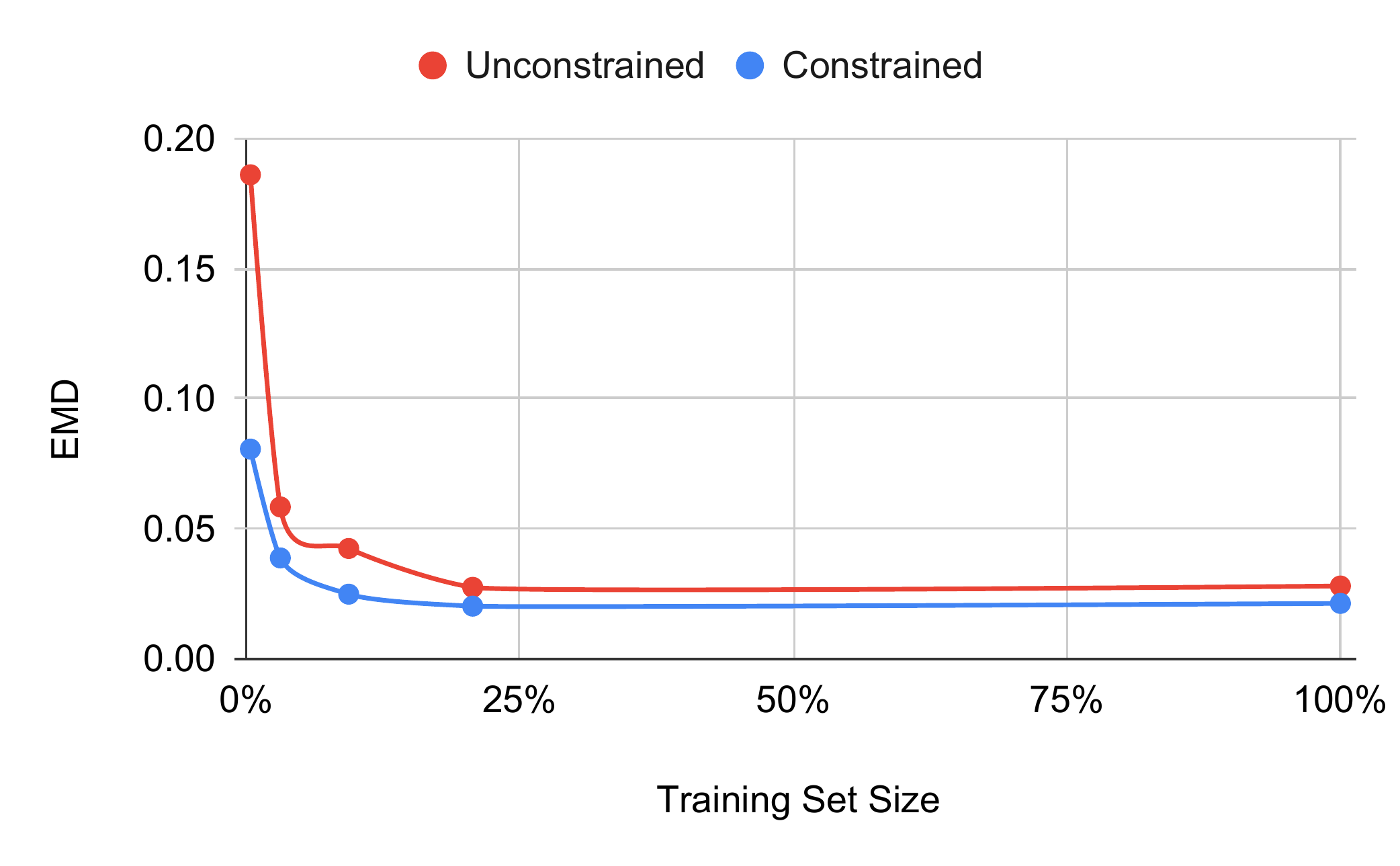}
    \caption{Accuracy of a constrained (blue) and unconstrained (red) model for varying training data set sizes. The first data point corresponds to 0.39\% of the training data while 100\% corresponds to the complete training data set as described in Sec.~\ref{sec:sim_data}.} 
    \label{fig:train_size_emd}
\end{minipage}
\end{figure}
We also evaluate our method's robustness to input noise and sampling density. 
First, we measure the change in EMD for varying input noise. The noise is normally distributed with a standard deviation based on the particle radius. The evaluation is based on the \texttt{WaterRamps} data set, and we also evaluate the GNS with noise as a comparison.
We perturb the input particles' position with the noise and evaluate the model with these perturbed inputs. Since GNS needs six input frames, we add the noise to all input frames. We keep the noise per particle constant for all six frames so that the velocity and acceleration derived by GNS from the input frames are not affected. In line with this treatment, we do not perturb the input velocity for the ASCC model. 
We show the evaluation results in Fig. \ref{fig:noise}. It can be seen that both models perform equally well in the presence of noise. Even with a considerably strong relative noise of 20\%, reasonable accuracy is still achieved. It is important to note that we only evaluate the first 50 frames for this comparison, as this time span is where the effects of the noise are most noticeable.

\begin{figure}[t]
 \centering
 \begin{subfigure}[b]{0.5\textwidth}
     \includegraphics[width=\textwidth, trim=200 0 200 0, clip]{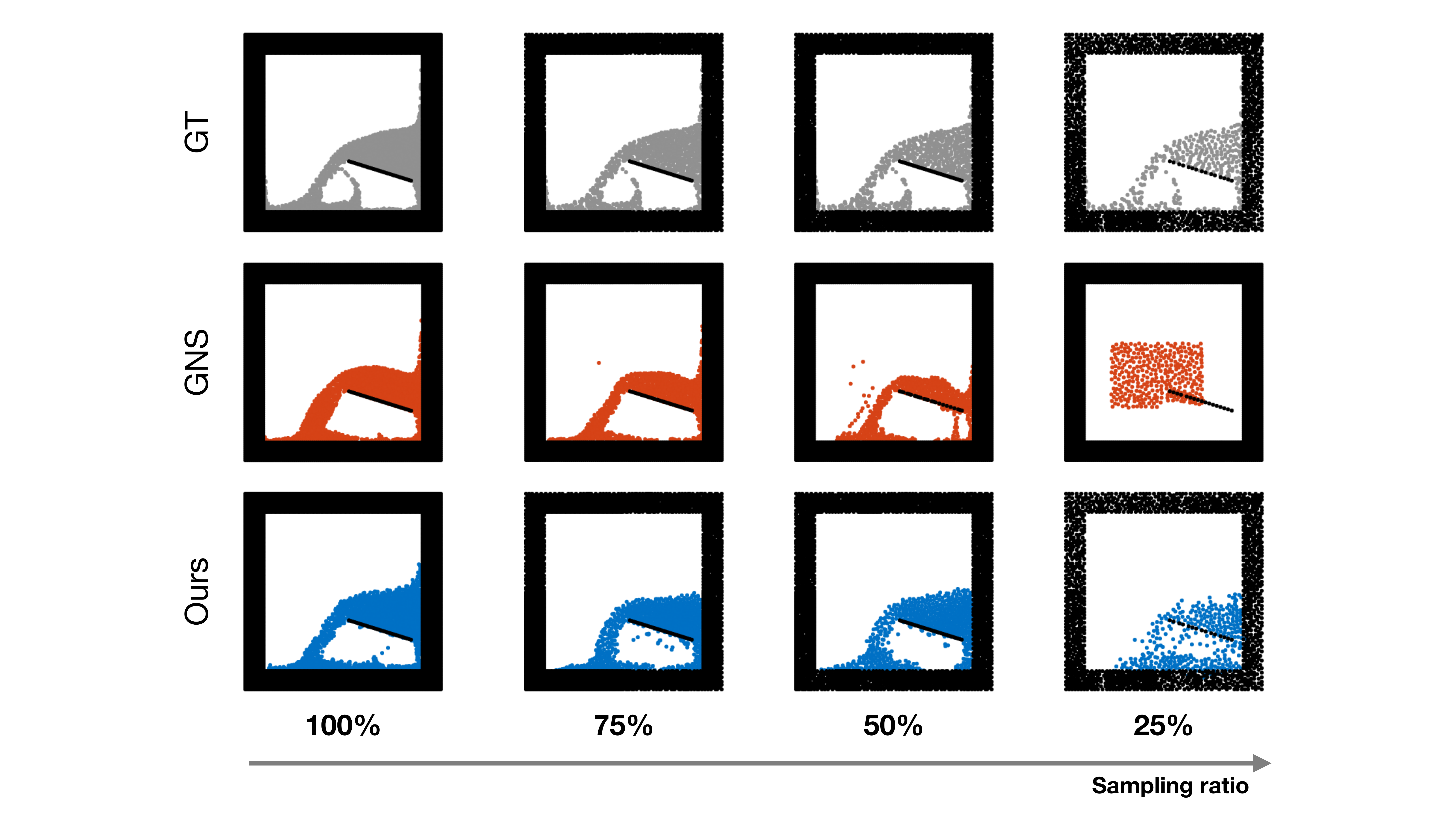}
    \caption{Frame 160 from a sequence of the \texttt{WaterRamps} test set, evaluated with different sampling ratios.} 
 \end{subfigure}
 \hspace{1cm}
 \begin{subfigure}[b]{0.4\textwidth}
 \centering
    \includegraphics[width=\textwidth]{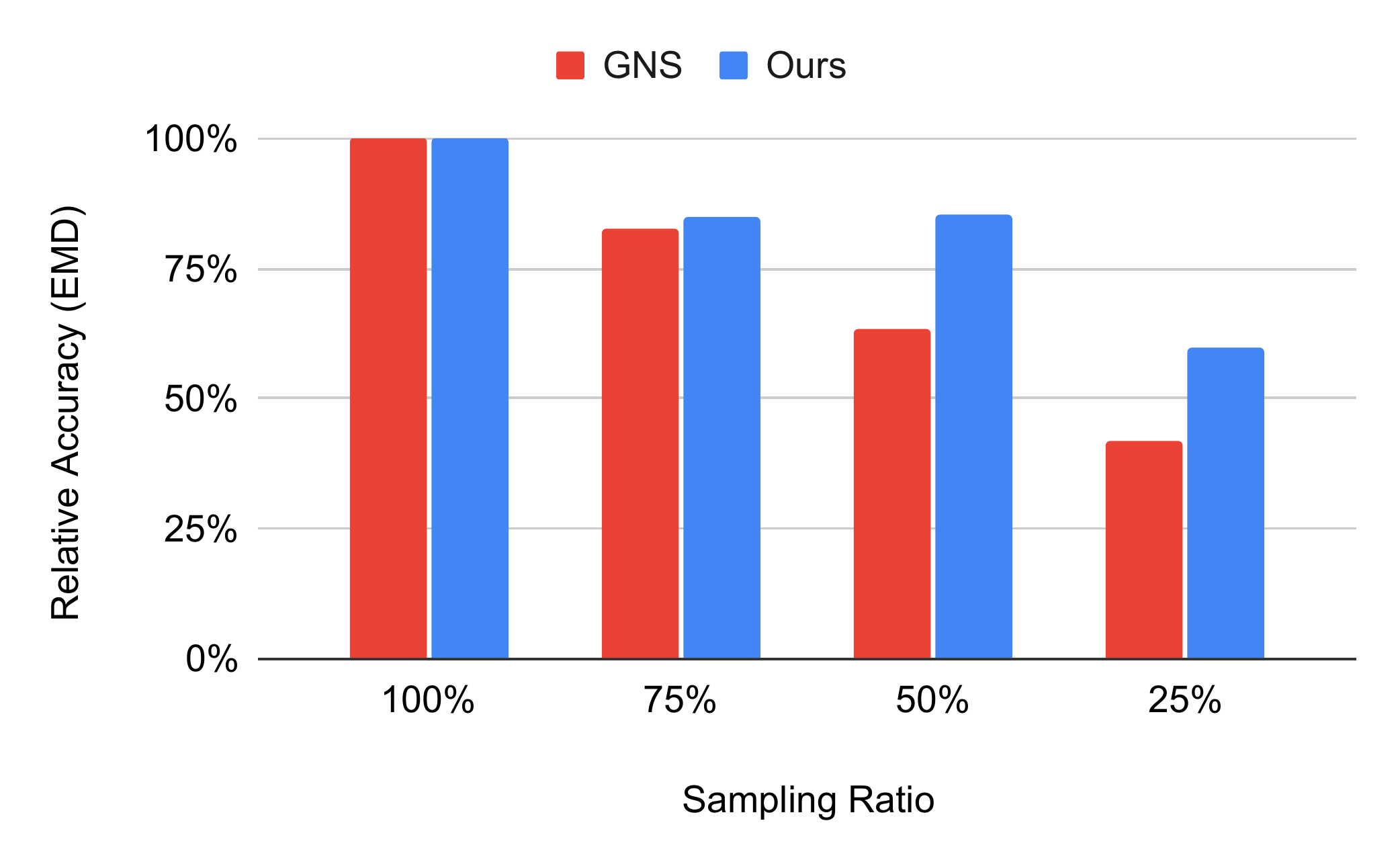}
    \caption{Relative accuracy of the models with different sampling ratios. The accuracy is evaluated in relation to the accuracy with a ratio of 100\%.} 
 \end{subfigure}
\caption{Qualitative and quantitative results of the sample efficiency evaluation.} 
\label{fig:sampling_rel}
\end{figure}

As a second test, we evaluate our model with data with a different sampling density than at training time. For this, we subsample the test data with different sampling factors, which reduces the number of input particles. We refer to this as the \textit{sampling ratio} in the following, where a sampling ratio of 100\% corresponds to the data with the original sampling density. 
Reducing the number of particles reduces the number of neighbors when evaluating CConv, and correspondingly down-scales the output of the convolutions.
To counteract this behavior, we multiply the kernels of the CConv with the subsampling factor. 
For comparison, we also evaluate a GNS with inputs with different sampling ratios. 
Here, we multiply the accumulated value of the edge features by the subsampling factor in the GNN to compensate for the reduced amount of neighbors. 
The results are given in Fig. \ref{fig:sampling_rel}. As can be seen, our model maintains high accuracy of 85\% up to a subsampling factor of 2. GNS, on the other hand, performs worse. 
While our model still produces meaningful results even with a sampling ratio of 25\%, the GNS does not manage to respond correctly to particle-based boundaries. The liquid volume falls in the wrong direction and through the obstacles. It is important to note that GNS processes the square border as an implicit representation, which is not affected by the sampling. Thus, the change in sampling density does not affect the fluid-border interaction, which artificially boosts the performance of the GNS. 
\paragraph{Sample Efficiency}
Incorporating inductive biases typically leads to improvements in terms of sample efficiency. To evaluate this aspect, we train two variants of models, one with an anti-symmetric constraint and one without the constraint, with different subsets of the training data and compare the resulting accuracy. 
The results in Figure \ref{fig:train_size_emd} show EMD as a function of the relative training data size for the \textit{WBC-SPH} data set. It is noticeable that the constrained method performs better throughout all tests. With a training data size of ca. 6\%, the constrained model achieves a performance similar to the unconstrained approach with more than 20\% of the data. This highlights the advantages of our constraints for conservation of momentum  in terms of sample efficiency.

\clearpage

\paragraph{Evaluation Details}
Below, we provide additional qualitative results as well as tables with numerical values corresponding to the graphs shown in the main paper.

\begin{figure}[h!]
	\label{fig:ablation_qual}
	\centering
    \begin{subfigure}[c]{0.13\textwidth}
	\includegraphics[width=\textwidth, trim= 1140 0 357 0, clip]{images/ablation/target.png} 
	\subcaption{Target}
	\end{subfigure}
    \begin{subfigure}[c]{0.13\textwidth}
	\includegraphics[width=\textwidth, trim= 1140 0 357 0, clip]{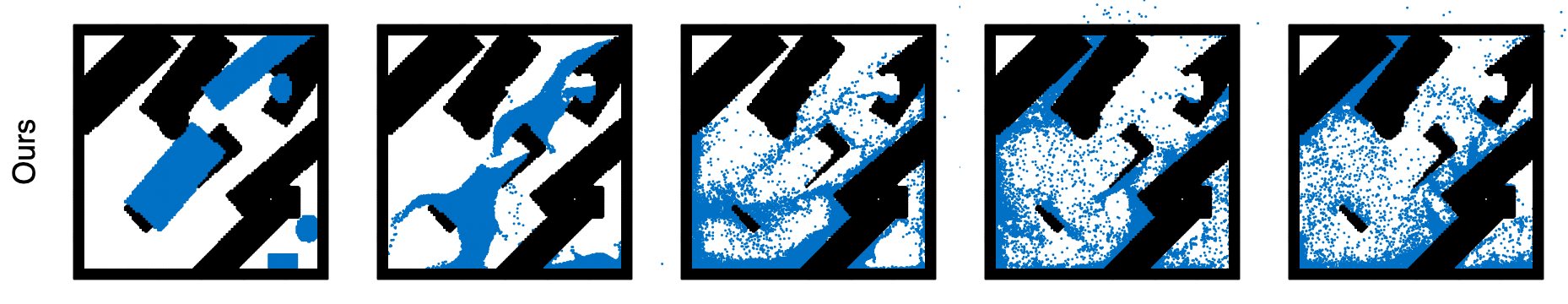} 
	\subcaption{Base}
	\end{subfigure}
    \begin{subfigure}[c]{0.13\textwidth}
	\includegraphics[width=\textwidth, trim= 1140 0 357 0, clip]{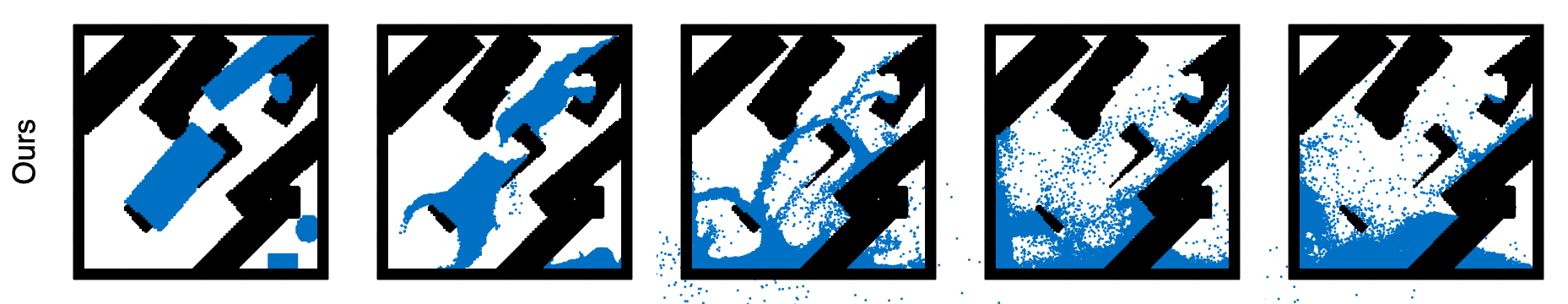} 
	\subcaption{ASCC}
	\end{subfigure}
    \begin{subfigure}[c]{0.13\textwidth}
	\includegraphics[width=\textwidth, trim= 1140 0 357 0, clip]{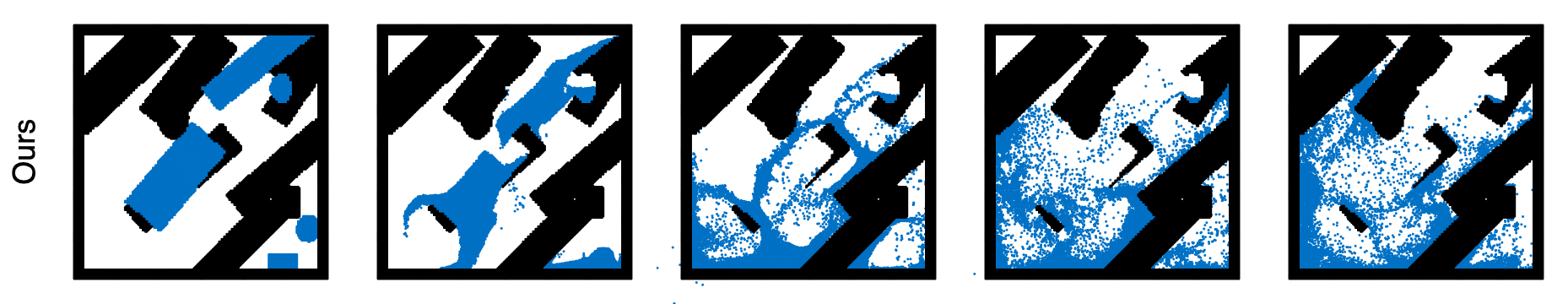} 
	\subcaption{Multi Sc.}
	\end{subfigure}
    \begin{subfigure}[c]{0.13\textwidth}
	\includegraphics[width=\textwidth, trim= 1140 0 357 0, clip]{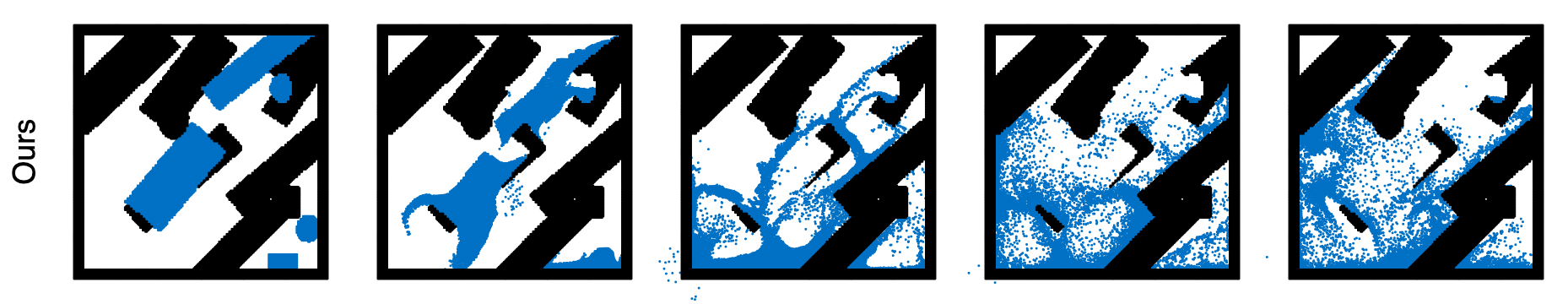} 
	\subcaption{Voxel.}
	\end{subfigure}
    \begin{subfigure}[c]{0.13\textwidth}
	\includegraphics[width=\textwidth, trim= 1140 0 357 0, clip]{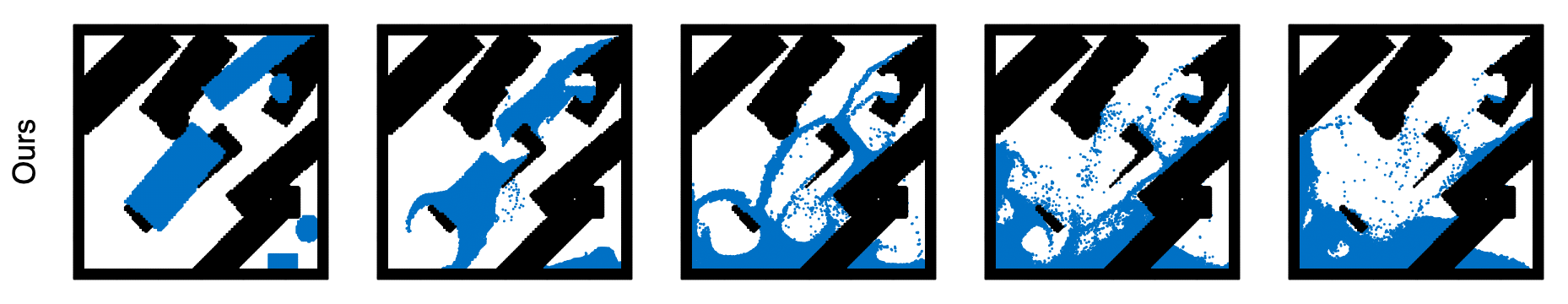} 
	\subcaption{Prepro.}
	\end{subfigure}
    \begin{subfigure}[c]{0.13\textwidth}
	\includegraphics[width=\textwidth, trim= 1140 0 357 0, clip]{images/ablation/ours.png}
	\subcaption{Ours}
	\end{subfigure}
	\caption{Frame 240 from a sample sequence for the ablation study. The gradual improvement in quality is clearly visible, with a big jump from voxelization to preprocessing. From then on, the results are much more stable.}
\end{figure}

\begin{figure}[h!]
	\centering
	\includegraphics[width=\textwidth]{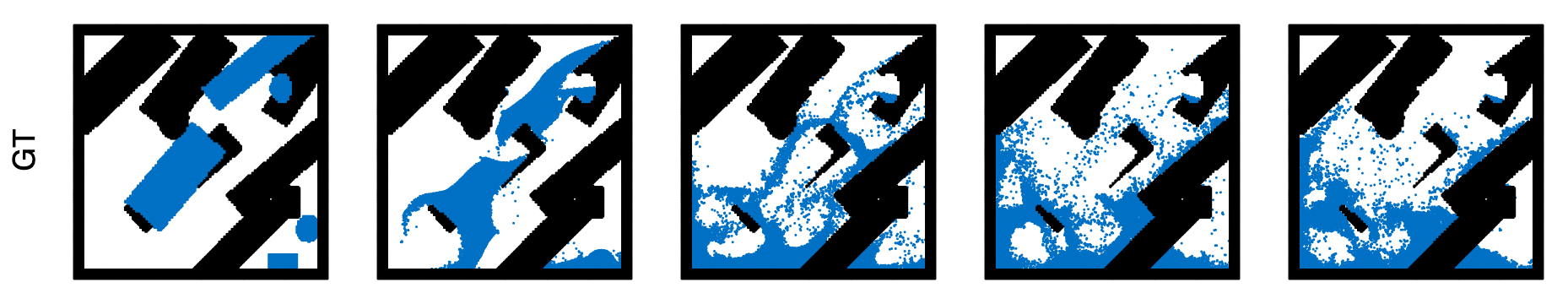} 
	\includegraphics[width=\textwidth]{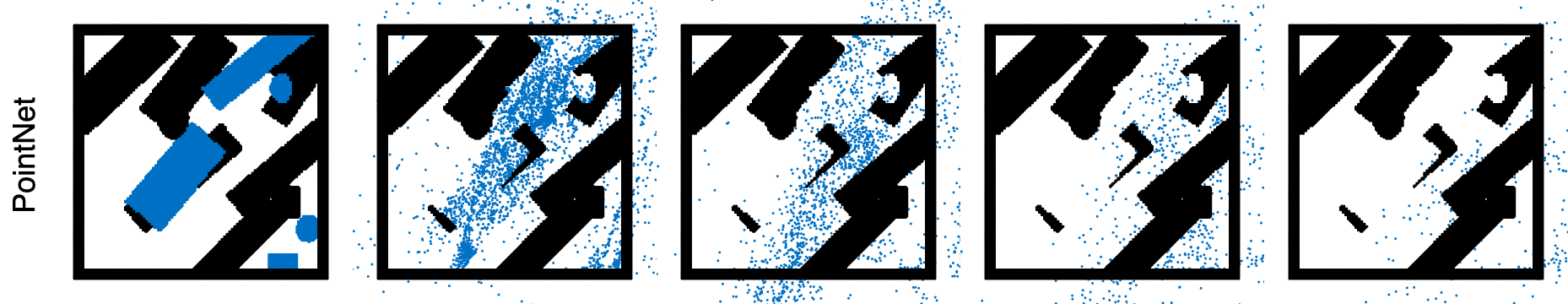}
	\includegraphics[width=\textwidth]{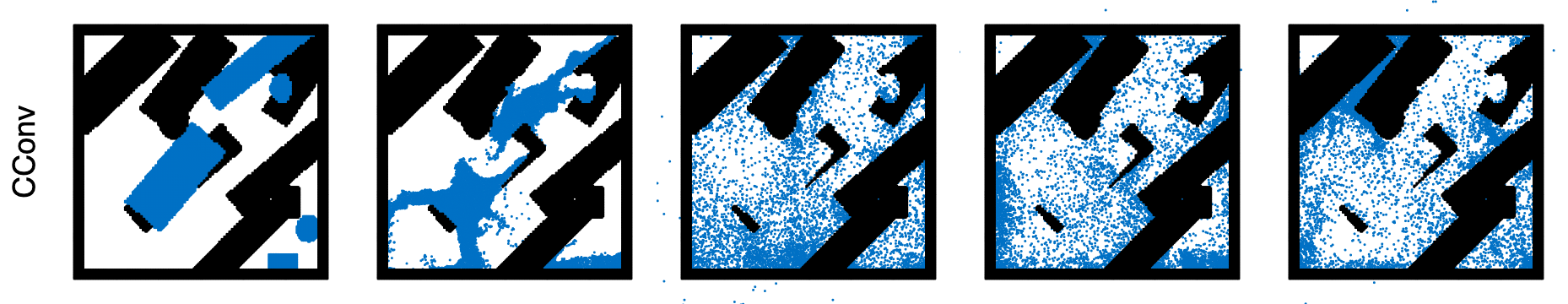}
	\includegraphics[width=\textwidth]{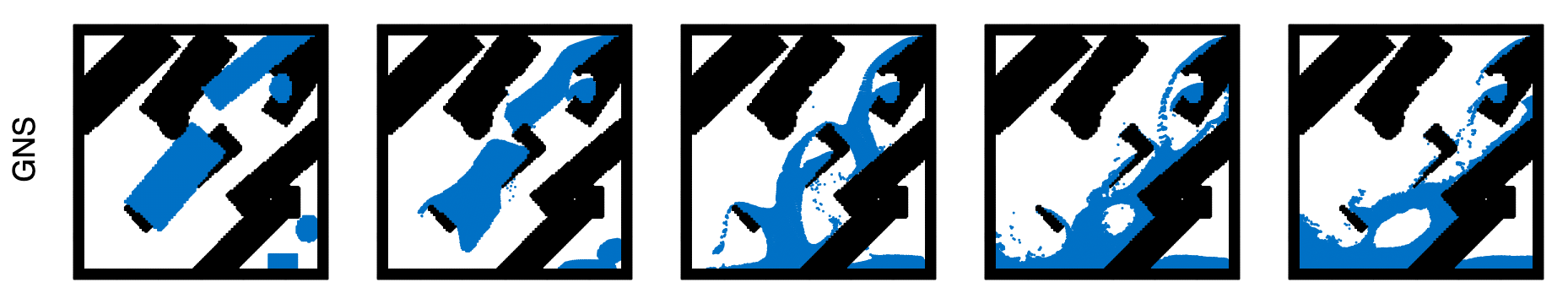}
	\includegraphics[width=\textwidth]{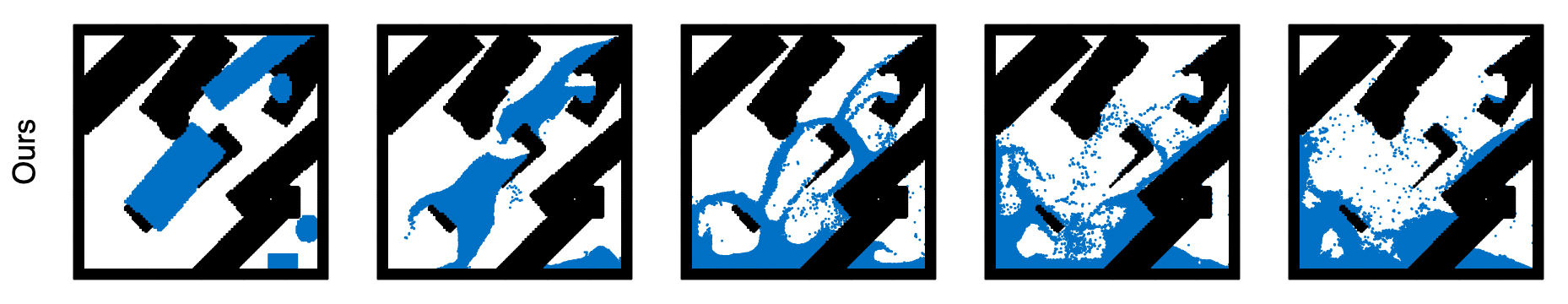}
	\caption{	\bu{A test sequence from our \texttt{WBC-SPH} data set.}
	}
\end{figure}

\begin{figure}[h!]
	\centering
	\includegraphics[width=\textwidth]{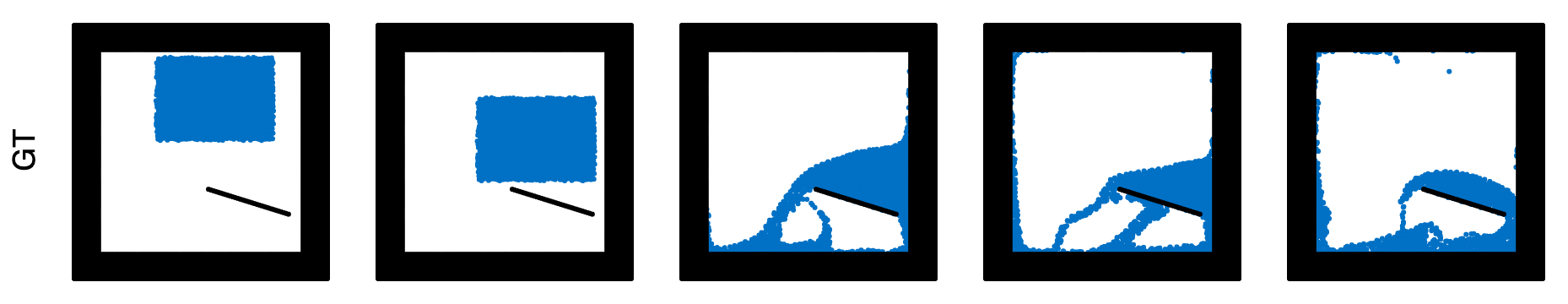} 
	\includegraphics[width=\textwidth]{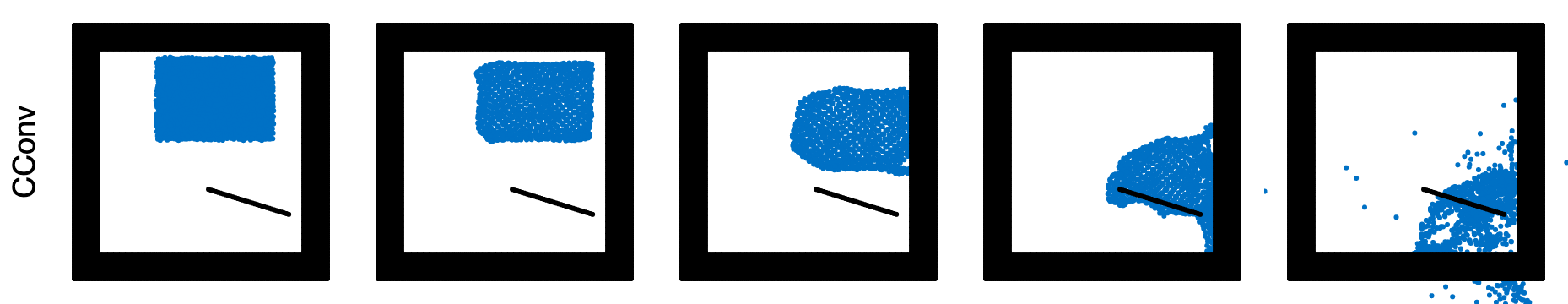} 
	\includegraphics[width=\textwidth]{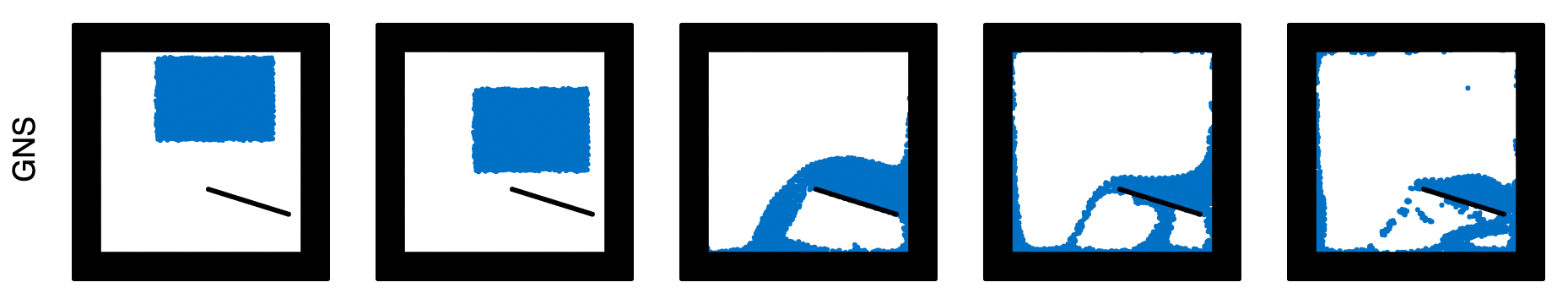} 
	\includegraphics[width=\textwidth]{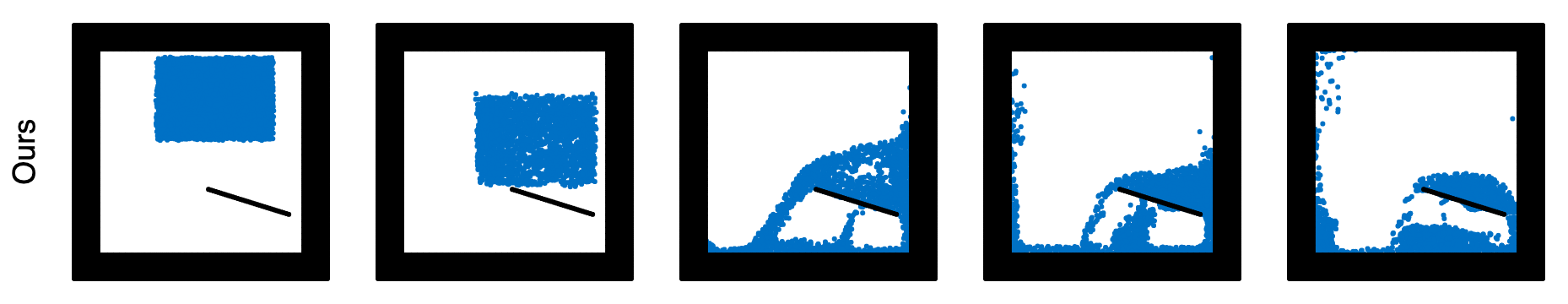} 
    \rule{\linewidth}{0.4pt}
	\includegraphics[width=\textwidth]{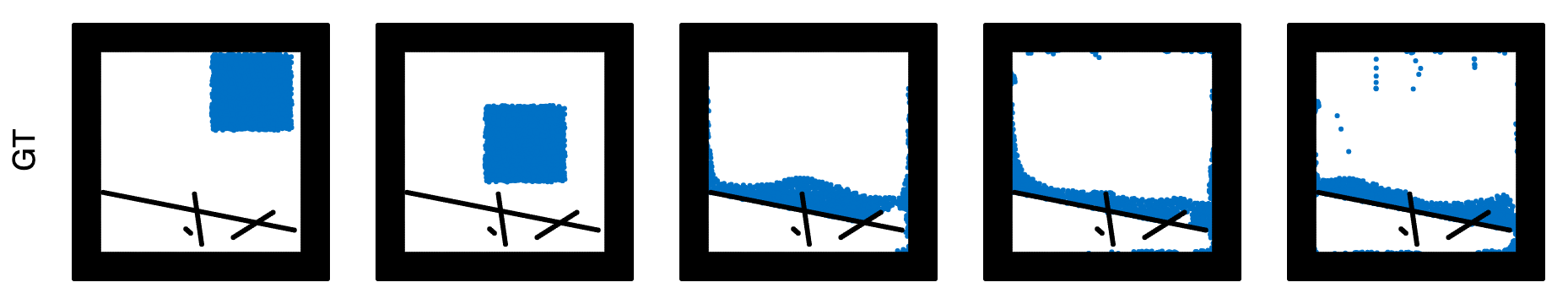}
	\includegraphics[width=\textwidth]{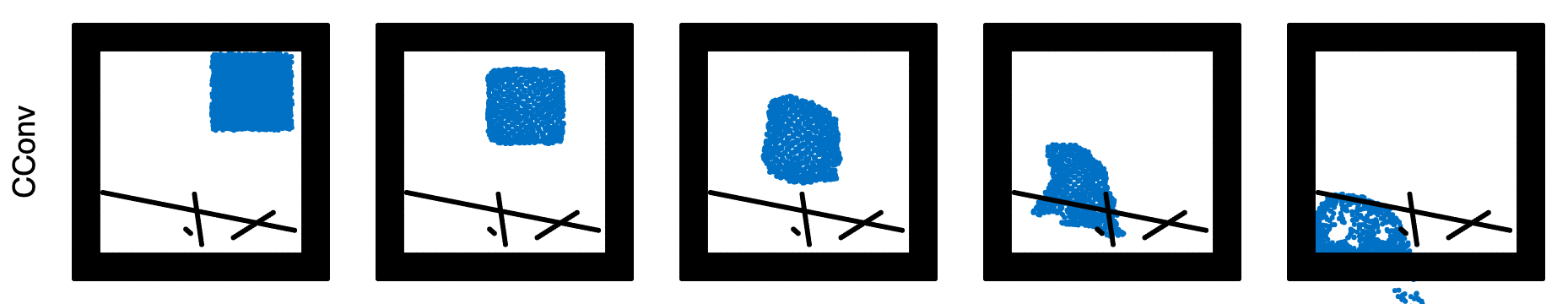}
	\includegraphics[width=\textwidth]{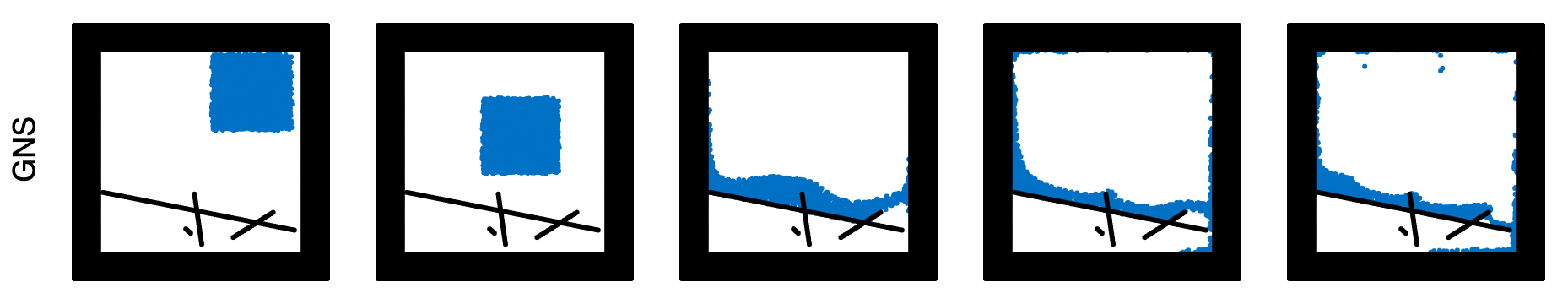}
	\includegraphics[width=\textwidth]{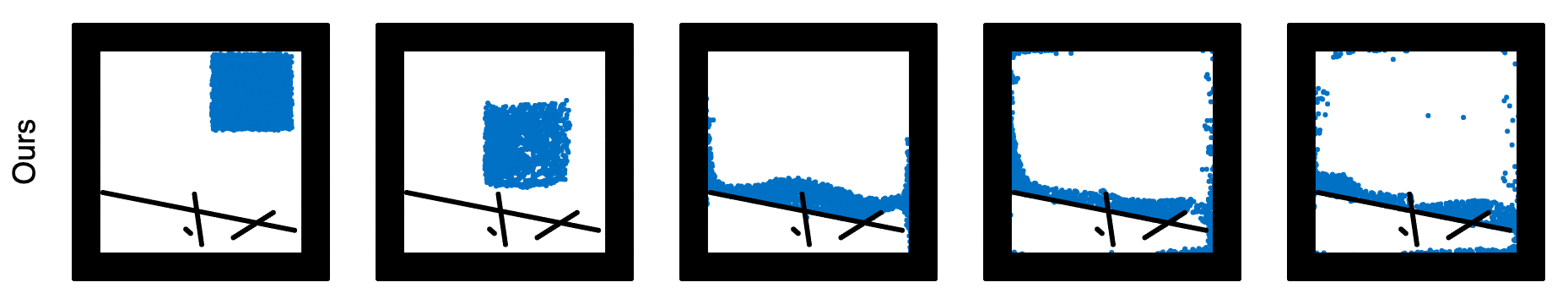} 
	\caption{Test sequences with methods trained on the \texttt{WaterRamps} data set.}
\end{figure}

\begin{figure}[h!]
	\centering
	\begin{overpic}[width=\textwidth, trim=0 20 0 200, clip]{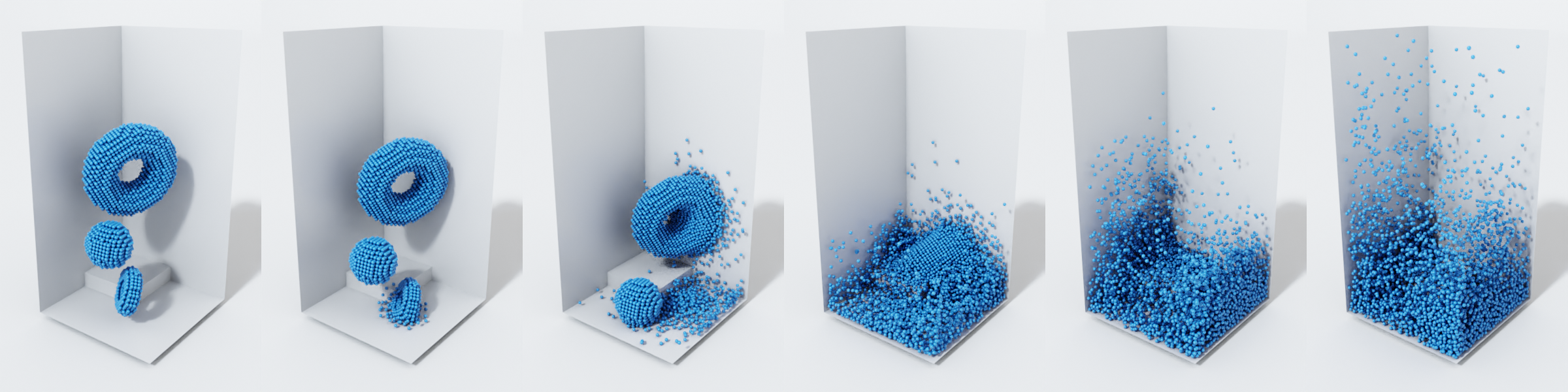}
	    \put(-3,6){\rotatebox{90}{GT}}\end{overpic}	    
	\begin{overpic}[width=\textwidth, trim=0 20 0 200, clip]{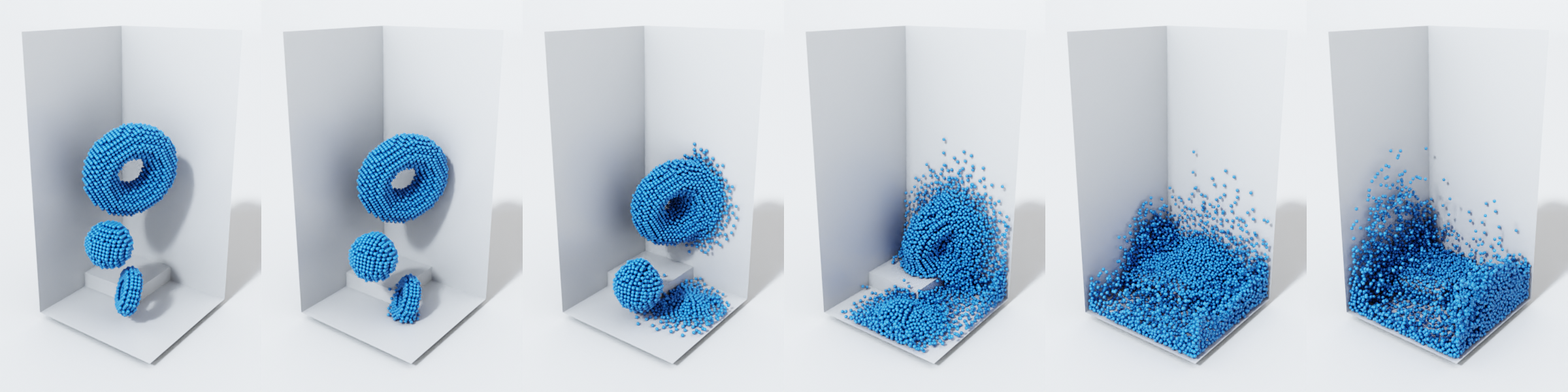}
	    \put(-3,6){\rotatebox{90}{CConv}}\end{overpic}	    
	\begin{overpic}[width=\textwidth, trim=0 20 0 200, clip]{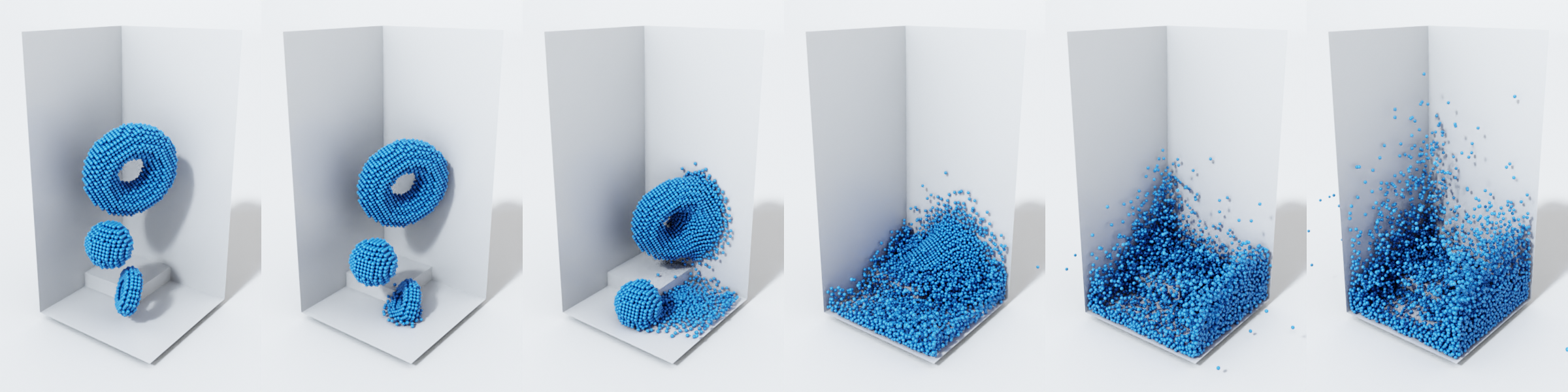}
	    \put(-3,6){\rotatebox{90}{Ours}}\end{overpic}	    
    \vspace{3mm}
    
	\begin{overpic}[width=\textwidth, trim=0 20 0 200, clip]{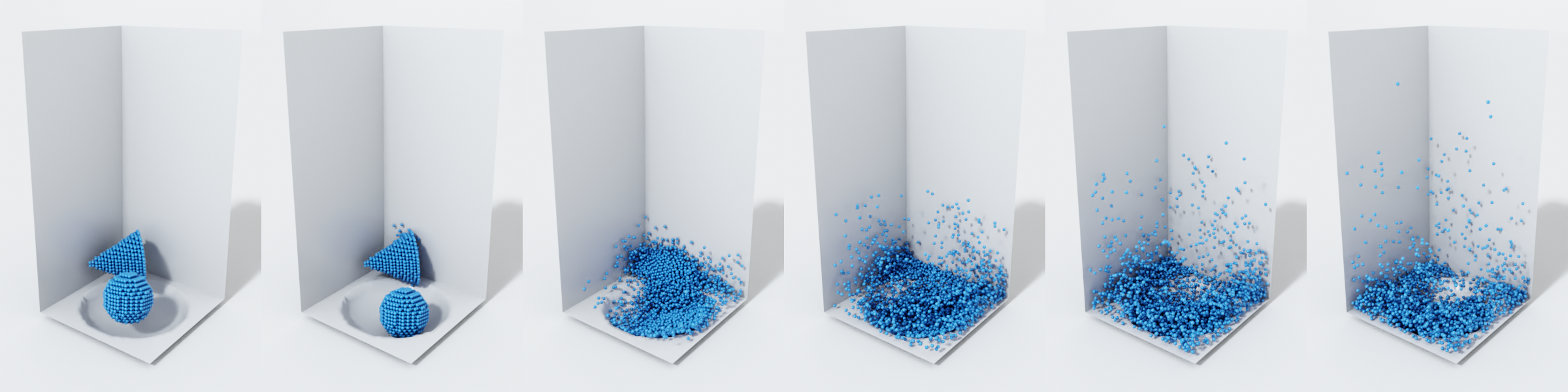}
	    \put(-3,6){\rotatebox{90}{GT}}\end{overpic}	    
	\begin{overpic}[width=\textwidth, trim=0 20 0 200, clip]{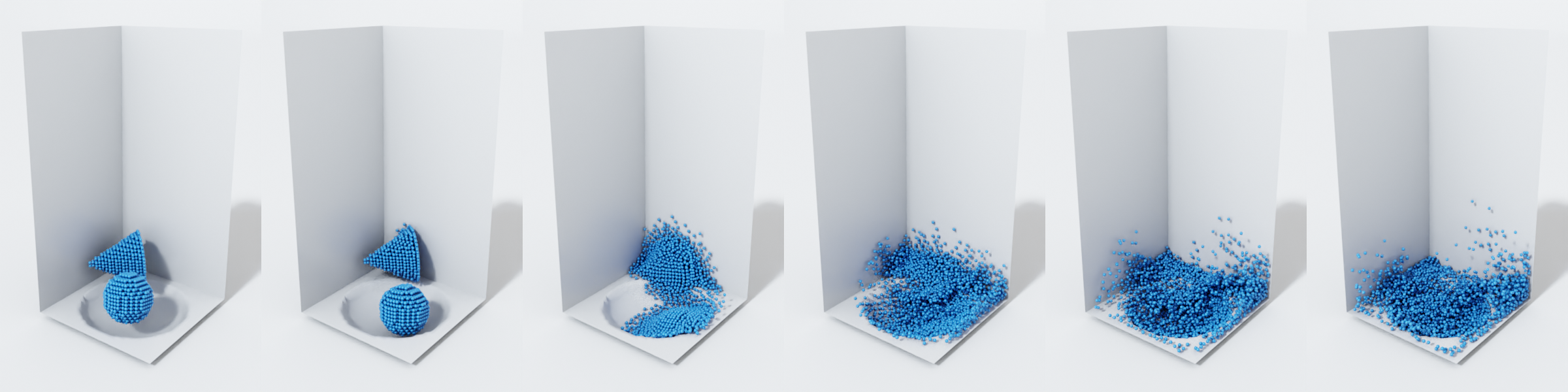}
	    \put(-3,6){\rotatebox{90}{CConv}}\end{overpic}	    
	\begin{overpic}[width=\textwidth, trim=0 20 0 200, clip]{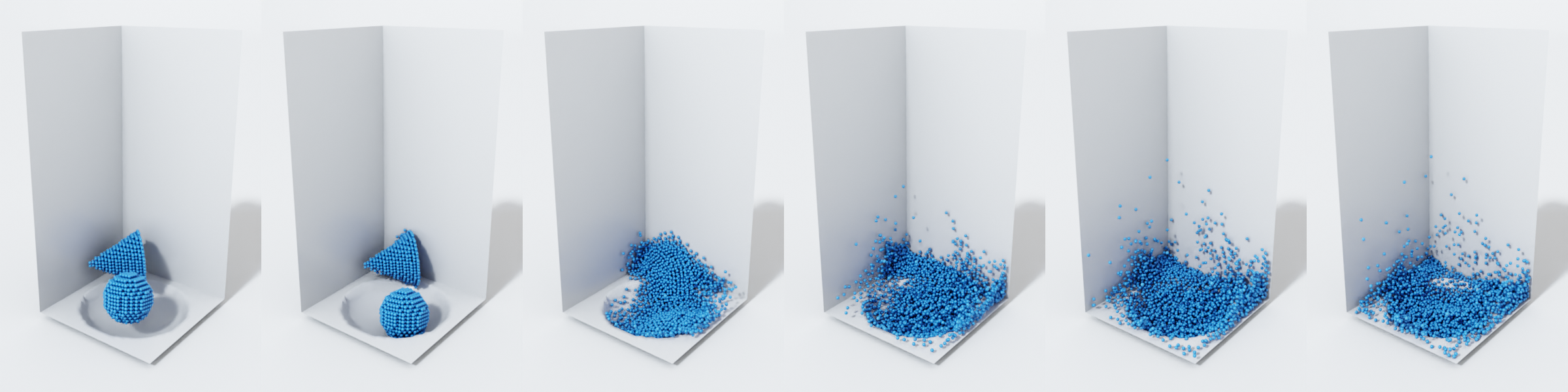}
	    \put(-3,6){\rotatebox{90}{Ours}}\end{overpic}	\caption{Test sequences based on the \texttt{Liquid3d} data set.}
\end{figure}

\begin{figure}[h!]
	\centering
	\includegraphics[width=\textwidth, trim=0 110 0 0, clip]{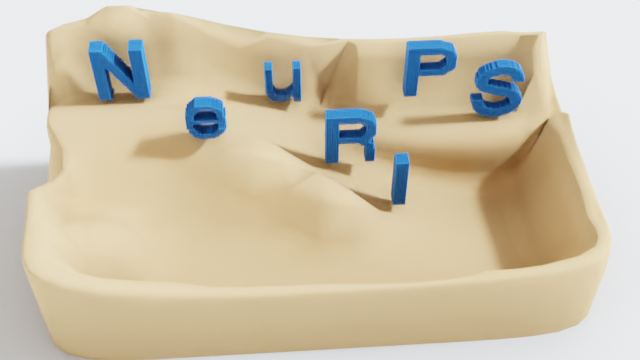} 
	\includegraphics[width=\textwidth, trim=0 110 0 0, clip]{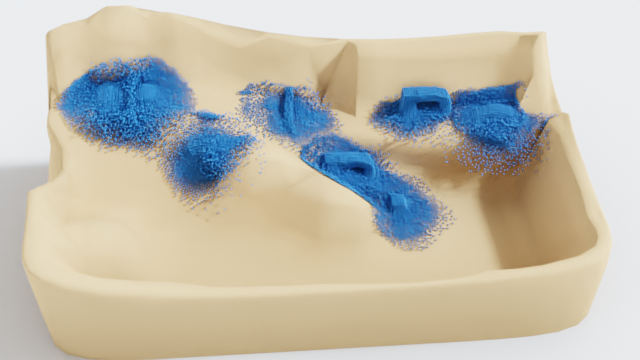} 
	\includegraphics[width=\textwidth, trim=0 50 0 60, clip]{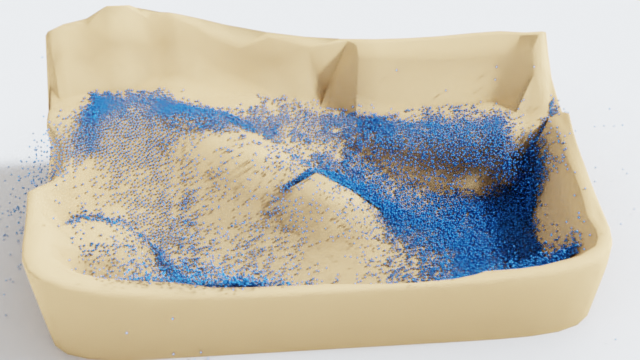} 
	\includegraphics[width=\textwidth, trim=0 10 0 100, clip]{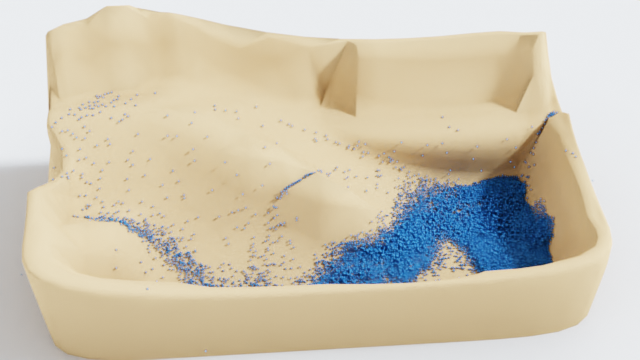} 
	\caption{A complex test sequence with a $5\times$ larger particle count than the training sequences from \texttt{Liquid3d} demonstrating scalability and generalization.}
	\label{fig:3d_big}
\end{figure}

\clearpage

\begin{table}    \centering
    \begin{tabular}{l|c|c}
         & Column & Free Fall \\
         \hline
         & RMSE ($\times 10^{-3}$) & RMSE ($\times 10^{-3}$) \\
         \hline
         SPH & 9.87 & 19.68 \\
         No Sym. & 0.04315 & 46.36843 \\
         ASCC & 0.06231 & 5.91239
    \end{tabular}
    \caption{Quantitative evaluation based on the  \texttt{Liquid Column} data set. Numbers correspond to Fig. \ref{fig:column} of the main paper.}
    \label{tab:column}
\end{table}
\begin{table}    \centering
    \begin{tabular}{l|c|c|c|c|c|c}
         & RMSE & Vel. Dist. & Momentum & Max. Dens. & EMD & Average \\
         \hline
         Base           & 1.227 & 0.619 & 0.000 & 0.510 & 0.072 & 0.486     \\
         ASCC           & 0.964 & 0.973 & 1.000 & 0.533 & 0.185 & 0.731  \\
         Multi Scale    & 1.125 & 0.929 & 1.000 & 0.546 & 0.408 & 0.802   \\
         Voxelize       & 1.227 & 0.913 & 1.000 & 0.548 & 0.385 & 0.815   \\
         Preprocess     & 0.964 & 1.002 & 1.000 & 0.834 & 0.922 & 0.945   \\
         Ours           & 1.000 & 1.000 & 1.000 & 1.000 & 1.000 & 1.000
    \end{tabular}
    \caption{    Quantitative evaluation for the ablation study. Numbers correspond to Fig. \ref{fig:ablation} of the main paper.}
    \label{tab:ablation}
\end{table}
\begin{table}    \centering
    \begin{tabular}{l|c|c|c|c|c|c||c|c}
         & \multicolumn{2}{|c|}{Random Gravity} 
         & \multicolumn{2}{|c|}{Tank}
         & \multicolumn{2}{|c||}{Two Drops}
         & \multicolumn{2}{|c}{Overall} \\
         \hline
         & RMSE & EMD & RMSE & EMD 
         & RMSE & EMD & RMSE & EMD \\
         \hline
         PointNet   & 0.11   & 47.3579   & 0.04   & 15.5390   & 0.07    & 1.9133    & 0.06  & 6.455210 \\
         CConv      & 0.16   & 221.5037  & 0.1    & 242.06389 & 0.21   & 9.4910     & 0.17  & 120.637445 \\
         GNS        & 0.1579 & 0.31152   & 0.512  & 0.2952    & 0.184   & 0.2628    & 0.1445  & 0.301795 \\
         Ours       & 0.12   & 0.134     & 0.05   & 0.00975   & 0.055  & 0.0117     & 0.07  & 0.041795
    \end{tabular}
    \caption{Quantitative evaluation based on the  \texttt{WBC-SPH} data set. Numbers correspond to Fig. \ref{fig:fluid_eval} of the main paper. RMSE was multiplied with $10^{-3}$.}
    \label{tab:wbcsph}
\end{table}
\begin{table}     \centering
    \begin{tabular}{l|c|c|c|c}
         & RMSE ($\times 10^{-3}$) & EMD & Params. ($\times 10^{-6}$)  \\
         \hline
         CConv      & 0.02  & 0.29296 & 0.18 \\
         GNS        & 0.092  & 0.08109 & 1.59 \\
         Ours 5steps   & 0.11  & 0.09155 & 0.47 \\ 
         Ours       & 0.11  & 0.06156 & 0.47 
    \end{tabular}
    \caption{Quantitative evaluation based on the  \texttt{WaterRamps} data set. Numbers correspond to Fig.~\ref{fig:gns_eval}, Fig.~\ref{fig:gns_quan}, and Fig.~\ref{fig:long_emd} of the main paper.}
    \label{tab:waterramps}
\end{table}

\begin{table}
    \centering
    \begin{tabular}{l|c|c}
         & Two Drops & Two Drops w/o Grav. \\
         \hline
         & EMD & EMD \\
         \hline
         GNS      & 0.08852  & 0.2012 \\
         Ours       & 0.05623  & 0.0682
    \end{tabular}
    \caption{Quantitative evaluation of the \texttt{Two Drops} generalization test, trained with \texttt{WaterRamps} data set. Numbers correspond to Fig. \ref{fig:mom} of the main paper.}
    \label{tab:mom}
\end{table}
\begin{table}
    \centering
    \begin{tabular}{l|c|c}
         & RMSE ($\times 10^{-3}$) & EMD \\
         \hline
         CConv      & 1.69  & 0.264 \\
         Ours       & 2.35  & 0.2143
    \end{tabular}
    \caption{Quantitative evaluation based on the  \texttt{Liquid3d} data set. Numbers correspond to Fig. \ref{fig:3d_qual} of the main paper.}
    \label{tab:liquid3d}
\end{table}
\begin{table}
    \centering
    \begin{tabular}{l|c|c}
         & Inference Time [ms] & Max. \# Particles (approx) \\
         \hline
         WBC-SPH      & 67.25  & 15k \\
         WaterRamps       & 17.62  & 2.3k \\
         Liquid3d       & 94.86  & 6k \\
         WBC-SPH (Solver)    & 10925  & -
    \end{tabular}
    \caption{Average inference time for single frames, and approximate maximum number of particles, corresponding to Fig. \ref{fig:time_cnt}.}
    \label{tab:time_cnt}
\end{table}
\begin{table}
    \centering
    \begin{tabular}{l|c}
         & Inference Time [ms] \\
         \hline
         CConv      & 2.57 \\
         GNS       & 30.63 \\
         Ours       & 10.98 
    \end{tabular}
    \caption{Average inference time for single frames corresponding to Fig. \ref{fig:time_inf}.}
    \label{tab:time_inf}
\end{table}
\begin{table}
    \centering
    \begin{tabular}{c|c|c}
         Noise Ratio & GNS (EMD) & Ours (EMD) \\
         \hline
        0\%	    & 0.01665 & 0.01409 \\
        1\%	    & 0.01673 & 0.0146 \\
        2\%		& 0.01715 & 0.01573 \\
        5\%	    & 0.02104 & 0.01909 \\
        10\%	& 0.02415 & 0.02093 \\
        20\%	& 0.02551 & 0.02264 \\
    \end{tabular}
    \caption{Accuracy evaluation for varying amounts of input noise corresponding to Fig. \ref{fig:noise}.}
    \label{tab:noise}
\end{table}
	
\begin{table}
    \centering
    \begin{tabular}{c|c|c|c|c}
         Sampling Ratio & GNS (EMD) & GNS (rel.) & Ours (EMD) & Ours (rel.) \\
         \hline
        100\%	    & 0.09555 & 0.10309 & 100\% & 100\%	\\
        75\%	    & 0.11544 & 0.12123 & 82.77\% & 85.04\% \\
        50\%		& 0.15092 & 0.01573 & 63.31\% & 85.35\% \\
        25\%	    & 0.22921 & 0.12078 & 41.69\% & 59.85\%
    \end{tabular}
    \caption{Relative accuracy evaluation for different sampling densities corresponding to Fig. \ref{fig:sampling_rel}.}
    \label{tab:sampling}
\end{table}
\begin{table}
    \centering
    \begin{tabular}{c|c|c}
         Train Set Size & Unconstrained (EMD) & Constrained (EMD) \\
         \hline
    0.39\%	& 0.1861 & 0.08066 \\	
    3.13\%	& 0.0584 & 0.03881 \\	
    9.38\%	& 0.0424 & 0.02488 \\	
    20.70\%	& 0.0275 & 0.02028 \\	
    100\%	& 0.0280 & 0.02133 \\	
    \end{tabular}
    \caption{Accuracy evaluation for different training set sizes corresponding to Fig. \ref{fig:train_size_emd}.}
    \label{tab:train_size}
\end{table}

\end{document}